\documentclass[journal,compsoc]{IEEEtran}
\ifCLASSINFOpdf
\else
\fi
\usepackage{amsmath}
\usepackage{tikz,mathpazo}
\usetikzlibrary{shapes.geometric, arrows, positioning}
\usetikzlibrary{calc} 

\usepackage{xcolor,soul}
\usepackage[colorlinks=true,linkcolor=blue,citecolor=blue]{hyperref}

\usepackage{comment}
\usepackage{longtable}
\usepackage[figuresright]{rotating}
\usepackage{booktabs}
\usepackage{framed,multirow}
\usepackage{multicol}
\newcommand{\etal}{{\em et al.}}       
\newcommand{\eg}{{\em e.g.}}           
\newcommand{\ie}{{\em i.e.}}           
\newcommand{\etc}{{etc}}         

\begin{document}
%
\title{Transformers in Medical Image Analysis:\\ A Review}
%
%
%

\author{Kelei~He,~
        Chen~Gan,~
        Zhuoyuan~Li,~
        Islem~Rekik,~
        Zihao~Yin,~
        Wen~Ji,~
        Yang~Gao,~
        Qian~Wang*,~\\
        Junfeng~Zhang*,~
        and~Dinggang~Shen*,~\IEEEmembership{Fellow,~IEEE}
\thanks{This work was supported in part by the National Nature Science Foundation of China under grant No. 62106101. This work was also supported in part by the Natural Science Foundation of Jiangsu Province under grant No. BK20210180. This work was also supported in part by the grant from Jiangsu Provincial Key R$\&$D Program under No. BE2020620 and BE2020723.}
\thanks{Co-corresponding authors: Junfeng~Zhang (jfzhang@nju.edu.cn); Dinggang~Shen (Dinggang.Shen@gmail.com); Qian Wang (wangqian2@shanghaitech.edu.cn).}
\thanks{K. He, C. Gan, Z. Li and I. Rekik contributed equally to this work.}
\thanks{K. He, Z. Li, and J. Zhang are with Medical School of Nanjing University, Nanjing, P. R. China. K. He, C. Gan, Z. Li, Z. Yin, W. Ji, Y. Gao, J. Zhang are also with National Institute of Healthcare Data Science at Nanjing University, Nanjing, P. R. China. Y. Gao is also with State Key Laboratory for Novel Software Technology, Nanjing University, Nanjing, P. R. China.}
\thanks{I. Rekik is with BASIRA lab, Faculty of Computer and Informatics Engineering, Istanbul Technical University, Istanbul, Turkey. She is also with School of Science and Engineering, Computing, University of Dundee, UK.}
\thanks{Q. Wang and D. Shen are with School of Biomedical Engineering, ShanghaiTech University, Shanghai, P. R. China. D. Shen is also with Department of Research and Development, Shanghai United Imaging Intelligence Co., Ltd., Shanghai, P. R. China.}}

%
%

\markboth{Journal of \LaTeX\ Class Files,~Vol.~14, No.~8, August~2015}%
{Shell \MakeLowercase{\textit{et al.}}: Bare Demo of IEEEtran.cls for IEEE Journals}
%



\maketitle


\begin{abstract}
Transformers have dominated the field of natural language processing, and recently impacted the computer vision area. In the field of medical image analysis, Transformers have also been successfully applied to full-stack clinical applications, including image synthesis/reconstruction, registration, segmentation, detection, and diagnosis. Our paper aims to promote awareness and application of Transformers in the field of medical image analysis. Specifically, we first overview the core concepts of the attention mechanism built into Transformers and other basic components. Second, we review various Transformer architectures tailored for medical image applications and discuss their limitations. Within this review, we investigate key challenges revolving around the use of Transformers in different learning paradigms, improving the model efficiency, and their coupling with other techniques. We hope this review can give a comprehensive picture of Transformers to the readers in the field of medical image analysis.
\end{abstract}

\begin{IEEEkeywords}
Transformers, Medical image analysis, Deep learning, Diagnosis, Registration, Segmentation, Image synthesis, Multi-task learning, Multi-modal learning, Weakly-supervised learning.
\end{IEEEkeywords}

%
\IEEEpeerreviewmaketitle

\section{Introduction}

\IEEEPARstart{T}{ransformers} \cite{vaswani_attention_2017} have dominated the field of natural language processing (NLP), including speech recognition \cite{dong2018speech}, synthesis \cite{li2019neural}, text to speech translation \cite{vila2018end}, and natural language generation \cite{topal2021exploring}. As a compelling instance of deep learning architectures, Transformer was firstly introduced to handle sequential inference tasks in NLP. While recurrent neural networks (RNNs) \cite{graves2013speech} (\eg, long short-term memory network (LSTM) \cite{sak2014long}) explicitly use a sequence of inference processes, Transformers remarkably capture long-term dependencies of sequential data with stacked self-attention layers. In this manner, Transformers are both efficient as they solve the sequential learning problem in one-shot, and effective by stacking very deep models. Several Transformer architectures trained on large-scale architectures have became widely popular in solving NLP tasks such as BERT \cite{devlin2018bert} and GPT \cite{radford2018improving,brown2020language} -- to just name a few.

Convolutional neural network (CNNs) and its variants have achieved the state-of-the-art in several computer vision (CV) tasks \cite{shen2017deep}, partially thanks to their progressively enlarged receptive fields that can learn hierarchies of structured image representations as semantics. Capturing vision semantics in images is usually regarded as the core idea of building successful networks in computer vision \cite{krizhevsky2012imagenet}. However, the long-term dependencies within images such as the non-local correlation of objects in the image are neglected in CNNs. Inspired by the aforementioned success of Transformers in NLP, Dosovitskiy \etal \cite{dosovitskiy2020vit} proposed the Vision Transformer (ViT) by formulating image classification as a sequence prediction task of the image patch (region) sequence, thereby capturing long-term dependencies within the input image. ViT and its derived instances have achieved the state-of-the-art performance on several benchmark datasets. Transformers have become very popular across a wide spectrum of computer vision tasks, including image classification \cite{dosovitskiy2020vit}, detection \cite{detr}, segmentation \cite{zheng2021rethinking}, generation \cite{parmar2018image}, and captioning \cite{li2019entangled}. Furthermore, Transformers also play an important role in video-based applications \cite{zhou2018end}.

\begin{figure*}
    \centering
    \includegraphics[width=\linewidth]{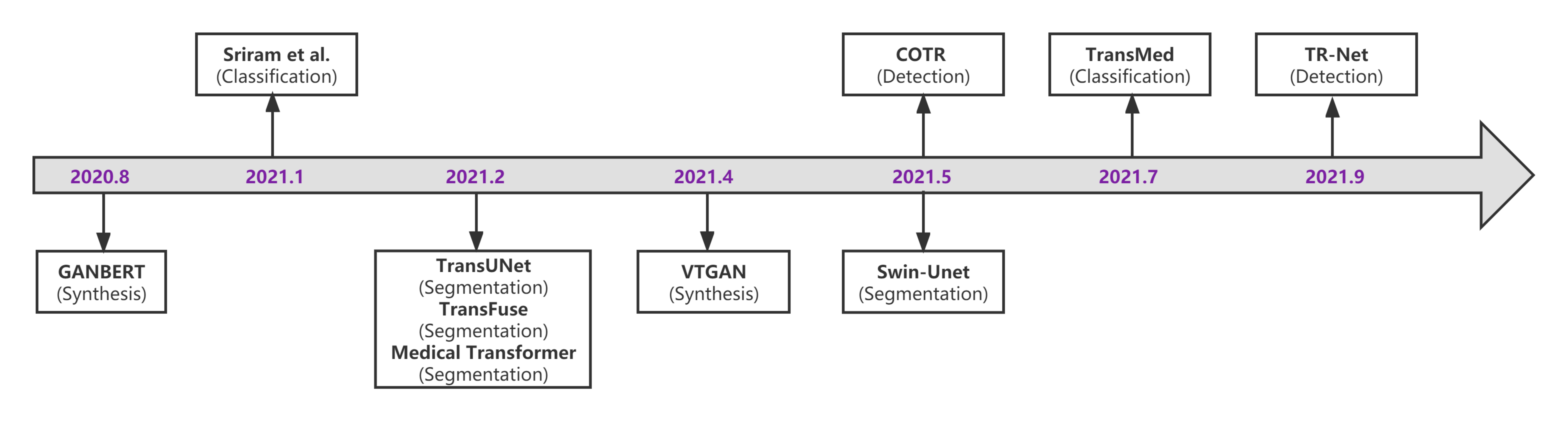}
    \caption{The development of Transformers in medical image analysis. The figures display selected methods in the applications of classification, detection, segmentation and synthesis.}
    \label{fig:roadmap}
\end{figure*}

Recently, Transformers have also cross-pollinated the field of medical image analysis for disease diagnosis \cite{PureTrans-COVID-VIT-gao2021covidvit,SwinTrans-MIA-COV19D-zhangmia,GLTrans-he2021globallocal} and other clinical purposes. For instance, the works in \cite{PureTrans-BrazilViT-costa2021covid,CNN+Trans-Multi-View-van2021multi} utilized Transformers to distinguish COVID-19 from other types of pneumonia using computed tomography (CT) or X-ray images, meeting the urgent need of treating COVID-19 patients fast and effectively. Besides, Transformers were successfully applied to image segmentation \cite{zhang2021pyramid}, detection \cite{xie2021cotr}, and synthesis \cite{watanabe2021generative}, remarkably achieving state-of-the-art results. Fig. \ref{fig:roadmap} displays the chronological adaptation of Transformers to different medical image applications, and will be further discussed in Section \ref{sec:application}.

Although many researches were devoted to customize Transformers to medical image analysis tasks, such customization stirred new challenges that remain unsolved. To encourage and facilitate the development of Transformer-based applications in medical image analysis, we extensively reviewed more than $170$ existing Transformer-based methods in the field, providing solutions for medical applications, and showing how Transformers were adopted in various clinical settings. Moreover, we present in-depth discussions on designing Transformer-based methods for solving more complex real-world tasks, including weakly-supervised/multi-task/multi-modal learning paradigms. This paper includes comparisons between Transformers and CNNs, and discusses new ways of improving the efficiency and interpretation of Transformer networks.

The following sections are organized as follows. Section \ref{sec:trans} introduces the preliminaries of Transformers and its development in vision. Section \ref{sec:application} reviews recent applications of Transformers in medical image analysis, and Section \ref{sec:discus} discusses the potential future directions of Transformers. Section \ref{sec:con} concludes the paper.

\section{Transformers}
\label{sec:trans}

\subsection{Preliminaries}

A typical Transformer leverages the attention mechanism in neural networks. Hence, we start by introducing the core principle of the attention mechanism, and then we give a detailed description of how the Transformer works.

\subsubsection{Attention mechanism}

For information exploration, human beings usually leverage the 'attention mechanism' to filter out irrelevant information while focusing on the meaningful parts of the data in daily life. Inspired by this observation, researchers designed the attention mechanism for deep learning that seep through homogeneous data while \emph{giving attention} to the most significant components or elements.

\noindent\textbf{Bahdanau attention.}

The attention mechanism was initially proposed by Bahdanau \etal \cite{bahdanau_neural_2016} for the language translation task, namely Bahdanau attention. 
This attention mechanism is calculated by the weighted sum of all annotations (\ie, the results of each input generated by the encoder) and the previous decoder. 

\subsubsection{Attention mechanism in computer vision}

Similar concepts have also been developed in the field of computer vision. For example, Hu \etal \cite{hu_squeeze-and-excitation_2019} introduced a novel attention mechanism, \ie, \textit{Squeeze-and-Excitation}, to execute \textit{feature re-calibration}, in which informative features to a particular visual task are emphasized, and the remaining features are regarded as less-important ones.

\begin{figure}[htbp]
    \centering
    \includegraphics[width=\linewidth]{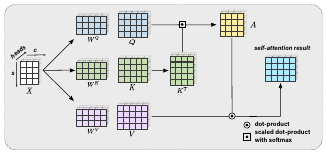}
    \caption{A brief illustration of self-attention mechanism.}
    \label{fig:self-attention}
\end{figure}

\noindent\textbf{Self-attention.}
In \cite{vaswani_attention_2017}, the attention mechanism is re-defined as a function working with queries, keys, and values, which are derived from the input vectors of the module, unlike that in Bahdanau attention.
The output is defined as a weighted sum of values, where the weight of each value is calculated as the attention between queries and keys. 

The self-attention operation is usually performed in matrix form to accelerate the calculation in parallel.
To briefly illustrate the concept of self-attention, we first describe it in an element-wise form. 

For each input $x_i \in \mathbb{R}^c$, $i=1,..,n$,
its corresponding vectors of query $q_i \in \mathbb{R}^d_q$, key $k_i \in \mathbb{R}^d_k$, and value $v_i \in \mathbb{R}^d_v$ are generated through the parameters, \ie, $W^q$, $W^k$, $W^v$, respectively. $d_q, d_k, d_v$ are the sizes of $q_i, k_i, v_i$, and also the number of features that are learnt from $x_i$.

\begin{equation}{}
\begin{array}{ccclcl}
    q_i = x_i \times W^q,& W^q \in \mathbb{R}^{c \times d_q},\\
    k_i = x_i \times W^k,& W^k \in \mathbb{R}^{c \times d_k},\\
    v_i = x_i \times W^v,& W^v \in \mathbb{R}^{c \times d_v},\\
    d_q = d_k.
\end{array}
\label{eq:qkv}
\end{equation}

The output is also a probability calculated by the weighted sum of the calculated weighting values,
\begin{equation}
    \alpha_{i j} = \operatorname{Softmax}(\frac{\alpha'_{i j}}{\sqrt{d_k}})=
    \frac{exp(\frac{\alpha'_{i j}}{\sqrt{d_k}})}{\sum_{j}exp(\frac{\alpha'_{i j}}{\sqrt{d_k}})}.
\end{equation}
\begin{equation}
    \alpha'_{i j}=q_i \times k_j^\text{T},
\end{equation}

where $\alpha'_{i j}$ measures the contribution of the $j^{th}$ element of the input to the $i^{th}$ element of the output. Through this operation, $\alpha'_{i j}$ is regarded as the attention assigned to the element $v_i$. Thereby the final output attentions can be computed as a weighted sum of all values as follows:

\begin{equation}
    z_{i}=\sum_{j}{\alpha_{i j} \times v_j}.
\end{equation}

The element-wise self-attention can be feasibly extended to matrices. In most cases, the query $q_i$, key $k_i$ and value $v_i$ for each input $x_i$ are generated using parallel matrix computation. $x_i, q_i, k_i, v_i$ can be stacked together to matrices, respectively. Let $X \in \mathbb{R}^{s \times c}$ denote the input matrix, $Q$ denote the query matrix, $K$ denote the key matrix, and $V$ denote the value matrix, where $s$ is the number of the samples, and each matrix is consisted of the elements, \ie, $X=[x_1;x_2;\cdots;x_s]^T$. Similarly, We compute the attention matrix $A$ and output matrix $Z$ as follows,

\begin{align}
    A &= \operatorname{Softmax}(\frac{Q \times K^{\text{T}}}{\sqrt{d_k}}) \in \mathbb{R}^{s \times s}, \\
    Z &= A \times V \in \mathbb{R}^{s \times d_v}.
\end{align}

\noindent\textbf{Multi-head self-attention.}
The work in \cite{vaswani_attention_2017} showed that applying multiple self-attentions to the same input can better capture hierarchical features. These self-attention layers work similarly to multiple kernels in convolution layers. Given $h$ self-attentions (heads), the module outputs the final result by concatenating the calculated attentions.

\begin{align}
Z_i = \operatorname{Attention}(Q \times W_i^Q, K \times W_i^K, V \times W_i^V),\\
\operatorname{MultiHead}(Q, K, V) =\operatorname{Concat}(Z_1, \cdots,Z_h) W^O,
\end{align}

where $W_i^Q$, $W_i^J$, $W_i^V$ denote linear projection matrices, which map matrices $Q, K, V$ into different subspaces, respectively. And $W^O$ is an output projection matrix that concatenates self-attention outputs of all attention heads.

\subsection{Architecture}

\begin{figure}
    \centering
    \includegraphics[width=\linewidth]{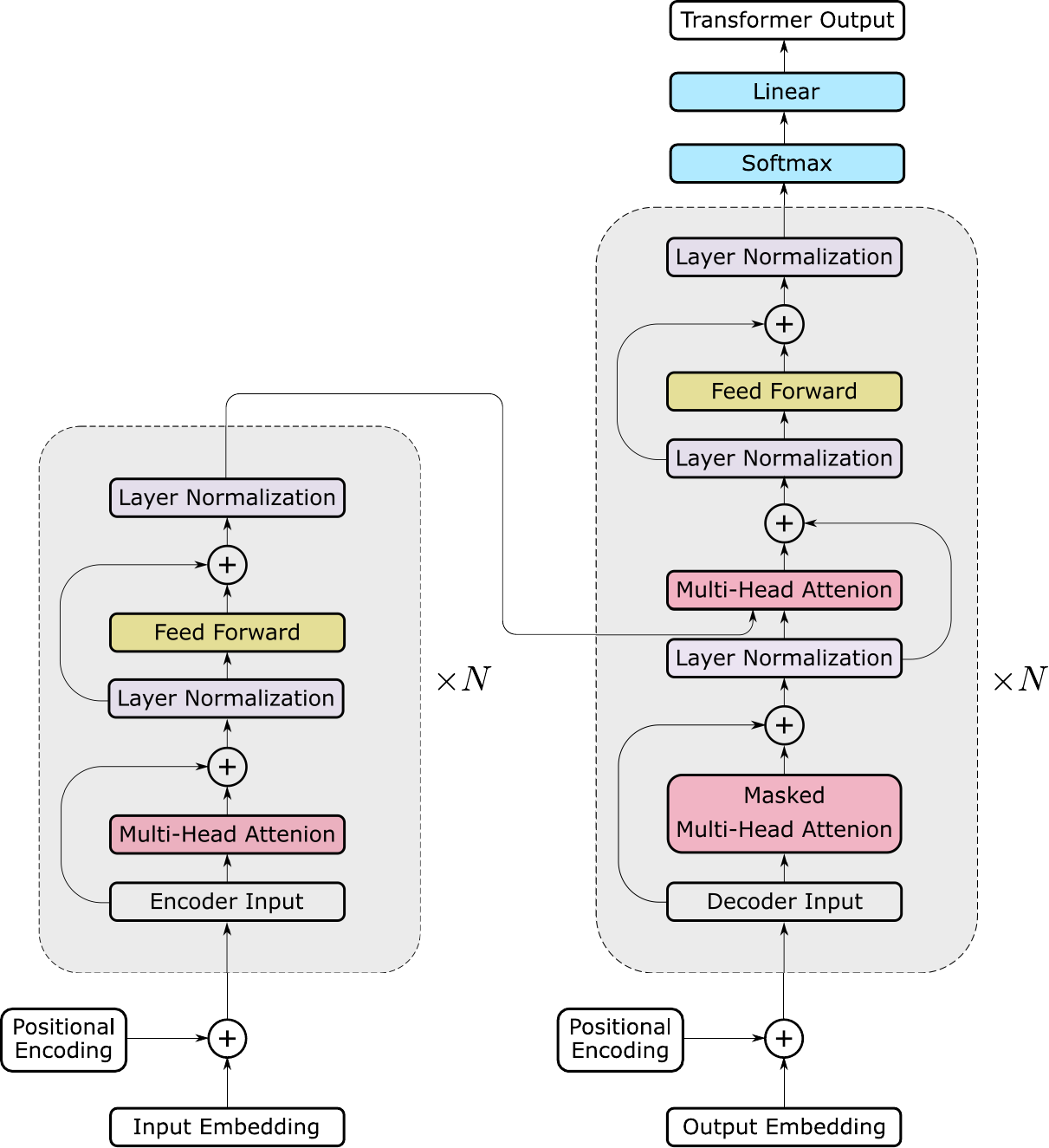}
    \caption{A brief illustration of a typical Transformer architecture \cite{chen2021vit}.}
    \label{fig:Transformer}
\end{figure}

In \cite{vaswani_attention_2017}, the authors proposed a typical \emph{Transformer} network, with an encoder-decoder structure. The encoder maps an input sequence $\{x_1, \dots, x_n\}$ to an output sequence $\{z_1, \cdots, z_n\}$ with the same length. The decoder generates the output $\{y_1, \cdots, y_m\}$ from the encoded representation $z$ in an element-wise manner, and takes the previous output as an additional input. A typical Transformer architecture is shown in Fig. \ref{fig:Transformer}, and will be described below.

\subsubsection{Encoder}

The encoder in a typical Transformer has $n=6$ stacked blocks, consisting of two types of layers, \ie, the multi-head attention layer and the feed-forward layer. Residual connections and layer normalization layers are also combined with the aforementioned layers. Concretely, in each block, the multi-head attention is firstly calculated, followed by a layer-wise normalization, calculating by the sum of the input and output of the multi-head attention. Then, a feed-forward layer is performed, followed by a layer-wise normalization of the sum of the feed-forward layer's input and output.

\subsubsection{Decoder}

The decoder also has $n=6$ blocks, similar to the encoder, with some minor modifications. Specifically, an additional self-attention layer is inserted on the top of the encoded output. Masking is employed in the first self-attention layer to block subsequent contributions to the state of the previous position,
since the prediction is based on a known state.
A linear layer and a Softmax layer are inserted after the output of the decoder to generate the final output.

\subsection{Vision Transformer}

The success of Transformers in NLP propagated to the computer vision (CV) research community where several efforts have been made to adapt Transformers to vision tasks. Till now, Transformer-based models in vision have been developed at an unprecedented pace, among which the most representative ones are DEtection TRansformer (DETR) \cite{detr}, Vision Transformers (ViT) \cite{dosovitskiy2020vit}, Data-efficient image Transformers (DeiT) \cite{Data-Efficient-Deit} and Swin-Transformer \cite{liu2021swin}.

\noindent\textbf{DETR.} DETR, proposed by Carion \etal \cite{detr} was the first work that applied Transformers to CV tasks, specifically for the task of object detection. Unlike the conventional object detection methods that consist of some hand-crafted processes, DETR is an end-to-end detection model which utilizes a Transformer encoder to model the relation between image features extracted by a CNN backbone, a Transformer decoder to generate object queries, and a feed-forward network to assign labels and bound the boxes around the objects. 

\begin{figure}[htbp]
    \centering
    \includegraphics[width=\linewidth]{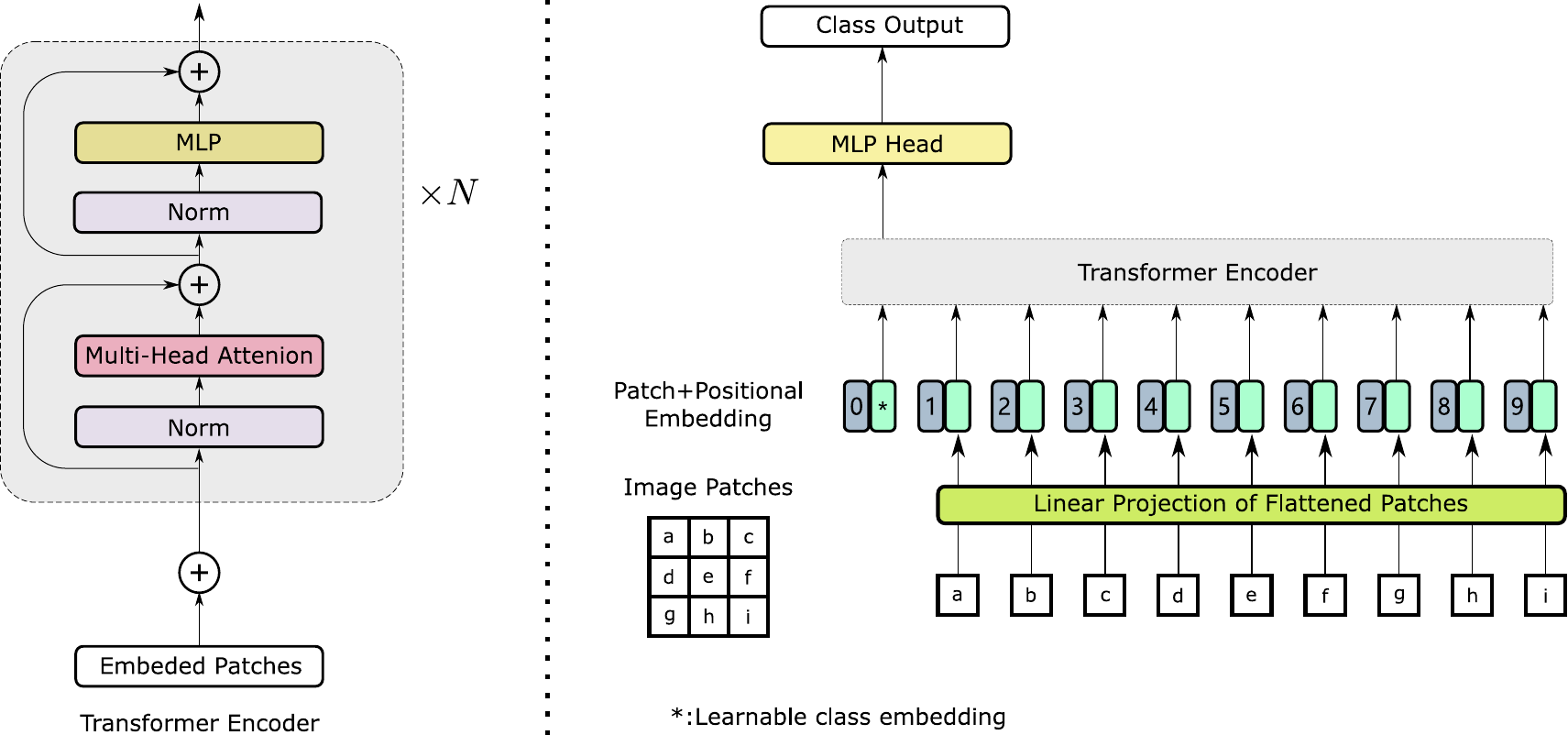}
    \caption{A typical architecture of ViT \cite{dosovitskiy2020vit}. It uses sequential image patches as the input, process it with the Transformer Encoder, and output the class prediction by a MLP head. The Transformer Encoder is built by a number of $N$ transformer blocks.}
    \label{fig:ViT}
\end{figure}

\noindent\textbf{ViT.} Following DETR, Dosovitskiy \etal \cite{dosovitskiy2020vit} proposed the vision Transformer (ViT), as shown in Fig.\ref{fig:ViT}. ViT is an image classification model that basically adopts the architecture of the conventional Transformer. In ViT, the input image is converted to a series of patches, each coupled with a positional encoding method that encodes the spatial positions of each patch to provide spatial information. The patches along with a class token are then fed into the Transformer to calculate the MHSA and output the learned embeddings of patches. The state of the class token from the output of the ViT serves as the image representation. At last, a multi-layer perception (MLP) is used to classify the learned image representation. Apart from the raw images, the feature maps from CNNs can also be fed into a ViT for relational mapping.

\noindent\textbf{DeiT.} In order to solve the problem of large-scale training data required by ViT, Touvron \etal \cite{Data-Efficient-Deit} proposed the DeiT to ensure its performance on small-scale data. They adopted a knowledge distillation framework with a teacher-student formulation, and attached a distillation token (this is a terminology for Transformers) after the input sequence to learn from the output of the teacher model. In addition, they argued that utilizing CNNs as the teacher model can facilitate the training of the Transformer as the student network to inherit the inductive bias.

\noindent\textbf{Swin-Transformer.} To reduce the cost of calculating the attention of high-resolution images, and deal with the varied patch sizes in scene understanding tasks (\eg, segmentation), Liu \etal \cite{liu2021swin} proposed the Swin-Transformer. They introduced a window self-attention to reduce the computational complexity, and used the shifted window attention to model cross-window relationships. Moreover, they also connected these attention blocks with patch merging blocks, which are used to merge neighboring patches to produce a hierarchical representation for handling variations in the scale of visual entities.

\subsection{Other techniques}

Meanwhile, recent works also validated MLP-based models, and examined the effectiveness of attention mechanism, convolution and other modules in the CNNs or ViTs. Though CNNs and ViTs have been dominant for quite some time, the success of some MLP-based models has also caused great repercussions. A representative work is MLP-Mixer \cite{Tolstikhin2021mlp-mixer}.
The MLP-Mixer was proposed by Tolstikhin \etal \cite{Tolstikhin2021mlp-mixer} in May 2021, which used a simple pure deep MLP architecture but showed competitive performance. The MLP-Mixer adopts the per-patch flattening instead of the full flattening while the positional encoding and class token are not added to the patch sequence as in ViT. Following patch embedding learning, the Mixer MLP block is composed of a Token-mixing MLP and a Channel-mixing MLP, where the former is used to aggregate inter-patch features, and the latter is used to integrate intra-patch features. The final class is predicted based on the features obtained following the Global Average Pooling.

Simultaneously with or following MLP-Mixer, many other MLP-based models were proposed, \eg, gMLP \cite{liu202gmlp}, ResMLP \cite{Touvron2021resmlp}, ASMLP \cite{lian2021asmlp}, CycleMLP \cite{Chen2021cyclemlp}, \etc. MLP-Mixer not only inspired further exploration of MLP-based model, but also raised stimulated further developments of neural architectures in CV. Since Transformers, CNNs and MLPs have shown competitive performance against each other, there is still no evidence as to which architecture is more suited for a particular CV learning task. For medical image analysis, we will also discuss the comparison of CNN and Transformer in Part C of Section \ref{sec:discus}.

\section{Transformers in medical image applications}
\label{sec:application}

\begin{figure}
    \centering
    \includegraphics[width=\linewidth]{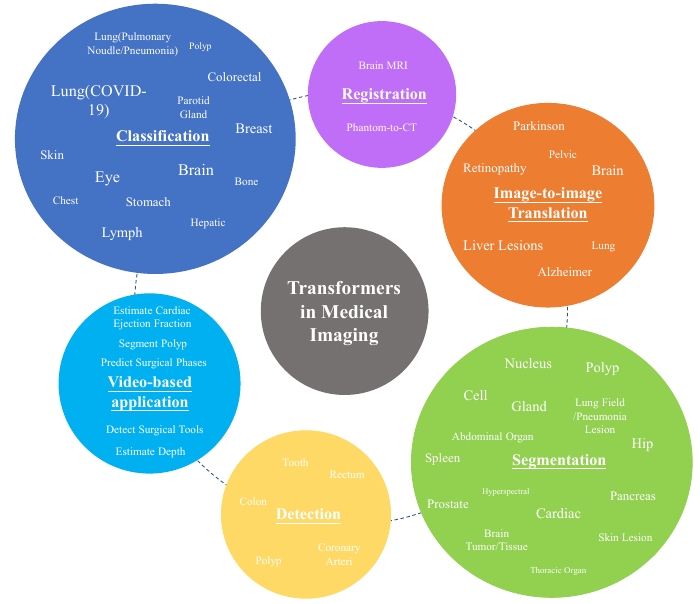}
    \caption{Applications of Transformers in medical image analysis, as reviewed in this work.}
    \label{fig:Applications}
\end{figure}

Transformers have been widely applied to full-stack clinical applications. In this section, we first introduce the Transformer-based medical image analysis applications, including classification, segmentation, image-to-image translation, detection, registration, and video-based applications. We categorize these applications according to their learning tasks as illustrated in Fig. \ref{fig:Applications}.

\subsection{Classification}

The methods using Transformers for both disease diagnosis and prognosis are formulated as the classification tasks, which can be divided into three categories, including: 

1) Applying ViTs directly to medical images;

2) Combining ViTs with convolutions for more representative local feature learning;

3) Combining ViTs with graph representations to better handle complex data.

This section will give a comprehensive overview of the three aforementioned Transformer categories used for classification tasks in medical images (Table \ref{Table:classification}).

\begin{table*}[htbp]
\renewcommand{\arraystretch}{0.5}
\centering
\caption{\label{Table:classification} Transformers used in medical image classification tasks.}
\setlength{\tabcolsep}{1pt}
\begin{tabular}{l|ccccc}
\toprule[1pt]
\textbf{Reference} & \textbf{Disease} & \textbf{Organ} & \textbf{Dataset} &\textbf{Highlight}  & \textbf{ACC(\%)} \\ 
\hline
\textbf{CT} &&&& \\
\cline{1-1}
\begin{tabular}{l}
\textbf{Costa \etal} \cite{PureTrans-BrazilViT-costa2021covid}
\end{tabular} 
& COVID-19
& Lung 
& COVIDx 
& \begin{tabular}{c} ViT with Performer \end{tabular}
 & \begin{tabular}{c} 91.0\\91.0\\96.0 \end{tabular}
\\

\begin{tabular}{l}
\textbf{COVID-VIT} \cite{PureTrans-COVID-VIT-gao2021covidvit}
\end{tabular}   
& COVID-19  
& Lung 
& \begin{tabular}{c}COV19-CT-DB 
\end{tabular} 
& \begin{tabular}{c} use sub-volumes for 3D images \end{tabular}
 & 76.6 
\\

\begin{tabular}{l}
\textbf{MIA-COV19D} \cite{SwinTrans-MIA-COV19D-zhangmia}
\end{tabular} 
& COVID-19 & Lung 
& \begin{tabular}{c}COV19-CT-DB 
\end{tabular} 
& \begin{tabular}{c} segment lung first, use Swin-Trans \end{tabular}
 & 94.3

\\

\begin{tabular}{l}
\textbf{Liang \etal} \cite{CNN+Trans-TransHybrid-liang2021hybrid}
\end{tabular} 
& COVID-19 
& Lung 
& COV19-CT-DB 
& \begin{tabular}{c} feature aggregation by Trans,\\CNN features, data resampling \end{tabular}
 & - 
\\

\begin{tabular}{l}
\textbf{Scopeformer} \cite{CNN+Trans-Scopeformer-barhoumi2021scopeformer}
\end{tabular} 
& \begin{tabular}{c} Intracranial 
\\ Hemorrhage \end{tabular} 
& Brain 
& \begin{tabular}{c} RSNA intracranial
\\ hemorrhage dataset 
\end{tabular} 
& \begin{tabular}{c} multiple CNNs,\\ GAN for domain alignment \end{tabular}
 & 98.0

\\

\begin{tabular}{l}
\textbf{Li \etal} \cite{li2021medical}
\end{tabular} 
& COVID-19 
& Lung 
& -
& \begin{tabular}{c} teacher-student model for\\ knowledge distillation\end{tabular}
 & -

\\

\begin{tabular}{c}
\textbf{Than \etal} \cite{than2021preliminary}
\end{tabular} 
& COVID-19 
& Lung 
& \begin{tabular}{c}COVID-CTset\cite{rahimzadeh2021fully}\end{tabular}
& \begin{tabular}{c}research on patch size\end{tabular}
& 95.4

\\

\begin{tabular}{l}
\textbf{Xia \etal} \cite{xia2021effective}
\end{tabular} 
& Pancreatic Cancer 
& Pancreas 
& -
& \begin{tabular}{c} Anatomy-Aware Transformer \\ with Localization Unet \end{tabular}
 & -

\\

\hline

\textbf{X-Ray} \\
\cline{1-1}

\begin{tabular}{l}\textbf{Park \etal} \cite{Pre-training-COVID-19-park2021vision}
\end{tabular} 
& COVID-19  
& Lung 
& -
& \begin{tabular}{c} pretrained backbone on CXRs\end{tabular}
 & -

\\

\begin{tabular}{l}\textbf{Tanzi \etal} \cite{tanzi-femur-fracture-x-ray} 
\end{tabular} 
& Femur Fracture  
& Bone 
& -  
& \begin{tabular}{c} unsupervised learning,\\ compare CNNs with ViTs \end{tabular} 
 & 77.0

\\

\begin{tabular}{l}\textbf{Van \etal} \cite{CNN+Trans-Multi-View-van2021multi} 
\end{tabular} 
&
\begin{tabular}{c}Mammography \\ Chest X-ray  \end{tabular}
&
\begin{tabular}{c} Breast \\  Lung \end{tabular}
&
\begin{tabular}{c} CBIS-DDSM \\ CheXpert \end{tabular}
& \begin{tabular}{c} Trans combine multi-view info
\end{tabular}
& \begin{tabular}{c} -\\- \end{tabular}

\\

\begin{tabular}{l} \textbf{Verenich \etal} \cite{CNN+Trans-MLL-ATT-verenich2021pulmonary} 
\end{tabular}  
& Chest X-Ray  
& Lung 
& \begin{tabular}{c} COVID-19 \\Radiology Dataset [\cite{dataset19-https://doi.org/10.48550/arxiv.2003.11597,covid19-database}] \end{tabular}    
&  \begin{tabular}{c} Transformer $\times$ CNN 
\end{tabular} 
 & \begin{tabular}{c}94.2 \\ 94.0 \end{tabular} 
\\ 

\begin{tabular}{l}\textbf{Liu \etal} \cite{PureTrans-VOLO-liu2021automatic} 
\end{tabular}  
& COVID-19  
& Lung
& \begin{tabular}{c} Cohen’s dataset \cite{cohen2020covid19} 
\\ COVID-19 database \cite{covid19-database} 
\end{tabular}
&  \begin{tabular}{c} outlooker attention
\end{tabular} 
 & \begin{tabular}{c} 99.0 \\  99.7 \end{tabular}

\\

\begin{tabular}{l} 
\textbf{Shome \etal} \cite{shome2021covid}
\end{tabular} 
& COVID-19 
& Lung 
& -
& \begin{tabular}{c} Grad-CAM-based visualization
\end{tabular} 
 & \begin{tabular}{c} 98.0 \\ 92.0 \end{tabular}

\\

\begin{tabular}{l}
\textbf{Krishnan \etal} \cite{krishnan2021vision}
\end{tabular} 
& COVID-19 
& Lung 
& \begin{tabular}{c}COVID19 X-ray database \\ COVID19, Pneumonia and Normal \\Chest X-ray PA Dataset \end{tabular} 
& \begin{tabular}{c}large-scale COVID19-dataset; \\pretrained ViT-B/32 model \end{tabular}
 & \begin{tabular}{c} 97.6 \end{tabular}

\\

\hline

\textbf{MRI} \\
\cline{1-1}

\begin{tabular}{l}\textbf{He \etal} \end{tabular} \cite{GLTrans-he2021globallocal}
& \textit{Brain Age} 
& Brain 
& \begin{tabular}{c}
BGSP,OASIS-3,NIH-PD,IXI 
\\ ABIDE-I,DLBS,CMI,CoRR 
\end{tabular} 
& \begin{tabular}{c}image-level and patch-level\\ fusion with attention\end{tabular} 
& -
\\

\begin{tabular}{l}\textbf{Kim \etal} \end{tabular} \cite{GraphLearning-DynamicGraph-kim2021learning}
& \begin{tabular}{c} \textit{Gender Classification} \\ \textit{Task Decoding}   \end{tabular}
& \begin{tabular}{c} Brain \\ Brain \end{tabular}
& \begin{tabular}{c} HCP-Rest \\ HCP-Task \end{tabular}
& \begin{tabular}{c}lspatio-temporal attention for \\brain graph representation \end{tabular}  
& \begin{tabular}{c}88.2 \\ 87.0 \end{tabular}

\\

\begin{tabular}{l} \textbf{mfTrans-Net} \end{tabular} \cite{CNN+Trans-mfTrans-zhao2021mftrans}
& \begin{tabular}{c}Hepatocellular 
\\ Carcinoma 
\end{tabular} 
& Hepatic 
&  -  
& \begin{tabular}{c}Trans combine multi-phase\\ info; multi-level learning \end{tabular}
& -

\\

\begin{tabular}{l} \textbf{3DMeT} \end{tabular} \cite{wang20213dmet}
& \begin{tabular}{c}  Knee Cartilage Defect\end{tabular} 
& Knee 
&  -  
& \begin{tabular}{c}Generalize Trans \\on 3D images\end{tabular}
& \begin{tabular}{c}66.4 \\ 70.2
\end{tabular}

\\



\hline

\begin{tabular}{l}\textbf{Histological}\\ \textbf{Image} \end{tabular} \\
\cline{1-1}

\begin{tabular}{l}\textbf{Gao \etal}  \end{tabular} \cite{CNN+Trans-i-ViT-gao2021instance}
& \begin{tabular}{c}Papillary Renal 
\\Cell Carcinoma\end{tabular}  
& Kidney 
& TCGA-KIRP  
& \begin{tabular}{c}instance-based patches; \\positions \& grade encodings \end{tabular}   
& \begin{tabular}{c} 89.2 \\ 93.0 \end{tabular}
\\

\begin{tabular}{l}\textbf{Chen \etal}\end{tabular} \cite{CNN+Trans-GasHis-Transformer-chen2021gashis}
& \begin{tabular}{c}Gastric Histopatho-
\\logical Image \end{tabular}
& Stomach
& HE-GHI-DS 
& \begin{tabular}{c}GIM and LIM modules; \\parallel structure  \end{tabular} 
& 98.0
\\ 

\begin{tabular}{l}\textbf{Zeid \etal}\end{tabular} \cite{zeid2021multiclass}
& \begin{tabular}{c}Colorectal Cancer\end{tabular}
& Colorectal 
& \begin{tabular}{c}Kather \cite{janowczyk2016deep} colorectal \\cancer histology dataset \end{tabular}
& \begin{tabular}{c}GIM and LIM modules; \\parallel structure  \end{tabular} 
& \begin{tabular}{c}93.3 \\ 94.8 \end{tabular}
\\ 

\begin{tabular}{l}\textbf{Ikromjanov \etal}\end{tabular} \cite{ikromjanov2022whole}
& \begin{tabular}{c}Prostate Cancer\end{tabular}
& Prostate 
& \begin{tabular}{c}Kaggle PANDA \\challenge dataset \end{tabular}
& \begin{tabular}{c}Classify according \\to Gleason grading  \end{tabular} 
& -
\\ 

\begin{tabular}{l}\textbf{Zhao \etal}\end{tabular} \cite{zhao2022improving}
& \begin{tabular}{c}cervical cancer\end{tabular}
& Cell 
& \begin{tabular}{c}a.Pap smear dataset \\b.SIPAKMeD c.Herlev \end{tabular}
& \begin{tabular}{c}taming trans × T2T-ViT\end{tabular} 
& -
\\ 

\hline
\begin{tabular}{l}\textbf{Others} \end{tabular} \\
\cline{1-1}

\begin{tabular}{l}\textbf{POCFormer} \end{tabular} \cite{LightTrans-USNet-perera2021pocformer}
& COVID-19 
& Lung 
& POCUS 
& \begin{tabular}{c}a lightweight Trans-based model \end{tabular}
&\begin{tabular}{c} 91.0\\ 95.0\\95.0 \end{tabular}

\\ 

\begin{tabular}{l}\textbf{Gheflati \etal}\end{tabular} \cite{gheflati2021vision}
& Breast Cancer
& Breast
& \begin{tabular}{c}BUSI \cite{BUSI} \\ Dataset B \cite{Dataset-B} \end{tabular}
& \begin{tabular}{c}ViT on breast ultrasound images \end{tabular}
& \begin{tabular}{c} 85.7\\ 86.0 \\86.7\\ 85.0\\ 86.4\end{tabular}
\\

\begin{tabular}{l}\textbf{Jiang \etal} \cite{ViT+CNN+Ensemble-ViT-CNN-jiang2021method}
\end{tabular} 
& 
\begin{tabular}{c}Acute Lymphoblastic 
\\ Leukemia \end{tabular}  
& Lymph 
& ISBI 2019 Dataset
& \begin{tabular}{c} ViT and CNN ensemble\end{tabular} 
& \begin{tabular}{c} 99.0 \end{tabular}
\\

\begin{tabular}{l}\textbf{Xie \etal} \cite{melane-skin-Xie}
\end{tabular} 
& Melanoma   
&Skin
&\begin{tabular}{c} ISIC-2017 \\ Skin Dataset\end{tabular}
& \begin{tabular}{c} SimAM with Swin-Trans\end{tabular}  
& -
\\

\begin{tabular}{l}\textbf{Li \etal} \cite{li2021out}
\end{tabular} 
& Skin Lesion   
& Skin
&\begin{tabular}{c} HAM10000 \\ DermNet\end{tabular}
& \begin{tabular}{c} Trans on Out-of-\\Distribution Detection\end{tabular}  
& -
\\

\begin{tabular}{l}\textbf{Yu \etal} \cite{yu2021end}
\end{tabular} 
& Melanoma   
& Skin
&\begin{tabular}{c} ISIC 2020 dataset \end{tabular}
& \begin{tabular}{c} Transformer $\times$ \\  contrastive learning \end{tabular}  
& -
\\

\begin{tabular}{l}\textbf{Wu \etal} \cite{wu2021scale}
\end{tabular} 
& Melanocytic Lesions   
& Skin
&\begin{tabular}{c} MPATH-Dx \end{tabular}
& \begin{tabular}{c} Encode multi-scale \\features with Trans \end{tabular}  
& 60.0
\\

\begin{tabular}{l}\textbf{TransEye \etal} \cite{yang2021fundus}
\end{tabular} 
& Fundus Disease   
& Eye
&\begin{tabular}{c} OIA \end{tabular}
& \begin{tabular}{c} Trans $\times$ CNN \end{tabular}  
& 84.1
\\

\toprule[1pt]
\end{tabular}\\
\end{table*}

\subsubsection{Applications of pure Transformers}

We call ViTs that are similar to the original proposed one \cite{dosovitskiy2020vit} as {\emph pure Transformers}. These methods usually do not contain significant structural changes. 
We introduce the literature of pure Transformers by image modality, \eg, X-Ray \cite{PureTrans-VOLO-liu2021automatic, tanzi-femur-fracture-x-ray}, computed tomography \cite{SwinTrans-MIA-COV19D-zhangmia, PureTrans-COVID-VIT-gao2021covidvit}, magnetic resonance imaging \cite{GLTrans-he2021globallocal}, ultrasound \cite{LightTrans-USNet-perera2021pocformer}, OCT \cite{Pure-Trans-Eye-song2021deep}, \etc

\noindent\textbf{X-Ray.}
X-Ray is an inexpensive and convenient imaging technique that is widely used in screening and diagnosis of several diseases, \eg, breast cancer, pneumonia, fracture, \etc
Especially during the COVID-19 pandemic, X-ray has played a very important role in the disease screening, and thus is a popular modality for AI researchers to use when designing Transformer-based methods. 
Liu \etal \cite{PureTrans-VOLO-liu2021automatic} proposed the Vision Outlooker (VOLO), a ViT model that replaced the original attention mechanism with the outlooker attention \cite{yuan2021volo}.
Their model achieved state-of-the-art (SOTA) performance for the diagnosis of COVID-19 without pretraining on ImageNet. 
Shome \etal \cite{shome2021covid} proposed a ViT based model for COVID-19 diagnosis by training the model on a self-collected large COVID-19 chest X-ray image dataset. 
They also used Grad-CAM \cite{selvaraju2017grad} to show the progression of COVID-19.
Krishnan \etal \cite{krishnan2021vision} applied an ImageNet-pretrained ViT-B/32 network to distinguish COVID-19 by using the patches from chest X-ray images as inputs.
Despite the effectiveness of ViTs for COVID-19, Tanzi \etal \cite{tanzi-femur-fracture-x-ray} applied a ViT model to classify femur fracture. Their work utilized clustering methods to validate the ViT ability in extracting features, and compared its performance against CNNs.
The aforementioned models reveal the importance of large data-scale datasets, which enhance the performance of Transformers. Therefore, since the scale of the dataset for COVID-19-related task \cite{PureTrans-VOLO-liu2021automatic, shome2021covid, selvaraju2017grad, krishnan2021vision} is larger than the femur fracture task \cite{tanzi-femur-fracture-x-ray}, the performance of the COVID-19-related task is also higher.

\noindent\textbf{Computed Tomography.}
Based on the high contrast between gas and tissue, CT is commonly used for thoracic disease diagnosis. Thus, the application of pure Transformers to CT images mainly focused on thoracic diseases.
For example, Than \etal \cite{than2021preliminary} studied the patch size of ViT for COVID-19 and Diseased Lungs Classification task. They found that the performance dropped with larger patch sizes, revealing a trade off between local and global information while the $32\times32$ patch resulted in the best accuracy. 
Costa \etal \cite{PureTrans-BrazilViT-costa2021covid} proposed to use ViT and its variants to distinguish COVID-19 Pneumonia and other Pneumonia from normal cases. By comparing the performance of several models, they showed that pretrained models such as Data-efficient image Transformer (DeiT) \cite{Data-Efficient-Deit} achieved competitive results. Meanwhile, the conventional ViT as well as its variants using Performer Encoder also achieved good results even without pretraining.
Li \etal \cite{li2021medical} designed a platform for COVID-19 diagnosis based on ViT. They converted the CT images into a series of flattened patches to fit the input of ViT for diagnosis.
They also adopted a teacher-student model to distill knowledge from a CNN pretrained on natural images.
Gao \etal \cite{PureTrans-COVID-VIT-gao2021covidvit} applied the ViT on both 2D and 3D CT scans to diagnose COVID-19. They proposed to construct an image sub-volume by extracting a fixed number of slices thereby 'normalizing' imaging sequences with a varying number of slices. 
They also proved that the ViT performance is better than that of DenseNet, which is a competitive CNN model.
Zhang \etal \cite{SwinTrans-MIA-COV19D-zhangmia} trained the popular Swin-Transformer on CT images. Specifically, the framework firstly segments the lung via an Unet, and then feeds the lung region to the feature extractor. Such strategy helped reduce the computation burden of the Transformer framework. 
The aforementioned works showed the importance of pre-training for CT image classification tasks since CT images are much harder to acquire than X-Rays. Also, the methods that reduce the computation complexity by attention mechanism are crucial to CT images, due to their large volume sizes.

\noindent\textbf{Magnetic Resonance Imaging.}
The Magnetic Resonance Imaging (MRI) has a better imaging quality particular for subtle anatomical structures including vessels and nerves, however it is time-consuming in acquisition. As MRI represents a powerful non-invasive imaging technology of soft-tissues, it is commonly used in neuroimaging studies. 
For instance, He \etal \cite{GLTrans-he2021globallocal} proposed a two-pathway network for brain age estimation. A global-pathway is designed to capture the global contextual information from the brain MRI, while a local-pathway is responsible for capturing the fine-grained information from local patches. The local and global contextual representations are then fused by a global-local attention mechanism. Next, the concatenation of fused features and local patches are fed into a revised global-local Transformer. Also, MRI has a wide spectrum of clinical application, \eg, cancer diagnosis, which makes it a strong candidate modality for training ViTs.

\noindent\textbf{Ultrasound.}
Ultrasound with Point-of-Care (POC) has expanded applicable scenarios, as specific positions are not necessary for acquiring the images. 
The work from Perera \etal \cite{LightTrans-USNet-perera2021pocformer} proposed a Transformer-based architecture to diagnose COVID-19 based on ultrasound clips. To ensure memory and time efficiencies, they proposed to replace the standard ViTs with Linformer, reducing the the space time complexity from $O(n^2)$ for conventional self-attention mechanism to $O(n)$.
Moreover, ultrasound has also became a prominent modality for breast cancer imaging thanks to its ease-of-use, low-cost and safety. Gheflati \etal \cite{gheflati2021vision} used ViTs to classify normal, malignant and benign breast tissues using ultrasound images. 
They also compared the performance of ViTs with various configurations against CNNs to demonstrate their efficiency. 

\noindent\textbf{Others.}
In addition to the abovementioned imaging modalities, other imaging technologies are adopted for the examination and diagnosis of specific diseases, \eg, dermoscopy images \cite{melane-skin-Xie}, fundus images \cite{aldahoul2021encoding}, histopathology images \cite{ikromjanov2022whole}. For instance, Xie \etal \cite{melane-skin-Xie} aims to detect melanoma using dermpscopy images. They proposed to combine the Swin-Transformer with a parameter-free attention module SimAM to learn better features for the target classification task. Considering that the features fed into the classifier contain rich semantic information but lack detailed information, they designed the output of the first three Swin-Transformer Blocks as three SimAM blocks input separately, and then all SimAM block outputs including the final feature map are concatenated together to form the new final feature map, which serves as the input to the final classification layer.
Li \etal \cite{li2021out} evaluated the performance of Transformers on the Out-of-Distribution detection tasks in medical image analysis. The original Vision Transformer and the Data-efficient image Transformer with multi-head, soft-distillation and hard distillation are included in their work. The performance of these models on skin lesion datasets HAM10000 and DermNet showed the limited performance and safety-critical problem of the Transformers for the OOD detection task.\
Ikromjanov \etal \cite{ikromjanov2022whole} applied ViT to assist pathologists to grade Prostate cancer (PCa) according to the Gleason grading system on whole slide histopathology images and showed promising results.

As shown in Table \ref{Table:classification}, despite the excellent performance of pure Transformers in several cases, \eg, COVID-19 X-Rays, further development is also necessary for other tasks.

\subsubsection{Applications of hybrid Transformers}

Although pure ViTs can achieve promising results without much modification, extensive efforts have been put into the exploration of combining ViTs with other learning components, to better capture complex data distributions or gain better performance. Typical cases are combinations of Transformers with 1) convolutional layers and 2) graph representations. We next introduce both categories.

\noindent\textbf{Transformers with convolutions.}
Vision Transformers focus more on modeling the global relationship within the data, while conventional CNNs pays more attention to the local texture. Such difference inspired researchers to combine the advantages of ViTs and CNNs. Also, the analysis of medical images involves not only the correlation of regions in the image, but also subtle textures. Hence, many works were devote to explore such CNN-ViT combination.

Most applications focused on the diagnosis of thoracic diseases, especially COVID-19 or other related diseases.
Benefiting from ViT's power of feature integration, Van \etal \cite{CNN+Trans-Multi-View-van2021multi} utilized a Transformer to conduct multi-view analysis of unregistered medical images, to classify chest X-rays. They proposed the Transformer-based approach to consider spatial information across different views at the feature-level by virtue of the trainable attention mechanism. 
They applied the Transformer to the intermediate feature maps produced by CNNs to retrieve features from one view, and transfer them to another view. 
Thus, additional context was added to the original view without requiring pixel-wise correspondences. 
Their work also contributed to the reduction of computational complexity by proposing to substitute a smaller number of visual tokens for the source pixels.
Verenich \etal \cite{CNN+Trans-MLL-ATT-verenich2021pulmonary} introduced global spatial information from ViTs to CNNs for pulmonary disease classification, while preserving spatial invariance and equivariance.
Liang \etal \cite{CNN+Trans-TransHybrid-liang2021hybrid} utilized a CNN to mine effective features, and a Transformer to conduct feature aggregation.
Additionally, an effective data sampling strategy is adopted to reduce the size of the inputs while preserving sufficient information for diagnosis. 
Park \etal \cite{Pre-training-COVID-19-park2021vision} designed a pretrained CNN backbone followed by a ViT for COVID-19 diagnosis. A large-scale public dataset for CXR classification is used in model pretraining. 
For the simple task of classifying thoracic diseases, existing methods are simple yet effective, where a CNN is used to extract the features, then a Transformer is used to capture high-level information.

\begin{figure}[htbp]
    \centering
    \includegraphics[width=\linewidth]{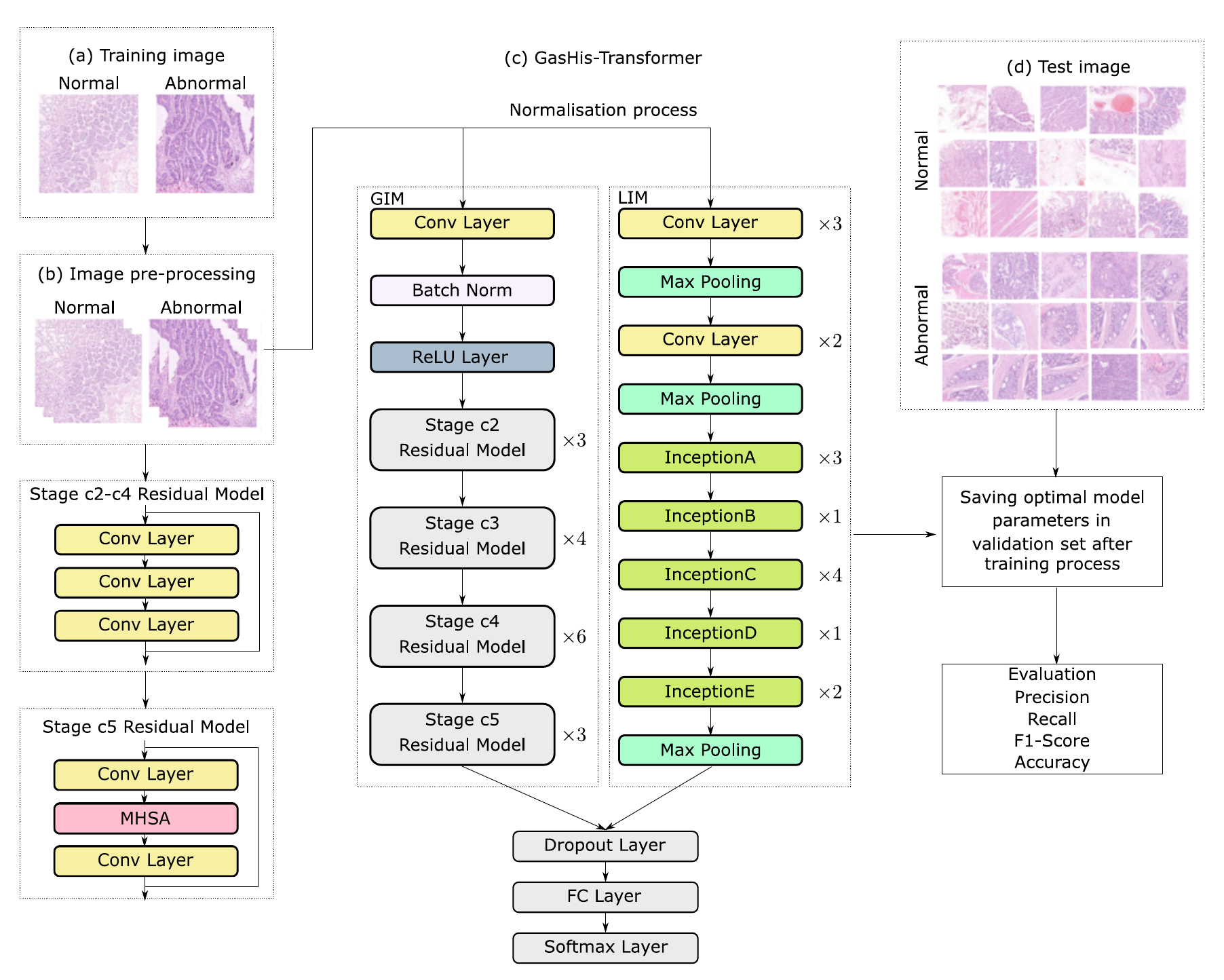}
    \caption{The structure of GasHis-Transformer model \cite{CNN+Trans-GasHis-Transformer-chen2021gashis}.}
    \label{fig:GasHisTrans}
\end{figure}

Other than COVID-19 diagnosis, 
Yassine \etal \cite{CNN+Trans-Scopeformer-barhoumi2021scopeformer} combined several CNNs with the ViT by feeding the extracted features into a ViT. 
They compared the number of CNNs as well as their pretraining configurations against the hybrid CNN-ViT model. 
It is worth mentioning that they pretrained the CNN on images generated from ImageNet dataset \cite{krizhevsky2012imagenet}, using a Generative adversarial network (GAN) pretrained on brain CT images. They claimed that further pretraining on the generated images could lead to a better inductive bias for the target computed tomography dataset as the dissimilarities of two domains are reduced.
Zhao \etal \cite{CNN+Trans-mfTrans-zhao2021mftrans} considered utilizing the combination of CNNs and Transformers to conduct multi-index quantification of hepatocellular carcinoma (HCC) using multi-phase contrast-enhanced magnetic resonance imaging (CEMRI). They proposed the mrTrans-Net where three parallel encoders, each followed by a non-local Transformer that extracts features from Arterial phase, PV phase and Delay phase. Next, a phase-aware Transformer is added to quantify the relevance of each phase for the target multi-phase CERMI information fusion and selection. The quantification is conducted not only after the phase-aware Transformer but also those non-local Transformers to form an enhanced loss function to constrain the quantification task.
Jiang \etal \cite{ViT+CNN+Ensemble-ViT-CNN-jiang2021method} explored the effectiveness of ensemble learning by treating ViTs and CNNs as base learners to diagnose Acute Lymphoblastic Leukemia based on microscopic images of B-lymphoid precursors and leukemic B-lymphoblast cells. They proposed an ensemble model based on the ViT and EfficientNet. As the two base models are complementary, the ensemble results showed some improvement. 
They also proposed a data enhancement method to handle normal/cancer cell imbalance in each image.
Chen \etal \cite{CNN+Trans-GasHis-Transformer-chen2021gashis} proposed the multi-scale vision Transformer model, shown in Fig.\ref{fig:GasHisTrans}, called GasHis-Transformer, to deal with gastric histopathological image classification. They designed the Global Information Module (GIM) and Local Information Module (LIM) (which is based on CNNs) to extract features. Moreover, they borrowed the parallel structure from the Inception-V3 to learn multi-scale local representations. Additionally, their model was robust to ten different adversarial attacks or conventional noises, and was generalizable to other cancer histopathological image classification tasks. 
Gao \etal \cite{CNN+Trans-i-ViT-gao2021instance} proposed the instance-based Vison Transformer (i-ViT) for papillary renal cell carcinoma subtyping. The i-ViT first extracts and selects instance features from instance-level patches which include a nucleus with parts of the surrounding background and the nuclei grade. Next, it aggregates these features for further capturing the cellular level and cell-layer level features. At last, the model encodes both of the fine-gained features into the final image-level representation, where grades and positions are embedded for subtyping.
Wang \etal \cite{wang20213dmet} proposed a 3D Transformer to outperform 3D CNNs. They used the 3D convolutional layer to extract features of 3D blocks, and the teacher-student network to learn Transformer weights from a CNN teacher.
Xia \etal \cite{xia2021effective} proposed the Anatomy-Aware Transformers for pancreatic cancer screening, and showed to win the radiologists.
Zeid \etal \cite{zeid2021multiclass} validated ViTs and its variants Compact Convolutional Transformer on the multiclass colorectal cancer (CRC) histology image classification task on a public CRC histology dataset.
Zhao \etal \cite{zhao2022improving} combined taming Transformers with T2T-ViT to handle the unbalanced samples with inconsistent image quality for cervical cancer classification task. 
Yu \etal \cite{yu2021end} adopted the Transformer encoder to model dependency among features of skin lesion to detect ugly duckling sign for melanoma identification. 
Yang \etal \cite{yang2021fundus} proposed a Transformer Eye (TransEye) fine-grained fundus disease image classification by jointly adopting the CNN and Transformer model.
Wu \etal \cite{wu2021scale} proposed the ScATNet to model the inter-patch and inter-scale representations at multiple input scales to diagnose melanocytic lesions in biopsy images. 
These hybrid Transformers for various applications contain rich innovations, including structure improvements, novel designed ViT modules, CNN modules, and also the learning strategies of pretraining and ensembling.

\noindent\textbf{Transformers with graphs.}
Learning with graphs is a common practice in MIA. The core concept of graph learning is to learn a compact representation of each sample (e.g., embeddings) while preserving the intrinsic inter-sample relationships via the data graph \cite{Graph-Representation-Learning-Survey}. As an attention-based network, Transformer is suitable to operate on graph data, including aggregating the node features and calculating the node relationships.

In the field of network neuroscience, the brain network is modeled as a graph where each node denotes an anatomical region of interest (ROI) and the edge connecting two nodes encodes their interaction (e.g., neural firing). Brain graphs play an important role in advancing our understanding of the brain as highly interconnected system in both health and disease \cite{Brain-Network-Analysis-liu2017complex,Bessadok2021}.
The work from Kim \etal \cite{GraphLearning-DynamicGraph-kim2021learning} leveraged the dynamic characteristics of the functional connectivity (FC) network by incorporating dynamic features into a compact brain graph representation. Specifically, they proposed the Spatio-Temporal Attention Graph Isomorphism Network (STAGIN) for learning the dynamic graph representation of the brain connectome with spatio-temporal attention.
The GNN is used to extract graph-level representations for the functional brain connectome at each timestep. 
At last, the model uses the Transformer encoder to obtain the final representation of a sequence of dynamic graphs. In detail, they concatenated encoded timestamp with node features to embed temporal information.
They claimed that the use of the Transformer not only improved the classification performance, but also improved the spatial-temporal interpretability. 
Such methods have validated the power of Transformers in mining both features and relationships for complex graphs, which draws more attention to this methodology.

We conclude the Transformers for medical image classification tasks as follows:
\begin{itemize}
    \item Transformers have achieved comparable or better performance in most tasks compared with CNNs.
    \item Transformers are hungry for large-scale datasets, which somewhat limits their applicability, especially in the medical image analysis field. Pretraining could be an alternative solution to alleviate this problem.
    \item The computation burden is high when training Transformers on large images. Hence, reducing the model complexity and developing light-weight models are key factors to improve their efficiency.
    \item Hybrid Transformers gained increasing attention as they have benefited from both sides of conventional networks (\ie, CNNs and GNNs) and Transformers.
\end{itemize}

\subsection{Segmentation}

{
\begin{table*}[htbp]
\renewcommand{\arraystretch}{0.5}
\centering
\caption{\label{Table:segmentation} Transformers for medical image segmentation tasks. Abbreviation: ACT-MOS, abdominal CT multi-organ segmentation; COPLE, COVID-19 Pneumonia Lesion segmentation; ACDC, Automated cardiac diagnosis challenge; MSD-01, Medical Segmentation Decathlon dataset Task01; KCCEE, Kvasir \cite{jha2020kvasir}, CVC-ClinicDB \cite{bernal2015wm}, CVC-ColonDB \cite{tajbakhsh2015automated}, EndoScene \cite{vazquez2017benchmark} and ETIS \cite{silva2014toward}. The performance is reported in terms of Dice (\%).}
\setlength{\tabcolsep}{1pt}
\begin{tabular}{cccccc}
\toprule[1pt]
\textbf{Reference} & \textbf{Task} & \textbf{DataSet} &\textbf{Performance}  &  \textbf{Highlight} \\
\hline

\begin{tabular}{c} TransUnet,\\ Chen \etal \cite{chen2021transunet} \end{tabular}
& \begin{tabular}{c} a. ACT-MOS \\ b. cardiac segmentation \end{tabular}
& \begin{tabular}{c} a. Synapse multi-organ CT\\ b. ACDC \end{tabular}
& \begin{tabular}{c} a. 77.48 \\b. 89.71\end{tabular}
&  \\

\begin{tabular}{c}TransClaw,\\ Yao \etal \cite{chang2021transclaw}\end{tabular}
& ACT-MOS
& \ Synapse multi-organ CT
& \ 78.09
& Claw Unet \\

\begin{tabular}{c} LeViT-Unet,\\ Xu \etal \cite{xu2021levit} \end{tabular}
& \begin{tabular}{c} a. ACT-MOS \\b. cardiac segmentation\end{tabular}
& \begin{tabular}{c} a. Synapse multi-organ CT \\b. ACDC\end{tabular}
& \begin{tabular}{c} a. 78.53 \\b. 90.32\end{tabular}
& LeViT \\

\begin{tabular}{c} Tunet,\\ Sha \etal \cite{sha2021Transformer}\end{tabular}
& \ pancreas segmentation
& \ CT82 datasets
& \ 79.66
&  \\

 Li \etal \cite{li2021more}
& \begin{tabular}{c} a. brain tumour segmentation \\b. ACT-MOS \end{tabular}
& \begin{tabular}{c} a.  MSD-01 \cite{simpson2019large} \\b. Synapse multi-organ CT\end{tabular}
& \begin{tabular}{c} a. 80.30\\ b. 74.75\end{tabular}
& \\

\begin{tabular}{c} UTNet,\\ Gao \etal \cite{gao2021utnet}\end{tabular}
& \begin{tabular}{c} cardiac segmentation\end{tabular}
& \ M\&Ms \cite{campello2021multi}
& \ 88.3
&  \\

\begin{tabular}{c}TransBTSV2,\\ Fu \etal \cite{fu2022tf}\end{tabular}	
& ab. brain tumor segmentation	
& \begin{tabular}{c} a. BraTS 2019 dataset \\ b. BraTS 2020 \\ c. LiTS2017 dataset (lesion, liver) \\d. KiTS2019 dataset (kidney, tumor) \end{tabular}
& \begin{tabular}{c} a. 85.18 \\b. 84.90 \\c. 71.20, 96.20  \\d. 97.37, 83.69 \end{tabular}
& 
\\

\begin{tabular}{c} UTNetV2, \\ Gao \etal \cite{gao2022multi}\end{tabular}
& cardiac segmentation	
& ACDC	
& 92.14
& 
\\

\hline

\begin{tabular}{c} HybridCTrm,\\ Sun \etal \cite{sun2021hybridctrm}\end{tabular}
& Brain tissue segmentation
& \begin{tabular}{c}\ a. MRBrainS \cite{mendrik2015mrbrains} \\b. iSeg-2017 \cite{wang2019benchmark}\end{tabular}
& \ a. 83.47 b. 87.16
& Dual-Path Network \\

\begin{tabular}{c} TransFuse,\\ Zhang \etal \cite{zhang2021transfuse}\end{tabular}
& \begin{tabular}{c} a. polyp segmentation \\b. skin lesion segmentation\\ c. hip segmentation \\d. prostate segmentation\end{tabular}
& \begin{tabular}{c} a. KCCEE  \\b. ISIC2017 \cite{codella2018skin} \\c. in-house dataset \\d. MSD dataset \end{tabular}
& \begin{tabular}{c} a. 92.0, 94.2, 78.1,\\ 89.4, 73.7 \\b. 87.2 c. - d. 76.4\end{tabular}
& Multi-level feature fusion \\

\begin{tabular}{c} CA-GANformer,\\ You \etal \cite{you2022class}\end{tabular}
& \begin{tabular}{c}a. ACT-MOS\\
b. liver tumor segmentation\end{tabular}
& \begin{tabular}{c}a. Synapse multi-organ CT\\
b. LiTS dataset\end{tabular}
& \ a. 82.55 b. 73.82
&  GAN\\

\begin{tabular}{c} ECT-NAS,\\ Xu \etal \cite{xu2021ect}\end{tabular}
& \begin{tabular}{c}a. ACT-MOS\\
b. cardiac segmentation\end{tabular}
& \begin{tabular}{c}a. Synapse multi-organ CT\\
b. ACDC\end{tabular}
& \ a. 78.97 b. 89.83
&  Searching strategy\\

\begin{tabular}{c} HyLT, \\Luo \etal \cite{luo2021hybrid}\end{tabular}
& \begin{tabular}{c}a. gland segmentation\\
b. nuclear segmentation\end{tabular}
& \begin{tabular}{c}a. GlaS dataset\\
b. MoNuSeg dataset\end{tabular}
& \ a. 90.86 b. 80.25
&  Multi-level feature fusion\\

\begin{tabular}{c} PHTrans, \\Liu \etal \cite{liu2022phtrans}\end{tabular}
& \begin{tabular}{c}a. ACT-MOS\\
b. cardiac segmentation\end{tabular}
& \begin{tabular}{c}a. BCV dataset\\
b. ACDC\end{tabular}
& \ a. 88.55 b. 91.79
&  Multi-level feature fusion\\

\begin{tabular}{c} nnFormer,\\ Zhou \etal \cite{zhou2021nnformer}\end{tabular}
& \begin{tabular}{c} a. ACT-MOS \\b. cardiac segmentation\end{tabular}
& \begin{tabular}{c} a. Synapse multi-organ CT \\b. ACDC\end{tabular}
& \ a. 87.40 b. 91.78
&  \\
\hline



\begin{tabular}{c} PMTrans,\\ Zhang \etal \cite{zhang2021pyramid}\end{tabular}
& \begin{tabular}{c} a. gland segmentation \\b. nuclear segmentation\end{tabular}
& \begin{tabular}{c} a. GlaS dataset \cite{sirinukunwattana2017gland} \\b. MoNuSeg dataset \cite{kumar2017dataset}\end{tabular}
& \ a. 81.48 b. 80.09
& Multi-resolution images  \\

\begin{tabular}{c} MedT,\\ Valanarasu \etal \cite{valanarasu2021medical}\end{tabular}
& \begin{tabular}{c} a. brain anatomy segmentation \\b. gland segmentation \\c. nucleus segmentation\end{tabular}
& \begin{tabular}{c} a. Brain US dataset \cite{valanarasu2020learning} \\b. GlaS dataset \\c. MoNuSeg dataset\end{tabular}
& -
& Global $+$ local \\
\begin{tabular}{c} CoTr,\\ Xie \etal \cite{xie2021cotr}\end{tabular}
& \ ACT-MOS
& \ BCV dataset \cite{landman20152015}
& \ 85.0
& Multi-scale features  \\

\begin{tabular}{c} MCTrans,\\ Ji \etal \cite{ji2021multi}\end{tabular}
& \begin{tabular}{c} a. cell segmentation \\b. polyp segmentation \\c. skin lesion segmentation\end{tabular}
& \begin{tabular}{c} a. Pannuke \cite{gamper2019pannuke} \\b. KCCEE \\c. ISIC2018 \cite{codella2019skin}\end{tabular}
& \begin{tabular}{c} a. 68.40\\ b. 92.30, 86.58,\\ 83.69, 86.20\\ c. 90.35\end{tabular}
& Multi-scale features \\

\begin{tabular}{c} D-Former,\\ Wu \etal \cite{wu2022d}\end{tabular}
& \begin{tabular}{c} a. ACT-MOS\\
b. cardiac segmentation\end{tabular}
& \begin{tabular}{c} a. Synapse multi-organ CT\\
b. ACDC\end{tabular}
& \ a. 88.83 b. 92.29
& Dilated Transformer  \\

\begin{tabular}{c} TF-Unet,\\ Fu \etal \cite{fu2022tf}\end{tabular}
& \begin{tabular}{c} a. ACT-MOS\\
b. cardiac segmentation\end{tabular}
& \begin{tabular}{c} a. Synapse multi-organ CT\\
b. ACDC\end{tabular}
& \ a. 85.46 b. 91.72
& Multi-scale features  \\

\begin{tabular}{c} UNETR,\\ Hatamizadeh \etal \cite{hatamizadeh2021unetr}\end{tabular}
& \begin{tabular}{c} a. brain tumour segmentation \\b. spleen CT segmentation\end{tabular}
& \begin{tabular}{c} a. MSD-01 \\b. MSD dataset Task09\end{tabular}
& \ a. 71.81 b. 95.82
& Multi-resolution  \\

\begin{tabular}{c} Swin UNETR,\\ Hatamizadeh \etal \cite{hatamizadeh2022swin}\end{tabular}
& \begin{tabular}{c} brain tumor segmentation\end{tabular}
& \begin{tabular}{c} BraTS2021 dataset\end{tabular}
& \ 91.3
& Swin transformer  \\
\hline

\begin{tabular}{c} TransAttUnet,\\ Chen \etal \cite{chen2021transattunet}\end{tabular}
& \begin{tabular}{c} a. skin lesion segmentation \\b. lung field segmentation \\c. COPLE \\d. nucleus segmentation \\e. gland Segmentation\end{tabular}
& \begin{tabular}{c} a. ISIC2018 dataset \\b. JSRT, Montgomery and NIH \cite{tang2019xlsor} \\c. Clean-CC-CCII dataset \cite{he2020benchmarking} \\d. Bowl dataset \cite{caicedo2019nucleus} \\e. GlaS dataset \end{tabular}
& \begin{tabular}{c} a. - b. 98.88 c. 86.57 \\d. 91.62 e. 89.11\end{tabular}
& Feature fusion  \\

\begin{tabular}{c} MT-Unet,\\ Wang \etal \cite{wang2021mixed}\end{tabular}
& \begin{tabular}{c} a. ACT-MOS \\b. cardiac segmentation\end{tabular}
& \begin{tabular}{c} a. Synapse multi-organ CT \\b. ACDC\end{tabular}
& \ a. 78.59 b. 90.43
&   \\

\begin{tabular}{c} AFTer-Unet,\\ Yan \etal \cite{yan2021after}\end{tabular}
& \begin{tabular}{c} a. ACT-MOS \\b. thoracic segmentation\end{tabular}
& \begin{tabular}{c} a. BCV dataset \\b. Thorax-85 \cite{chen2021deep},\\ SegTHOR \cite{lambert2020segthor}\end{tabular}
& \begin{tabular}{c} a. 81.02 \\b. 92.32, 92.10 \end{tabular}
& \ Axial fusion \\

\begin{tabular}{c} S²WinTOUnet,\\ Zhang \etal \cite{zhang3992963s2wintounet}\end{tabular}
& \begin{tabular}{c}
skin lesion  segmentation\end{tabular}
& \begin{tabular}{c} ISIC2018 dataset\end{tabular}
& \begin{tabular}{c} 90.4 \end{tabular}
& \begin{tabular}{c} star-shaped Window\\ self-attention\end{tabular}
 \\
\hline

Karimi \etal \cite{karimi2021convolution}
& \begin{tabular}{c} a. brain cortical plate segmentation \\b. pancreas segmentation \\c. hippocampus segmentation\end{tabular}
& -
& \ a. 87.9 b. 82.6 c. 88.1
&  \\

\begin{tabular}{c} Swin-Unet,\\ Cao \etal \cite{cao2021swin}\end{tabular}
& \begin{tabular}{c} a. ACT-MOS \\b. cardiac segmentation\end{tabular}
& \begin{tabular}{c} a. Synapse multi-organ CT \\b. ACDC\end{tabular}
& \ a. 79.13 b. 90.00
& Swin-Transformer  \\

\begin{tabular}{c} DS-TransUNet,\\ Lin \etal \cite{lin2021ds}\end{tabular}
& \begin{tabular}{c} a. polyp segmentation \\b. skin lesion segmentation \\c. gland segmentation \\d. nucleus segmentation\end{tabular}
& \begin{tabular}{c} a. Kvasir, CVC-ColonDB, EndoScene, \\ETIS, CVC-ClinicDB b. ISIC2018 \\c. GlaS Dataset d. Bowl dataset\end{tabular}
& \begin{tabular}{c} a. 93.5, 79.8, 91.1,\\ 77.2, 93.8 \\b. - c. 87.19 d. - \end{tabular}
& Swin-Transformer \\

\begin{tabular}{c} MISSFormer,\\ Huang \etal \cite{huang2021missformer}\end{tabular}
& \begin{tabular}{c} a. ACT-MOS \\b. cardiac segmentation\end{tabular}
& \begin{tabular}{c} a. Synapse multi-organ CT \\b. ACDC\end{tabular}
& \ a. 81.96 b. 87.90
& Multi-scale feature \\
\toprule[1pt]
\end{tabular}\\
\end{table*}
}

Transformer-based methods have also been applied to a variety of segmentation tasks, including abdominal multi-organ segmentation \cite{chen2021transunet, chang2021transclaw, xu2021levit, li2021more, zhou2021nnformer, xie2021cotr, wang2021mixed, yan2021after, cao2021swin, huang2021missformer, you2022class, wu2022d, xu2021ect, fu2022tf, liu2022phtrans}, thoracic multi-organ segmentation \cite{yan2021after}, cardiac segmentation \cite{chen2021transunet, xu2021levit, gao2021utnet, gao2022multi, zhou2021nnformer, wang2021mixed, cao2021swin, huang2021missformer, ning2021cac, wu2022d, xu2021ect, fu2022tf, liu2022phtrans}, pancreas segmentation \cite{sha2021Transformer, karimi2021convolution}, brain tumour/tissue segmentation \cite{li2021more, sun2021hybridctrm, valanarasu2021medical, hatamizadeh2021unetr, karimi2021convolution, wang2021transbts, Self-Pre-training-BrainEncoder-jun2021medical, ranem2022continual, laiton2022deep, rao2022improving, zhang3992963s2wintounet, hatamizadeh2022swin, fu2022tf, liang2022transconver, hatamizadeh2022unetformer}, polyp segmentation \cite{zhang2021transfuse, ji2021multi, lin2021ds, wang2022medical, huang2021transde}, liver and hepatic lesion segmentation \cite{you2022class, hille2022joint, wang2021multi, li2022rdctrans, zhang3992963s2wintounet, fu2022tf, wang2022multiscale}, kidney tumor segmentation \cite{wang2021multi, fu2022tf}, skin lesion segmentation \cite{zhang2021transfuse, ji2021multi, chen2021transattunet, lin2021ds, wang2021boundary, zhang3992963s2wintounet, huang2021transde} , prostate segmentation \cite{zhang2021transfuse, wang2022multiscale}, gland segmentation \cite{zhang2021pyramid, valanarasu2021medical, lin2021ds, chen2021transattunet, luo2021hybrid}, nucleus segmentation \cite{zhang2021pyramid, valanarasu2021medical, chen2021transattunet, lin2021ds, luo2021hybrid}, cell segmentation \cite{ji2021multi, zhang2021multi, prangemeier2020attention}, spleen segmentation \cite{hatamizadeh2021unetr}, lung field/COVID-19 pneumonia lesion segmentation \cite{chen2021transattunet}, retinal vessel segmentation \cite{chen2022pcat}, and hyperspectral pathology image segmentation \cite{yun2021spectr}. Several remarkable methods are listed and detailed in Table \ref{Table:segmentation}.

In most medical image segmentation tasks, the U-shaped convolutional neural network architecture, known as Unet, has achieved tremendous success. However, due to the use of convolution operations, Unet is also limited in modeling long-term dependencies. To overcome this limitation, researchers have made efforts in designing robust hybrid Transformers combined with the Unet architecture, which will be introduced in the first part. Besides, several methods also apply pure Transformers to the segmentation tasks, and will be introduced in the second part of this section.

\subsubsection{Hybrid Transformers}

To build Transformers coupled with the popular U-shaped architecture, we find that existing researches pay most attention to the following three questions including:

1) inserting Transformer layers at different levels of the U-shaped architecture;

2) combining Transformer and CNN using different strategies;

3) using multi-scale features or attention mechanisms. 

We detail below each of these three categories.

\begin{figure}[htbp]
    \centering
    \includegraphics[width=\linewidth]{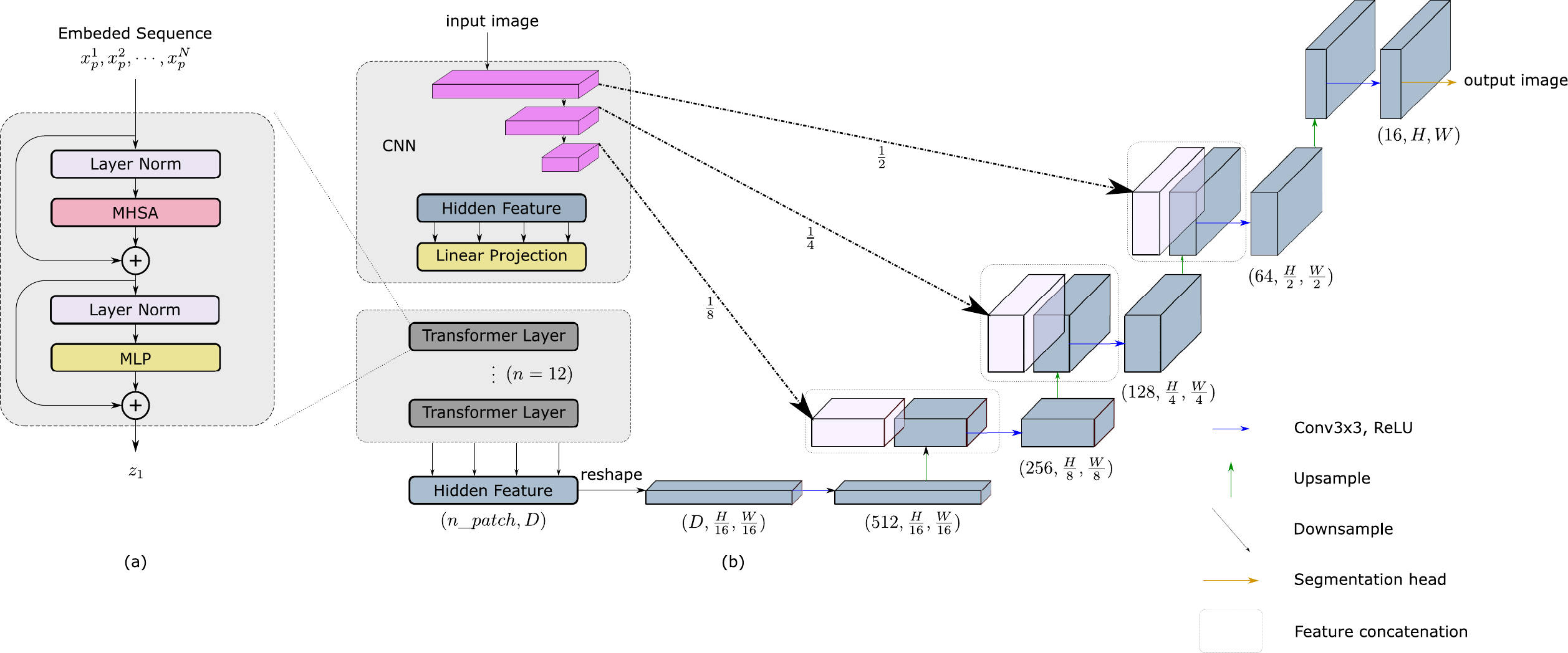}
    \caption{Overview of TransUNet for medical image segmentation. (a) Schematic design of the Transformer layer; (b) Architecture of TransUNet \cite{chen2021transunet}. Transformer layers are inserted into the encoder of the Unet. 
    \label{fig:TransUNet}} 
\end{figure}

\begin{figure}[htbp]
    \centering
    \includegraphics[width=\linewidth]{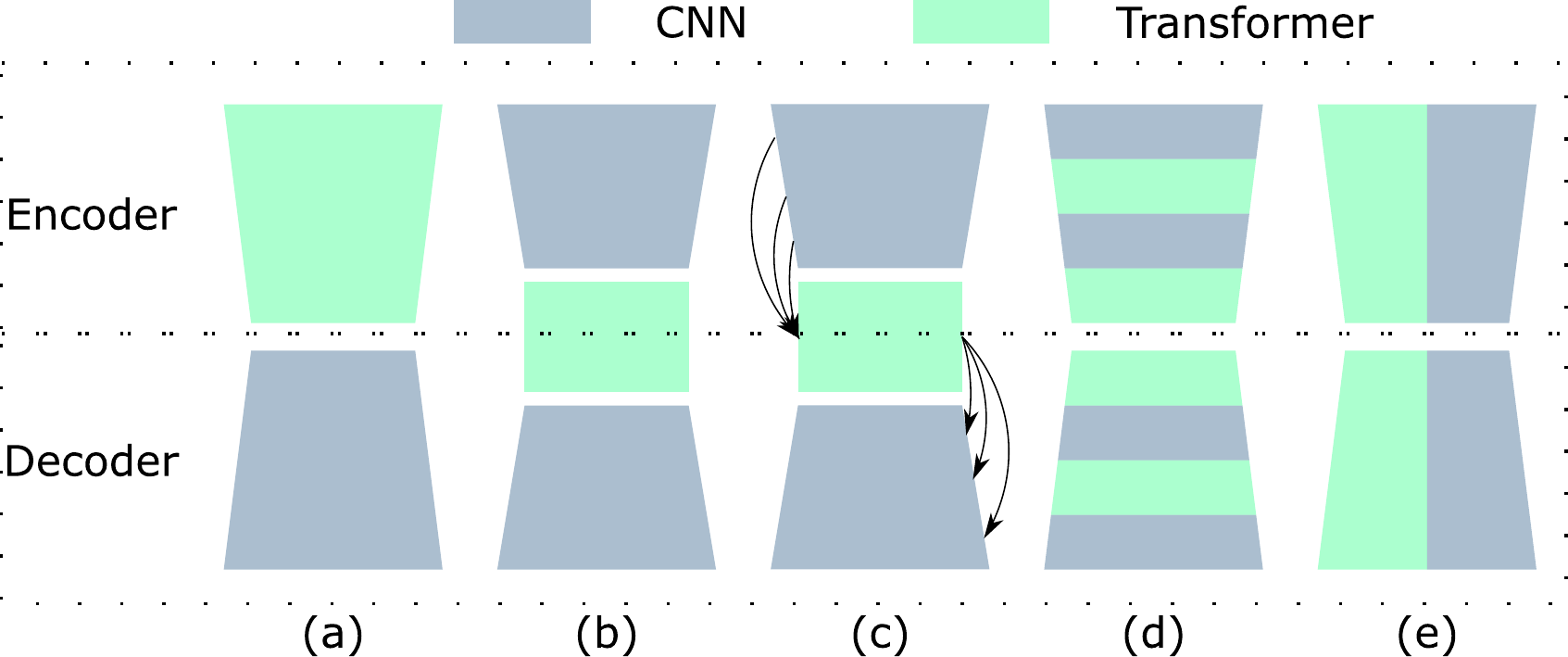}
    \caption{Comparison of several hybrid architectures of Transformer and CNN \cite{liu2022phtrans}.
    \label{fig:Hybrid architectures}} 
\end{figure}

\noindent\textbf{Location of Transformer in U-shaped architecture.}

To insert Transformer layers into the U-shaped architecture, an intuitive idea is to insert a whole Transformer between the encoder and decoder blocks, to build the long-term dependencies in between high-level vision concepts. Following this idea, Chen \etal \cite{chen2021transunet} proposed TransUNet, shown in Fig. \ref{fig:TransUNet}, which extracts high-resolution spatial features by CNN, and then encodes the global context by a Transformer. The self-attentive features encoded by the Transformer are then upsampled and combined with features at multiple scales extracted from the encoding path using skip-connections for precise localization. TransUNet achieved superior performance compared with V-Net, AttnUNet, and ViT in multi-organ and cardiac segmentation tasks. 
Similar to TransUNet, Yao \etal \cite{chang2021transclaw} combined a Transformer network with a Claw Unet architecture and outperformed TransUnet in synapse multi-organ segmentation. For another instance, Xu \etal \cite{xu2021levit} proposed LeViT-UNet, which integrates a LeViT Transformer into the Unet architecture. In \cite{sha2021Transformer}, Sha \etal designed a Transformer-Unet by adding Transformer modules to Unet, which outperformed TransUnet.

Different from the above studies that inserts the Transformer right after the encoder block, Li \etal \cite{li2021more} added an Attention Upsample (AU) component to the decoder. They also proposed the Window Attention Decoder (WAD) and Window Attention Upsampling (WAU), working on local windows, to reduce memory and computation costs. Gao \etal \cite{gao2021utnet} presented a UTNet that applies self-attention modules in both encoder and decoder blocks to capture long-range dependencies at multiple scales with minimal overhead. They proposed an efficient self-attention mechanism along with relative position encoding, which reduces the complexity of the self-attention operation significantly from $O(n^2)$ to approximate $O(n)$. An upgrade of their work, \ie, UTNetV2 \cite{gao2022multi}, which further proposed an efficient bidirectional attention (BMHA). Fu \etal \cite{fu2022tf} proposed TF-Unet, which is built on the intertwined backbone of convolution and Transformer at multiple scales. Besides, several works are proposed to improve the strategies of feature concatenation \cite{ning2021cac, you2022class}.

\begin{table*}[htbp]
\renewcommand{\arraystretch}{0.5}
\centering
\caption{\label{Table:generation} Transformers for image-to-image translation tasks in medical images. Abbreviations for the metrics: FID, Frechet Inception Distance; KID, Kernel Inception Distance; PSNR, Peak Signal-to-Noise Ratio; SSIM, Structural Similarity
Index; RMSE, Root Mean Square Error; MAE, Mean Absolute Error; NMSE, Normalized Mean Square Error.}
\setlength{\tabcolsep}{1pt}
\begin{tabular}{c|ccccc}
\toprule[1pt]

\textbf{Reference} & \textbf{Application} & \textbf{DataSet} & \textbf{Metrics} & \textbf{Task} \\
\hline
\\
\textbf{GIT, Watanabe \etal \cite{watanabe2021generative}} & Parkinson & 
    \begin{tabular}{ccc}
          the Parkinson’s Progression \\ Marker Initiative database \cite{marek2011parkinson}
    \end{tabular} 
    & \begin{tabular}{ccc}
          -
    \end{tabular} & \begin{tabular}{ccc}
         Image synthesis \\ (SPECT)
    \end{tabular}& \\
\\
\textbf{VTGAN, kamran \etal \cite{kamran2021vtgan}} 
    & Retinopathy
    & 
    \begin{tabular}{ccc}
          Fundus \& Fluorescein \\
          angiograms (FA) \cite{hajeb2012diabetic}
    \end{tabular}
    &\begin{tabular}{ccc}
          FID, KID
    \end{tabular}
    & \begin{tabular}{ccc}
          Image synthesis \\ (Fundus $\to$ FA)
    \end{tabular} 
    & \\
\\
\textbf{GANBERT, Shin \etal \cite{shin2020ganbert}} & Alzheimer & ADNI \footnote{http://adni.loni.usc.edu/} &\begin{tabular}{ccc}
 PSNR, SSIM, RMSE
\end{tabular}
& \begin{tabular}{ccc}
           Image synthesis \\ (MRI $\to$ PET)
    \end{tabular} & \\
\\
\textbf{Hu \etal \cite{hu2021data}} & Brain & IXI \footnote{http://brain-development.org/ixidataset/} &\begin{tabular}{ccc}
 PSNR, SSIM
\end{tabular}
& \begin{tabular}{ccc}
Image synthesis \\ (MRI $\to$ T1/T2)
    \end{tabular} & \\
\\
\textbf{SLATER, Korkmaz \etal \cite{korkmaz2021unsupervised}} & Brain & IXI\footnote{http://brain-development.org/ixidataset/}; fastMRI \cite{knoll2020fastmri} &
\begin{tabular}{ccc}
PSNR, SSIM
\end{tabular}
& Zero-shot MRI Reconstruction & \\
\\
\textbf{CyTran, Ristea \etal \cite{ristea2021cytran}} & Lung & Coltea-Lung-CT-100W \cite{ristea2021cytran} &
\begin{tabular}{ccc}
MAE, SSIM, RMSE
\end{tabular}
& \begin{tabular}{ccc}
          CT Translation \\ (Non-Contrast $\to$ Contrast) 
    \end{tabular} & \\
\\
\textbf{ResVit, Dalmaz \etal \cite{dalmaz2021resvit}} & Brain; Pelvic & 
    \begin{tabular}{ccc}
          IXI; BRATS \cite{menze2014multimodal,bakas2017advancing,bakas2018identifying}; \\
          pelvic MRI-CT datase \cite{nyholm2018mr} 
    \end{tabular}
&
\begin{tabular}{ccc}
PSNR, SSIM
\end{tabular}
& Multi-model Image synthesis & \\
\\
\hline
\\
\textbf{T$^2$Net, Feng \etal \cite{feng2021task}} & Brain & IXI; Clinical Dataset &
\begin{tabular}{ccc}
PSNR, SSIM, NMSE
\end{tabular}
& 
\begin{tabular}{ccc}
MRI Reconstruction \\ \& Super-resolution
\end{tabular}
& \\
\\
\textbf{PTNet, Zhang \etal \cite{zhang2021ptnet}} & Infant Brain & dHCP \cite{makropoulos2018developing} 
&
\begin{tabular}{ccc}
PSNR, SSIM
\end{tabular}
&\begin{tabular}{ccc}
MRI synthesis \\ \& Super-resolution 
\end{tabular} & \\
\\
\hline
\\
\textbf{TED-net, Wang \etal \cite{wang2021ted}} & Liver Lesions & 
    \begin{tabular}{ccc}
          2016 NIH-AAPMMayo Clinic LDCT \\
          Grand Challenge dataset \cite{mccollough2016tu}
    \end{tabular}
&
\begin{tabular}{ccc}
SSIM, RMSE
\end{tabular}
& Low-dose CT Denoising & \\
\\
\textbf{Eformer, Luthra \etal \cite{luthra2021eformer}} & Liver Lesions & 
    \begin{tabular}{ccc}
          2016 NIH-AAPMMayo Clinic LDCT \\
          Grand Challenge dataset \cite{mccollough2016tu}
    \end{tabular}
&
\begin{tabular}{ccc}
PSNR, SSIM, RMSE
\end{tabular}
& Low-dose CT Denoising & \\
\\
\toprule[1pt]
\end{tabular}\\
\end{table*}

\begin{table*}[htbp]
\renewcommand{\arraystretch}{1}
\centering
\caption{\label{Table:detection} Transformers for Detection Tasks.}
\setlength{\tabcolsep}{5pt}
\begin{tabular}{l|cccc}
\toprule[1pt]
\textbf{Reference} & \textbf{Disease} & \textbf{Organ} & \textbf{Dataset} &\textbf{Highlight}\\
\hline

\begin{tabular}{l}\textbf{RDFNet} \cite{jiang2021RDFNet}
\end{tabular} 
& Caries  
& Tooth 
& -
& \begin{tabular}{c} Transformer for feature extraction \end{tabular}
\\

\begin{tabular}{l}\textbf{COTR} \cite{shen2021COTR}
\end{tabular} 
& Polyp Lesion  
&\begin{tabular}{c}Colon\& \\ Rectum \end{tabular}
&\begin{tabular}{c} ETIS-LARIB \\ CVC-ColonDB\end{tabular}
& \begin{tabular}{c} Convolutions $\times$ Transformer \end{tabular} 
\\

\begin{tabular}{l}
\textbf{TR-Net} \cite{ma2021Transformer}
\end{tabular} 
&\begin{tabular}{c}Coronary Arteries \\ Significant Stenosis \end{tabular}
&\begin{tabular}{c}Coronary \\ Arteries \end{tabular} 
& -
& 
\\

\begin{tabular}{l}
\textbf{CT-CAD} \cite{kong2021ctcad}
\end{tabular} 
&\begin{tabular}{c}Chest Abnormality Detection \end{tabular}
&\begin{tabular}{c} Chest\end{tabular} 
& \begin{tabular}{c}Vinbig Chest \\Chest Det 10 \end{tabular}
& \begin{tabular}{c}context-aware\\ feature extractor \end{tabular}
\\

\begin{tabular}{l}
\textbf{Tao \etal} \cite{tao2021spine}
\end{tabular} 
&\begin{tabular}{c}Vertebrae Detection \end{tabular}
&\begin{tabular}{c} Spine\end{tabular} 
& \begin{tabular}{c}VerSe 2019 challenge\\MICCAI-CSI 2014 challenge
 \end{tabular}
& \begin{tabular}{c}inscribed sphere-based\\ object detector \end{tabular}
\\

\toprule[1pt]
\end{tabular}\\
\end{table*}

\noindent\textbf{Strategies of bridging Transformer and CNN.}

Unlike the aforementioned methods which combine Transformer and U-shaped architectures within a single inference path, other works explored different Transformer-CNN coupling strategies. Sun \etal \cite{sun2021hybridctrm} used Unet and Transformer encoders to generate representations independently, and then integrated their representations for subsequent decoding. Similarly, Li \etal \cite{li2021x} proposed X-Net, which used a CNN and a Transformer to extract the local and global features simultaneously. Zhang \etal \cite{zhang2021transfuse} proposed TransFuse, which also combines Transformers and Unet in a parallel style. Better than the abovementioned work, a novel fusion technique, \ie, BiFusion module, was proposed to efficiently fuse the multi-level features from both branches. Luo \etal \cite{luo2021hybrid} also used bi-directional cross-attention to fuse the local information extracted by the convolution operations and the global information learned by the self-attention mechanisms. Liu \etal \cite{liu2022phtrans} proposed PHTrans, which introduces the parallel hybird module in deep stages, where convolution blocks and the modified 3D Swin Transformer learn local features and global dependencies separately, then a sequence-to-volume operation unifies the dimensions of the outputs to achieve feature aggregation.

Zhou \etal \cite{zhou2021nnformer} claimed that most of the recently proposed Transformer-based segmentation approaches simply treat Transformers as assisted modules to help encode global context in convolutional representations, without investigating how to optimally combine self-attention with convolution. To address this issue, they introduced the nnFormer with an interleaved architecture based on empirical combination of self-attention and convolution. Xu \etal \cite{xu2021ect} proposed ECT-NAS method to search efficient CNN-Transformers architecture for medical image segmentation based on a multi-scale space search.

\noindent\textbf{Multi-scaling.}
The multi-scale strategy for Transformers in MIA uses features in a multi-scale manner or takes multi-scale images as inputs.

(1) Multi-resolution images. Zhang \etal \cite{zhang2021pyramid} proposed a pyramidal network architecture, namely Pyramid Medical Transformer (PMTrans), which captures multi-range relations by working on multi-resolution images. Valanarasu \etal \cite{valanarasu2021medical} added gated axial Transformer layers in the encoder, which contains the basic building block of both height and width gated multi-head attention blocks. The whole image and patches were used to learn global and local features correspondingly, and a Local-Global training strategy (LoGo) was proposed to further boost the overall performance.

(2) Multi-scale features. Different from TransUNet, which only uses Transformer to process the low-resolution feature maps learned from the previous layer, Xie \etal \cite{xie2021cotr} proposed a deformable Transformer (DeTrans) to process the multi-scale and high-resolution feature maps. Ji \etal \cite{ji2021multi} proposed Multi-Compound Transformer (MCTrans), which embeds the multi-scale convolutional features as a sequence of tokens, and performs intra- and inter-scale self-attention. Different from these works which use CNN to extract features, Hatamizadeh \etal \cite{hatamizadeh2021unetr} introduced UNEt TRansformers (UNETR) that utilize a pure Transformer as encoder to learn sequence representations of the input volume. The Transformer encoder is directly connected to a decoder via skip connections at different resolutions to compute the final semantic segmentation output. Zhang \etal \cite{zhang3992963s2wintounet} proposed S$^2$WinTOUnet, which used the star-shaped Window self-attention to obtain fine-grained details and coarse-grained semantic information.

(3) Multi-level attention. Chen \etal \cite{chen2021transattunet} proposed TransAttUnet, in which the multi-level guided attention and multi-scale skip connection are jointly designed to effectively enhance traditional U-shaped architecture. Both Transformer Self Attention (TSA) and Global Spatial Attention (GSA) are incorporated into TransAttUnet to effectively learn the non-local interactions between the encoded features. Wang \etal \cite{wang2021mixed} proposed the Mixed Transformer Module (MTM), which calculates self-affinities through well-designed Local-Global Gaussian-Weighted Self-Attention (LGG-SA) and then mines inter-connections between data samples through External Attention (EA). Wu \etal \cite{wu2022d} proposed the Dilated Transformer, which conducts self-attention for pair-wise patch relations that are captured alternately in local and global scopes.

(4) Multi-axial fusion. Yan \etal \cite{yan2021after} applied axial fusion Transformer to fuse the inter-slice and intra-slice information, which reduces the computational complexity of calculating self-attention in 3D space.

To conclude, the aforementioned methods all leveraged additional features learned using a feature fusion strategy for more effective learning.

\subsubsection{Pure Transformer}

Excluding the aforementioned variants of the Unet architecture combining the Transformer with convolutions, Karimi \etal \cite{karimi2021convolution} tried to use simple self-attention between adjacent image patches without convolution operations. A 3D image is divided into $n^3$ 3D patches ($n = 3$ or $5$), and a 1D embedding is learned for each patch. Through the self-attention between patch embeddings, the network outputs the segmentation result of the center patch. Methods under such assumption could be easily recognized as pure Transformers.

Cao \etal \cite{cao2021swin} proposed an Unet-like pure Transformer for medical image segmentation by feeding the tokenized image patches into the Transformer-like U-shaped Encoder-Decoder architecture with skip-connections for semantic feature learning in a local-global manner. 
Lin \etal \cite{lin2021ds} went a step further and proposed the DS-TransUNet, which first adopts dual-scale encoder sub-networks based on Swin-Transformer to extract the coarse and fine-grained feature representations of different semantic scales. A well-designed Transformer Interactive Fusion (TIF) module was also proposed to effectively establish global dependencies between features of different scales through the self-attention mechanism. 
To better leverage the natural multi-scale feature hierarchies of Transformers, Huang \etal \cite{huang2021missformer} proposed the MISSFormer, which has two appealing designs: 1) An enhanced Transformer Block as a feed forward network with better feature consistency, long-range dependencies and local context; and 2) An enhanced Transformer Context Bridge to model the long-range dependencies and local context of multi-scale features generated by the hierarchical Transformer encoder.

\subsection{Image-to-image translation}

Transformer models also have demonstrated their strong learning ability in many image-to-image translation applications including image synthesis \cite{parmar2018image}, reconstruction \cite{liang2021swinir}, and super-resolution \cite{yang2020learning}. However, in the field of medical image analysis, works (\eg, \cite{watanabe2021generative,kamran2021vtgan}) on image-to-image translation have recently started to emerge. Here we list existing Transformer-based image-to-image translation methods in Table \ref{Table:generation} as well as their evaluation metrics.

\subsubsection{Image synthesis}

\begin{figure}[htbp]
    \centering
    \includegraphics[width=\linewidth]{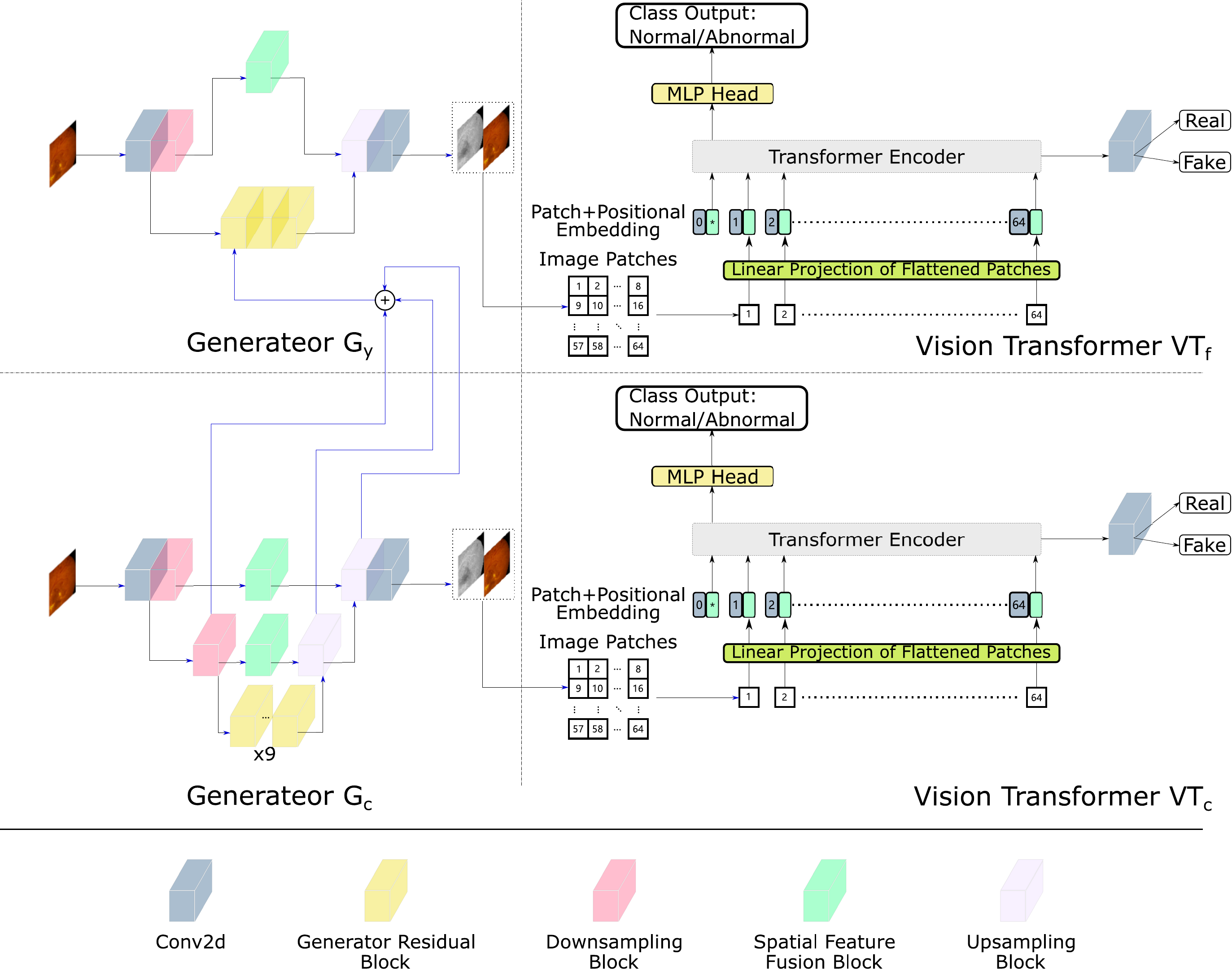}
    \caption{Overview of the architecture of VTGAN, which uses the Coarse and Fine generators $G_f$ , $G_c$ as generator, and the Vision Transformers $VT_f$, $VT_c$ as discriminators. \cite{kamran2021vtgan}}
    \label{fig:VTGAN}
\end{figure}

In the medical field, image synthesis remains very challenging due to inter-subject variability and the fact that anatomical hallucinations (e.g., hallucinating a white spot in a brain MRI) might be detrimental to diagnostic tasks. In recent years, generative adversarial learning is widely-used to tackle image synthesis tasks. Therefore, Transformers were combined with a generative adversarial learning paradigm for image synthesis. For example, Hu \etal \cite{hu2021data} introduced a double-scale discriminator GAN for cross-modal medical image synthesis, consisting of a transformer-based global discriminator and a CNN-based local discriminator. Watanabe \etal \cite{watanabe2021generative} proposed a generative model architecture based on a Transformer decoder block, thanks to its powerful ability in modeling time series. During the data processing, they normalized the pixel values of single photon emission computed tomography (SPECT) images by the specific/nonspecific binding ratio (SNBR). And during the training process, they adopted the Transformer decoder to construct the auto-regression model, and trained the model on $[^{123}I]$FP-CIT SPECT images of Parkinson’s progressive marker initiative database in an unpaired manner. The trained model can generate SPECT images that have characteristics of Parkinson’s disease patients. Kamran \etal \cite{kamran2021vtgan} proposed a Transformer-based conditional generative adversarial network, shown in Fig. \ref{fig:VTGAN} that can simultaneously perform semi-supervised image synthesis from fundus photographs to Fluorescein Angiography (FA) for retinal disease diagnosis.

To tackle the problem that the intensity range of Positron Emission Tomography (PET) is often wide and dense even heavily biased toward zero, Shin \etal \cite{shin2020ganbert} built a generative adversarial network by utilizing the Bidirectional Encoder Representations from Transformers (BERT), namely GANBERT, to generate PET images from MRI images. Luo \etal \cite{luo20213d} proposed a 3D-Transformer GAN to reconstruct the high-quality PET image at low dose. In order to overcome the limitation of scarce access to large medical datasets, Korkmaz \etal \cite{korkmaz2021unsupervised} introduced an unsupervised reconstruction method based on Zero-Shot Learned Adversarial TransformERs (SLATER) to perform the MRI synthesis. SLATER is an unconditional adversarial architecture consisting of a synthesizer, a discriminator, and a mapper. The synthesizer adopts the cross-attention Transformer blocks to capture long-range relationships and the mapper maps noise and latent variables onto MR images. Ristea et al. \cite{ristea2021cytran} proposed an architecture--CyTran which is based on generative adversarial convolutional Transformers and integrated the cycle-consistency loss for unpaired Computed Tomography (CT) images translation between contrast and non-contrast CT scans.

In addition to the work on image synthesis between two modalities, Transformer models were also successfully used in multi-modal medical image synthesis. For example, Dalmaz \etal \cite{dalmaz2021resvit} proposed a generative adversarial approach, namely ResViT, for multi-modal medical image synthesis. The generator in ResViT is based on the encoder-decoder architecture with a central bottleneck that comprises the aggregated residual Transformer (ART) blocks capable of synergistically preserving local and global contexts.

\subsubsection{Image super-resolution}

Super-resolution imaging (SR) is a class of techniques that enhance the resolution of an imaging system and it is also a popular sub-field of image synthesis. Outstanding contributions have been made by the Transformer models on super-resolution tasks in medical image analysis. For instance, Feng \etal \cite{feng2021task} introduced a task Transformer network (T$^2$Net) to jointly learn the Image reconstruction and super-resolution tasks in MRI. It is a multi-task framework, including a super-resolution branch and a common resolution branch, and the authors designed the Transformer module to embed the similarity and align the gap between the two branches. Zhang \etal \cite{zhang2021ptnet} proposed a high-resolution synthesizer based on Pyramid Transformer (PTNet) and used for MRI synthesis of infant brains. PTNet consists of a performer encoder (PE), a performer decoder (PD), and a Transformer bottleneck which inherits the U-structures as well as multi-resolution pyramid structures.

\subsubsection{Image denoising}

Image Denoising is the task of removing noise from an image. It is a fundamental step to several clinical applications. For example, Wang \etal \cite{wang2021ted} used the Transformer for low dose CT (LDCT) denoising for the first time. They developed an Encoder-decoder Dilation network based on Token-to-Token (T2T) vision Transformer, namely TED-net. TED-net is also a U-structure model that utilizes the dilation in the T2T stage to enlarge the receptive field. Luthra et al. \cite{luthra2021eformer} proposed Edge enhancement based Transformer (Eformer) which uses Transformer blocks to construct an encoder-decoder architecture for medical image denoising. Transformer models and their applications in the task of low dose CT (LDCT) denoising remain scarce.

\subsection{Detection}

The meaning and terminology of 'detection' varies across technical and clinical fields. In technical area, it often refers to checking the existence of diseases or lesions, while, in clinical practice, it often means diagnosis or disease classification, as discussed above. In computer vision, detection aims to identify the location of objects in an input image and predict their categories/classes. In this section, detection refers to the application of object detection. 

Transformers dealing with detection tasks using medical images are often combined with CNN blocks, where a CNN is used to extract features from medical images while the Transformer architecture is used to enhance the extracted features for downstream detection.
Shen \etal \cite{shen2021COTR} proposed a DETR-based model, namely COTR, for the detection of the polyp in the colon. DETR \cite{detr} is a primer method for object detection in computer vision. COTR is composed of a CNN for feature extraction, Transformer encoder layers interleaved with convolutional layers for feature encoding and recalibration, Transformer decoder layers for object querying, and a feed-forward network for detection prediction. They proposed to insert convolutional layers into the Transformer encoder for high-level image feature reconstruction and convergence acceleration. Ma \etal \cite{ma2021Transformer} proposed a TR-Net which combines CNN and Transformer nets to detect significant stenosis in multiplanar reformatted (MPR) image. Their model employs a shallow 3D-CNN to extract local semantic features of coronary regions while ensuring the model efficiency. Next, Transformer encoders are used to learn
correlations between different regions of the local stenosis at each position of a coronary artery. Thus, the TR-Net can accurately detect stenosis after aggregating information from local semantic features and global semantic features.
Jiang \etal \cite{jiang2021RDFNet} constructed a YOLOv5s-based Transformer for the detection of caries, called RDFNet. The model uses the FReLU activation function to activate the complex visual-spatial information of images for efficiency boosting.
Kong \etal \cite{kong2021ctcad} proposed CT-CAD, a context-aware hybrid Transformer for end-to-end chest abnormality detection on X-Ray images. 
Tao \etal \cite{tao2021spine} designed a Spine-Transformer to address the automatic detection and localization of vertebrae in arbitrary Field-Of-View Spine CT. They formulated the detection as an one-to-one set prediction problem.

\subsection{Registration}

Transformers have several advantages in image registration tasks thanks to their self-attention mechanism which enables a more precise spatial mapping between moving and fixed images. Chen \etal pioneered the first Transformers for image registration. Inspired by the architecture of TransUnet \cite{chen2021transunet}, they proposed ViT-V-Net \cite{chen2021vit}, which combines ViT and V-Net by simply altering the network architecture of VoxelMorph (a conventional registration network) \cite{balakrishnan2019voxelmorph}. The proposed ViT-V-Net produced superior performance against benchmark methods. Afterward, they extended their work and presented TransMorph \cite{chen2021transmorph}, for volumetric medical image registration. In this method, the Swin-Transformer \cite{liu2021swin} was used as the encoder network to capture the spatial correspondence between the input moving and fixed images. Next, a ConvNet decoder mapped the information provided by the Transformer encoder onto a dense displacement field. Long skip connections were deployed to maintain the flow of local information between the encoder and decoder stages. Obviously, Transformers-based registration methods are rare, that needs more explorations and research interests.

\subsection{Video-based applications}

Because of the limited receptive field, CNNs cannot fully utilize the global temporal and spatial information in continuous video frames, whereas Transformers can overcome such defects. Ji \etal \cite{ji2021progressively} proposed PNS-Net (Progressively Normalized Self-attention Network) for accurate polyp segmentation from colonoscopy videos. Kondo \etal \cite{kondo2021lapformer} proposed LapFormer to detect the surgical tools in laparoscopic surgery videos. Czempiel \etal \cite{czempiel2021opera} introduced OperA to predict surgical phases from long video sequences. Reynaud \etal \cite{reynaud2021ultrasound} firstly adopted the Transformer architecture, which contains a Residual AutoEncoder Network and a BERT model, to analyze videos of arbitrary length. Long \etal \cite{long2021dssr} firstly applied Transformer to estimate surgical scene depth.

\section{Discussion}
\label{sec:discus}

\begin{table*}[htbp]
\renewcommand{\arraystretch}{0.5}
\centering
\caption{\label{Table:multimodal} Transformers for multi-modal learning.}
\setlength{\tabcolsep}{5pt}
\begin{tabular}{l|ccccc}
\toprule[1pt]
\textbf{Reference} & \textbf{Disease} & \textbf{Organ} & \textbf{Dataset} &\textbf{Highlight} & \textbf{Task} \\
\hline
\textbf{Multi-modal} \\
\cline{1-1}
\begin{tabular}{l} \textbf{CLIMAT} \end{tabular} \cite{Prognosis-CLIMAT-nguyen2021climat}
& Alzheimer's Disease 
& Brain 
& ADNI 
& \begin{tabular}{c} use multi-Trans to mimic radiologist\\ and general practitioner interactions\end{tabular}  
& Prognosis \\
 
\begin{tabular}{l}\textbf{Zheng \etal} \end{tabular} \cite{GraphLearning-GrapBrain-zheng2021multimodal}
& Alzheimer’s Disease 
& Brain 
& TADPOLE  
& \begin{tabular}{c} propose a modal-attentional\\ multi-modal fusion \end{tabular}
& Prediction\\ 

\begin{tabular}{l}\textbf{Qiu \etal}\end{tabular} \cite{GraphLearning-GraphAD-9433842}
& Alzheimer’s Disease 
& Brain
& ADNI  
& \begin{tabular}{c}build a graph based on rich multi-modal \\features; propose Graph Trans to classify \end{tabular} 
& Classification
\\ 
 
& \begin{tabular}{c} Autism Spectrum \end{tabular} 
& Brain 
& ABIDE  
&  
& Prediction\\

\begin{tabular}{l} \textbf{BERTHop} \end{tabular} \cite{monajatipoor2021berthop}
& Thoracic Disease  
& Chest 
& OpenI 
& \begin{tabular}{c}incorporates PixelHop++ into a Trans-based\\ model; adopt in-domain pretrained BERT \end{tabular}
& Diagnosis \\
 
\begin{tabular}{l}\textbf{DRT} \end{tabular} \cite{Pure-Trans-Eye-song2021deep}
& Glaucoma
& Eye
& ZOC-OCT\&VF 
& \begin{tabular}{c}use two relation module to extract inter-\\modal relation; use Trans to fuse features\end{tabular} 
& Diagnosis \\

\begin{tabular}{l}\textbf{TransMed} \cite{CNN+Trans-TransMed-dai2021transmed}
\end{tabular}
& Parotid Gland Tumor  
& \begin{tabular}{c} Parotid 
\\ Gland \end{tabular}  
& *  
& \begin{tabular}{c}use Trans to capture cross-modality \\mutual information and fuse features \end{tabular} 
& Prediction \\

\begin{tabular}{l}\textbf{Jacenków} \cite{prior-cxr}
\end{tabular}
& Thoracic Diseases  
& \begin{tabular}{c}  
\\ thorax
\end{tabular}  
& MIMIC-CXR  
& \begin{tabular}{c}use text to assist image classification \end{tabular} 
& Classification \\

\toprule[1pt]
\end{tabular}\\
\end{table*}

Transformers have been successfully applied to plenty of applications in almost all fields of medical image analysis. However, the deployment of machine learning methods in real clinical applications might lead to a poor performance due to several challenges. Among them, the most urging challenge lies in label scarcity, especially in scene understanding tasks, \eg, segmentation and detection, which usually needs pixel-wise precise labeling. Learning from noisy labels present a bigger challenge. Besides, building advanced CADx methods require the use of multi-modality clinical data in a multi-task manner -- a versatile learning approach nevertheless difficult in design.

\subsection{Transformers under different learning scenarios}

\subsubsection{Multi-task learning}

Building models with multiple tasks help improve their generalizability, which is highly demanded in the field of medical image analysis. A frequently used framework is to unify classification and segmentation into one model. \cite{ azzuni2022color, MT-TransUNet}. For instance, Chen \etal \cite{MT-TransUNet} proposed the Multi-Task TransUNet (MT-TransUNet) to jointly learn the segmentation and classification of skin lesion. With Local details (\eg, skin color, texture) and long-range context (\eg, skin lesion shape, physical size) extracted by CNNs and ViTs, the method achieved SOTA performance and obtained efficiency improvements in model parameters and inference speed.
Besides, Sui \etal \cite{sui2021cst} combined the detection with segmentation tasks, and proposed a novel transfer learning method, \ie, CST, with a Transformer-based framework for joint colorectal cancer region detection and tumor segmentation. For detection, the generated region proposals of the input images, as well as the position features obtained by the encoder-decoder module, were used as the input to a DETR network. For segmentation, the model used image patches as input, which were projected into a sequence of embeddings.

\subsubsection{Multi-modal learning}

Using multiple modality data provides complementary evidence for diagnosis. For example, researchers have explored the combination of Optical Coherence Tomography (OCT) and Visual Field (VF) test to aid in the diagnosis of eye diseases. Song \etal \cite{Pure-Trans-Eye-song2021deep} proposed to employ Transformers for Glaucoma diagnosis. 
The proposed model utilized the attention mechanism to model the pairwise relation between OCT features and VF features. 
Next, the attention mechanism is applied again to calculate the regional relation of features between the visual field areas and the quadrants of retinal nerve fiber layer. 
The complementary information is passed from one modality to another by utilizing the Transformer model.

Monajatipoor \etal \cite{monajatipoor2021berthop} proposed a Transformer-based vision-and-language model which combined the efficient PixelHop++ model with the BERT model. Specifically, the BERT model was pretrained using in-domain knowledge. The model is proved to be effective when trained on small-scale datasets. The extracted vision features and the word embeddings are fed into the Transformer for final diagnosis. 
Although the model decreased the need of massive annotations of medical images, the pretraining of the language model still needed a large amount of clinical reports.
Jacenków \etal \cite{prior-cxr} combined the text with CXR for disease classification. They observed that the interpretation and reporting of an image is affected by the scan request text, which served as the indication field in the radiology report. Zheng \etal \cite{GraphLearning-GrapBrain-zheng2021multimodal} focused on the feature fusion of multi-modal information, by considering the latent inter-modal correlation. They proposed the Transformer-like modal-attentional feature fusion approach (MaFF) to extract rich information from each modality while mining the inter-modal relationships. Next, an adaptive graph learning mechanism (AGL) is utilized to construct latent robust graphs for downstream tasks based on the fused features. The method achieved significant improvement for the prediction of AD and Autism.
Dai \etal \cite{CNN+Trans-TransMed-dai2021transmed} proposed the TransMed for the diagnosis of parotid gland tumor. TransMed combines the advantages of CNN and Transformer networks to capture both low-level textures and cross-modality high-level relationships. The model first processes multi-modal images as sequences by chaining and sending them to a CNN for feature extraction. The feature sequences are then fed into the Transformers to learn the relationship between the sequences as well as conducting feature fusion. 
Their work leveraged Transformers to capture mutual information from images of different modalities which showed better performance and efficiency. 
Nguyen \etal \cite{Prognosis-CLIMAT-nguyen2021climat} attempted to mimic the interaction between radiologist and general practitioner for the diagnosis and prognosis of knee osteoarthritis.
They proposed a Clinically-Inspired Multi-Agent Transformers (CLIMAT) framework with a tri-Transformer architecture. Firstly, a feature extractor with the combination of Transformer and CNN, is used to predict the current state of a disease. 
Next, the non-image auxiliary information is fed into another Transformer to extract context embedding. 
Finally, an additional Transformer-based general practitioner module forecasts the disease trajectory based on the current state and context embedding. 

To conclude, Transformer is regarded as a promising approach to bridge CV and NLP tasks \cite{han2020survey}. Under this assumption, Radford \etal \cite{radford2021learning} built a multi-modal Transformer, \ie, CLIP, that provides zero-shot ability for recognizing images from their text descriptions without image labeling. Such strength also points out to a potential way of building more robust and accurate computer aided diagnosis (CADx) methods for real clinical applications, where multiple data types, \eg, clinical, laboratory and imaging data are considered as diverse source of information.

\begin{table*}[htbp]
\renewcommand{\arraystretch}{0.5}
\centering
\caption{\label{Table:multitask} Transformers for weakly-supervised learning.}
\setlength{\tabcolsep}{1pt}
\begin{tabular}{l|ccccc}
\toprule[1pt]
\textbf{Reference} & \textbf{Disease} & \textbf{Organ} & \textbf{Dataset} &\textbf{Highlight} & \textbf{Task} \\

\hline

\begin{tabular}{l} \textbf{Weakly} \\  \textbf{Supervised} \end{tabular} \\
\cline{1-1}

\begin{tabular}{l}\textbf{Li \etal} \cite{MIL-Self-Attention-MILTrans-li2020deep}
\end{tabular} 
& Diabetic Retinopathy  
& Eye 
& Messidor   
& \begin{tabular}{c}Induced Self-Attention to model\\relation of instances within a bag\end{tabular}
& Classification\\

\begin{tabular}{l}\textbf{Rymarczyk \etal} \cite{MIL-SA-AbMILP-rymarczyk2021kernel}
\end{tabular} 
& Diabetic Retinopathy 
&  Eye 
& Messidor   
& \begin{tabular}{c}self-attention with\\ Attention-based MIL Pooling\end{tabular}
&  Classification\\

\begin{tabular}{l}
\textbf{Yang \etal} \cite{MIL-NoduleSATs-yang2020relational}
\end{tabular}
& \begin{tabular}{c} Multiple Nodule 
\\ Malignancy\end{tabular} 
& Lung & LIDC-IDRI  
& \begin{tabular}{c}inter-solitary-nodule relationships\end{tabular}  
& Classification \\
& Lung Nodule  
& Lung 
& LUNA16, Tianchi Val  
& 
& Detection\\

\begin{tabular}{l}\textbf{MIL-VT} \cite{MIL-MIL-VT-yu2021mil}
\end{tabular} 
& Diabetic Retinopathy 
& Eye 
& APTOS2019 
& \begin{tabular}{c}MIL-head to provide complementary\\information of patches to the class token\end{tabular} 
& Classification\\
& Retinal Fundus Disease  
& Eye 
& RFMiD2020
& 
& Diagnosis\\

\begin{tabular}{l} \textbf{TransMIL} \cite{MIL-TransMIL-shao2021transmil} 
\end{tabular} 
& \begin{tabular}{c}Breast Cancer
\\ Metastasis\end{tabular} 
& Breast 
& CAMELYON16  
& \begin{tabular}{c}explore both morphological and spatial\\information between different instances\end{tabular}
& Detection\\
& Lung Cancer  
& Lung 
& TCGA-NSCLC  
&  
& Classification\\
& Kidney Cancer  
& Kidney 
& TCGA-RCC 
& 
& Classification \\

\hline

\begin{tabular}{l} \textbf{Self-Supervised} \end{tabular} \\
\cline{1-1}

\begin{tabular}{l}\textbf{Park \etal} \cite{Pre-training-COVID-19-park2021vision}
\end{tabular} 
& COVID-19  
& Lung 
& *
& \begin{tabular}{c}pretrain model on large scale data; eva-\\luate necessity of self-supervised pretrain \end{tabular}
& Diagnosis\\ 
 
\begin{tabular}{l}\textbf{Jun \etal} \cite{Self-Pre-training-BrainEncoder-jun2021medical}
\end{tabular}
& Alzheimer’s Disease 
& Brain 
& ADNI 
& \begin{tabular}{c}pretrain Trans with masked encoding\\vector prediction as SSL proxy task\end{tabular}
& Diagnosis\\
& \textit{Brain Age} 
& Brain 
& ADNI  
& 
& Prediction\\

\begin{tabular}{l}\textbf{TransPath} \cite{Transfer-Transpath-wang2021transpath}
\end{tabular}  
& Colorectal Cancer  
& Colorectal 
& NCT-CRC-HE   
& \begin{tabular}{c}collect approximately 2.7 million\\images for self-supervised pretraining \end{tabular}
& Classification \\
& Breast Cancer 
& Breast  
& PatchCamelyon 
& 
& Classification  \\
& Colorectal Polyps  
& Colorectal
& MHIST  
&
& Classification  \\

\begin{tabular}{l}\textbf{Truong \etal} \cite{CNN-Transfer-truong2021transferable}
\end{tabular}
& \begin{tabular}{c} Axillary Lymph \\ Node Cancer \end{tabular}
& Lymph
& PatchCam 
& \begin{tabular}{c}validate ViT-based Self-Supervised\\method by comparison\end{tabular}
& Classification\\
& Diabetic Retinopathy
& Eye
& APTOS  
& 
& Classification\\
& Pneumonina
& Chest
& \begin{tabular}{c} Pneumonina \\ chest X-ray\end{tabular}  
& 
& Classification\\
& Thorax Disease
& Chest
& \begin{tabular}{c} NIH chest X-ray\end{tabular}  
& 
& Classification\\

\begin{tabular}{l}\textbf{Sriram} \cite{Transfer-Learning-COVIDPrognosis-sriram2021covid}
\end{tabular} 
& COVID-19 
& Lung 
& \begin{tabular}{c}
NYU COVID 
\end{tabular} 
& \begin{tabular}{c}adopt Momentum Contrastive Learning\\for Self-Supervised Learning pretrain\end{tabular} 
& Prognosis \\

\begin{tabular}{l}\textbf{Chen} \cite{Transfer-Learning-COVIDPrognosis-sriram2021covid}
\end{tabular} 
& Tissue Phenotyping 
& Tissue 
& \begin{tabular}{c}
TCGABRCA cohort\\CRC-100K BreastPathQ
\end{tabular} 
& \begin{tabular}{c}DINO-based knowledge distillation \\ applied on ViT\end{tabular} 
& Classification \\

\toprule[1pt]
\end{tabular}\\
\end{table*}

\subsubsection{Weakly-supervised learning}

One of the weakly-supervised conditions in medical images is that the ROI for a certain disease is relatively small in the image, whereas only image-level labeling is available.
To solve this problem, multiple instance learning (MIL) was adopted as a suitable solution. In MIL, the training samples present sets of instances, called bags. The supervision is provided only for bags, and the individual labels of the instances contained in the bags are not provided \cite{MIL-Survey2018}. 

Although many existing MIL methods assume that positive and negative instances are sampled independently from a positive and a negative distribution \cite{MIL-Survey2018}, instances in a bag are relational, especially in medical image analysis. The learning scenario of MIL does not follow the i.i.d assumption, since the relationships between instances are not neglected. In such situation, ViTs can be leveraged to build correlations between instances for better high-level representations. 
Li \etal \cite{MIL-Self-Attention-MILTrans-li2020deep} proposed the Transformer-based MIL framework, with an induced attention block which calculates the attention while bypassing the quadratic computational complexity caused by the pairwise dot product. The feature aggregator of the framework is also based on multi-head attentions. It merges the previously mentioned features into bag representations.
Yang \etal \cite{MIL-NoduleSATs-yang2020relational} treated the multiple pulmonary nodules of a patient as a bag, and each nodule as an instance. Unlike conventional MIL methods that use the pooling operation to get bag-level representations, they proposed to use a 3D DenseNet to learn solitary-nodule-level representation at the voxel level. Next, the generated representations are fed into the Transformer to learn the nodule relationships from the same patient. To reduce the computational burden, they applied in-group scaled-dot production attention, extracted from split channel features. Shao \etal \cite{MIL-TransMIL-shao2021transmil} focused on the correlation between different instances as opposed to simply assuming that instances are independent and identically distributed. To this end, they proposed a Transformer-based MIL framework to deal with the whole slide image classification problem. Their framework used Transformer layers to aggregate morphological information, and proposed the Pyramid Position Encoding Generator (PPEG) to extract the spatial information. Besides, they adopt the Nystrom method to calculate approximated self-attentions, which can reduce computational complexity from $O(n^2)$ to $O(n)$. 
Rymarczyk \etal \cite{MIL-SA-AbMILP-rymarczyk2021kernel} paid more attention to the attention mechanism. Their work contributed to revise the Attention-based MIL Pooling (AbMILP) which aggregates information from a varying number of instances. They proposed the Self-Attention Attention-based MILPooling (SA-AbMILP) to model the dependencies between different instances within a bag. They also proposed to extend the calculation of attentions by introducing different kernels, which played a same role as the dot production. They evaluated their work on histological, microbiological, and retinal datasets. The work from Yu \etal \cite{MIL-MIL-VT-yu2021mil} explored the applicability of ViTs on retinal disease classification in fundus images. They proposed the Multiple Instance Learning enhanced Vision Transformer (MIL-VT) by adding a plug-and-play multiple instance learning head to the ViT to exploit the features extracted from individual patches.

Another weakly-supervised example is the semi-supervised learning, which requires only a small amount of labeled data to exploit the knowledge from a large amount of unlabeled data. Luo \etal \cite{luo2021semi} first combined a CNN and Transformer for semi-supervised medical image segmentation. They introduced the cross teaching between CNN and Transformer, where the prediction of a network is used as the pseudo label to supervise the other network. Zhao \etal \cite{zhaocontext} proposed a context-aware network called CA-Net for semi-supervised LA segmentation from 3D MRI. CA-Net contains two main modules, a Trans-V module that combines Transformer and V-Net to learn contextual information, and a discriminator to calculate an adversarial loss for learning the unlabeled data. Xiao \etal \cite{xiao4081789efficient} simultaneously used Dual-Teacher structure of CNN and transformer to guide the student segmentation model.

\subsubsection{Self-supervised learning} 

Successful training of the Transformer relies on \emph{large-scale} annotated data, which are rarely available in real clinical facilities. The paradigm of self-supervised learning (SSL) was created to handle such issue. Self-supervised learning aims at improving the performance of downstream tasks (\eg, classification, detection and segmentation), by transferring the knowledge from the related unsupervised upstream task (\ie, learning of vision concepts), and pretrains the model using its self-contained information in the unlabeled data \cite{Transfer-Survey-zhuang2020comprehensive}. 
The practice of training SLL ViTs is generally rooted in pretraining the model on ImageNet followed by a fine-tuning step on the target medical image dataset. This led to boosting the performance of ViTs in comparison with CNNs and achieving the SOTA accuracy \cite{CNN-VIT-Pretrain-matsoukas2021time, luo2022self, malkiel2021pre, xie2021unified}. 

The work from Truong \etal \cite{CNN-Transfer-truong2021transferable} evaluated the transferability of self-supervised features in medical images. They pretrained features using DINO, a self-supervised ViT. They used the ViT as a backbone, and showed its outperformance in comparison with SimCLR and SwAV. Park \etal \cite{Pre-training-COVID-19-park2021vision} proposed to use the public large-scale CXR classification dataset to pretrain the backbone network. The features extracted by the pretrained backbone model are then fed into a ViT to diagnose COVID-19. 
Jun \etal \cite{Self-Pre-training-BrainEncoder-jun2021medical} proposed a self-supervised transfer learning framework that can better represent the spatial relationships in 3D volumetric images, to facilitate downstream tasks. They converted the 3D volumetric images into sequences of 2D image slices from three views, and fed them into the pretrained backbone network, which consists of a convolutional encoder and a Transformer. The pretraining of the Transformer was implemented through masked encoding vectors, which served as a proxy task for SSL. The downstream tasks include brain disease diagnosis, brain age prediction and brain tumor segmentation, using 3D volumetric images. They also explored a parameter-efficient transfer learning framework for 3D medical images. Wang \etal \cite{Transfer-Transpath-wang2021transpath} collected a large public histopathological image dataset to pretrain their proposed hybrid CNN-Transformer framework. Moreover, they designed the token-aggregating and excitation (TAE) module to further enhance global weight attention by taking all tokens into consideration.
Sriram \etal \cite{Transfer-Learning-COVIDPrognosis-sriram2021covid} explored the application of Transformers for COVID-19 prognosis. They proposed a Multiple Image Prediction (MIP) model, which takes a sequence of images along with the corresponding scanning time as input. To deal with missing COVID-19 images, they used the Momentum Contrast Learning, which is a self-supervised method to pretrain the feature extractor network. In addition to the features extracted from X-rays, they also proposed the Continuous Positional Embedding (CPE) to add information based on the time-step. The concatenation of features and continuous positional embeddings are fed into the Transformer to predict the possibility of an adverse event.
Chen \etal \cite{chen2022self} showed that Vision Transformers using DINO-based knowledge distillation is able to learn data-efficient and interpretable features in histology images by training various self-supervised models with validation on different weakly-supervised tissue phenotyping tasks. It is worth noting that they leveraged an excellent performance on different attention heads in the ViT while learning distinct morphological phenotypes.

\subsection{Model-improvement: quantification, acceleration and interpretation}

Several works focused on the model efficiency within the medical imaging field.
A natural idea is to simplify the attention mechanism, which demands the largest workload in Transformers. Gao \etal \cite{gao2021utnet} proposed an efficient self-attention mechanism and position encoding, which significantly reduced the complexity of the self-attention operation from $O(n^2)$ to approximate $O(n)$. This circumvented the hurdle that Transformers require a huge amount of data to learn vision inductive bias. Their hybrid-layer design initializes Transformers as convolutional networks without the need of pretraining.
Besides, the aforementioned Vision Outlooker (VOLO) adopted by Liu \etal \cite{PureTrans-VOLO-liu2021automatic} replaced the standard self-attention with outlook attention which performs the internal self-attention mechanism, reducing the original space time complexity. Li \etal \cite{li2022transbtsv2} redesigned the Transformer block in their work TransBTSV2, pursuing a shallower but wider architecture compared with the conventional Transformer-based methods. Inspired by the dilated convolution kernels, Wu \etal \cite{wu2022d} conducted the global self-attention in a dilated manner, enlarging the receptive fields without increasing the patched and reducing the computational costs. 
Xu \etal \cite{xu2021ect} built a multi-scale searching space composed of a multi-branch parallel searching block, which connects CNN and Transformer in parallel. They also proposed an efficient resource constrained search strategy to simultaneously optimize the accuracy and costs (\eg, Params. and FLOPs) of the model.

We have seen fewer works attempting to solve the model efficiency problem in MIA rather than in CV. However, as medical images come in larger sizes and smaller quantities, there is an urging need of solving this problem in the field. Thus, we would like to see more works in this specific research direction.

\subsection{Comparison with convolutional neural networks}

Convolutional Neural Networks were dominant in CV prior to the emergence of ViTs, including the field of medical image analysis. Lots of efforts have been invested in improving the performance of CNN-based classifiers both in natural and medical images. Several works were proposed to investigate whether CNN-based methods can still work on ViTs. Meanwhile, as ViTs have ranked on the top of several benchmarks, lots of researches have focused on the performance comparison between ViTs and CNNs. 

Large scale datasets are required for obtaining desirable performance with Transformer. However, in the medical image analysis field, available images and annotations are limited. To alleviate this problem, many methods adopted convolutional layers in ViTs to boost the performance with limited medical images, and also leveraged the power of transfer learning and self-supervised learning. 
Matsoukas \etal \cite{CNN-VIT-Pretrain-matsoukas2021time} explored whether transfer learning and self-supervised learning regimes can benefit ViTs. They conducted several experiments to compare the performance of a CNN (\ie, ResNet50) and a ViT (\ie, DEIT-S) using different initialization strategies: 1) randomly initialized weights, 2) transfer learning using ImageNet pretrained weights, 3) self-supervised pretraining on the target dataset, with the same initialization in 2). They evaluated these methods on APTOS 2019, ISIC 2019 and CBIS-DDSM dataset. It can be concluded that the standard procedures, \eg, initialization using ImageNet pretrained weights, and leveraging self-supervised learning, can bridge the performance gap between CNN and ViT. 
Krishnamurthy \etal \cite{CNN-ViT-Transfer-krishnamurthy2021} adopted the transfer learning scheme in both CNNs and ViTs for Pneumonitis diagnosis. They first pretrained the models on ImageNet and fine-tuned the classifier on their private dataset. However, their comparison was based on fine-tuning with frozen backbone layers, which limits the performance of feature extraction when adapted to the target domain. 
Truong \etal \cite{CNN-Transfer-truong2021transferable} assessed the transferability of self-supervised features in medical imaging tasks. They chose ResNet-50 as the backbone, and pretrained it using three self-supervised methods, SimCLR, SwAV and DINO. DINO used the ViT as the backbone, that consistently outperformed other self-supervised techniques as well as the supervised baseline by a large margin. They proposed a model-agnostic technique, \ie, Dynamic Visual Meta-Embeddings (DVME), to combine pretrained features from multiple self-supervised learning methods with self-attention. 

For the task of multi-scale cell image classification, Liu \etal \cite{CNN-ViT-CellRatio-liu2021aspect} proposed an experimental platform to compare multiple deep learning methods, including CNNs and ViTs. They validated the performance of deep learning models on standard and scaled data, by changing the internal cell ratios of the images. The results suggested that the deep learning models, including ViTs, are robust to the change of internal cell ratio in cervical cytopathological images. 
For shoulder implant X-Ray manufacturer classification, Zhou \etal \cite{Compare-Implant-zhou2021shoulder} compared the performance of various models, including traditional machine learning methods, CNN-based deep learning methods, and ViTs. The results showed that ViTs achieved the best performance in these tasks, and transfer learning improved ViT by a large margin. 
Altay \etal \cite{TransAD-altay2020preclinical} aimed at early pre-clinical prediction of AD using MRI. They compared Transformers against the baseline 3D CNN model and 3D recurrent visual attention model, and showed that Transformers achieve the best accuracy and F1 scores.
Adjei-Mensah \etal \cite{cnn-vit-low-resolution} showed that the CNNs outperformed the ViTs on low-resolution medical image recognition.
Galdran \etal \cite{cassidy2021diabetic} also showed that CNNs outperformed ViTs on Diabetic Foot Ulcer Classification in a few-data regime.

In summary, existing works have not shown that ViTs can outperform CNNs in all scenarios, particularly in both few-shot and low-resolution of medical image analysis. Thus, similar to the methods in computer vision, building hybrid models with convolutions was adopted by most recent works.

\section{Conclusion}
\label{sec:con}

Transformers are now transforming the field of computer vision. Also, researches using Transformers are undergoing a rapid growth in the field of medical image analysis. However, most of the current Transformer-based methods are naturally and simply applied to medical imaging problems without drastic changes. In other words, advanced methodologies, \eg, weakly-supervised learning, multi-modal learning, multi-task learning, and model improvement, are rarely explored. Also, we only see a few works focusing on general problems of the model, \eg, parallelization, interpretability, quantification and safety. These indicate future directions of medical Transformers.

\section*{Acknowledgement}

This work was supported in part by the National Nature Science Foundation of China under grant No. 62106101. This work was also supported in part by the Natural Science Foundation of Jiangsu Province under grant No. BK20210180.


%





\ifCLASSOPTIONcaptionsoff
  \newpage
\fi



%
\bibliographystyle{IEEEtran}
\bibliography{IEEEabrv,TIM}

\begin{thebibliography}{100}
\providecommand{\url}[1]{#1}
\csname url@samestyle\endcsname
\providecommand{\newblock}{\relax}
\providecommand{\bibinfo}[2]{#2}
\providecommand{\BIBentrySTDinterwordspacing}{\spaceskip=0pt\relax}
\providecommand{\BIBentryALTinterwordstretchfactor}{4}
\providecommand{\BIBentryALTinterwordspacing}{\spaceskip=\fontdimen2\font plus
\BIBentryALTinterwordstretchfactor\fontdimen3\font minus
  \fontdimen4\font\relax}
\providecommand{\BIBforeignlanguage}[2]{{%
\expandafter\ifx\csname l@#1\endcsname\relax
\typeout{** WARNING: IEEEtran.bst: No hyphenation pattern has been}%
\typeout{** loaded for the language `#1'. Using the pattern for}%
\typeout{** the default language instead.}%
\else
\language=\csname l@#1\endcsname
\fi
#2}}
\providecommand{\BIBdecl}{\relax}
\BIBdecl

\bibitem{vaswani_attention_2017}
\BIBentryALTinterwordspacing
A.~Vaswani, N.~Shazeer, N.~Parmar, J.~Uszkoreit, L.~Jones, A.~N. Gomez,
  L.~Kaiser, and I.~Polosukhin, ``Attention {Is} {All} {You} {Need},''
  \emph{arXiv:1706.03762 [cs]}, Dec. 2017, arXiv: 1706.03762. [Online].
  Available: \url{http://arxiv.org/abs/1706.03762}
\BIBentrySTDinterwordspacing

\bibitem{dong2018speech}
L.~Dong, S.~Xu, and B.~Xu, ``Speech-transformer: a no-recurrence
  sequence-to-sequence model for speech recognition,'' in \emph{2018 IEEE
  International Conference on Acoustics, Speech and Signal Processing
  (ICASSP)}.\hskip 1em plus 0.5em minus 0.4em\relax IEEE, 2018, pp. 5884--5888.

\bibitem{li2019neural}
N.~Li, S.~Liu, Y.~Liu, S.~Zhao, and M.~Liu, ``Neural speech synthesis with
  transformer network,'' in \emph{Proceedings of the AAAI Conference on
  Artificial Intelligence}, vol.~33, no.~01, 2019, pp. 6706--6713.

\bibitem{vila2018end}
L.~C. Vila, C.~Escolano, J.~A. Fonollosa, and M.~R. Costa-Jussa, ``End-to-end
  speech translation with the transformer.'' in \emph{IberSPEECH}, 2018, pp.
  60--63.

\bibitem{topal2021exploring}
M.~O. Topal, A.~Bas, and I.~van Heerden, ``Exploring transformers in natural
  language generation: Gpt, bert, and xlnet,'' \emph{arXiv preprint
  arXiv:2102.08036}, 2021.

\bibitem{graves2013speech}
A.~Graves, A.-r. Mohamed, and G.~Hinton, ``Speech recognition with deep
  recurrent neural networks,'' in \emph{2013 IEEE international conference on
  acoustics, speech and signal processing}.\hskip 1em plus 0.5em minus
  0.4em\relax Ieee, 2013, pp. 6645--6649.

\bibitem{sak2014long}
H.~Sak, A.~Senior, and F.~Beaufays, ``Long short-term memory based recurrent
  neural network architectures for large vocabulary speech recognition,''
  \emph{arXiv preprint arXiv:1402.1128}, 2014.

\bibitem{devlin2018bert}
J.~Devlin, M.-W. Chang, K.~Lee, and K.~Toutanova, ``Bert: Pre-training of deep
  bidirectional transformers for language understanding,'' \emph{arXiv preprint
  arXiv:1810.04805}, 2018.

\bibitem{radford2018improving}
A.~Radford, K.~Narasimhan, T.~Salimans, and I.~Sutskever, ``Improving language
  understanding by generative pre-training,'' \emph{cs.ubc.ca}, 2018.

\bibitem{brown2020language}
T.~B. Brown, B.~Mann, N.~Ryder, M.~Subbiah, J.~Kaplan, P.~Dhariwal,
  A.~Neelakantan, P.~Shyam, G.~Sastry, A.~Askell \emph{et~al.}, ``Language
  models are few-shot learners,'' \emph{arXiv preprint arXiv:2005.14165}, 2020.

\bibitem{shen2017deep}
D.~Shen, G.~Wu, and H.-I. Suk, ``Deep learning in medical image analysis,''
  \emph{Annual review of biomedical engineering}, vol.~19, pp. 221--248, 2017.

\bibitem{krizhevsky2012imagenet}
A.~Krizhevsky, I.~Sutskever, and G.~E. Hinton, ``Imagenet classification with
  deep convolutional neural networks,'' \emph{Advances in neural information
  processing systems}, vol.~25, pp. 1097--1105, 2012.

\bibitem{dosovitskiy2020vit}
A.~Dosovitskiy, L.~Beyer, A.~Kolesnikov, D.~Weissenborn, X.~Zhai,
  T.~Unterthiner, M.~Dehghani, M.~Minderer, G.~Heigold, S.~Gelly, J.~Uszkoreit,
  and N.~Houlsby, ``An image is worth 16x16 words: Transformers for image
  recognition at scale,'' \emph{ICLR}, 2021.

\bibitem{detr}
N.~Carion, F.~Massa, G.~Synnaeve, N.~Usunier, A.~Kirillov, and S.~Zagoruyko,
  ``End-to-end object detection with transformers,'' in \emph{European
  Conference on Computer Vision}.\hskip 1em plus 0.5em minus 0.4em\relax
  Springer, 2020, pp. 213--229.

\bibitem{zheng2021rethinking}
S.~Zheng, J.~Lu, H.~Zhao, X.~Zhu, Z.~Luo, Y.~Wang, Y.~Fu, J.~Feng, T.~Xiang,
  P.~H. Torr \emph{et~al.}, ``Rethinking semantic segmentation from a
  sequence-to-sequence perspective with transformers,'' in \emph{Proceedings of
  the IEEE/CVF Conference on Computer Vision and Pattern Recognition}, 2021,
  pp. 6881--6890.

\bibitem{parmar2018image}
N.~Parmar, A.~Vaswani, J.~Uszkoreit, L.~Kaiser, N.~Shazeer, A.~Ku, and D.~Tran,
  ``Image transformer,'' in \emph{International Conference on Machine
  Learning}.\hskip 1em plus 0.5em minus 0.4em\relax PMLR, 2018, pp. 4055--4064.

\bibitem{li2019entangled}
G.~Li, L.~Zhu, P.~Liu, and Y.~Yang, ``Entangled transformer for image
  captioning,'' in \emph{Proceedings of the IEEE/CVF International Conference
  on Computer Vision}, 2019, pp. 8928--8937.

\bibitem{zhou2018end}
L.~Zhou, Y.~Zhou, J.~J. Corso, R.~Socher, and C.~Xiong, ``End-to-end dense
  video captioning with masked transformer,'' in \emph{Proceedings of the IEEE
  Conference on Computer Vision and Pattern Recognition}, 2018, pp. 8739--8748.

\bibitem{PureTrans-COVID-VIT-gao2021covidvit}
X.~Gao, Y.~Qian, and A.~Gao, ``Covid-vit: Classification of covid-19 from ct
  chest images based on vision transformer models,'' 2021.

\bibitem{SwinTrans-MIA-COV19D-zhangmia}
L.~Zhang and Y.~Wen, ``Mia-cov19d: A transformer-based framework for covid19
  classification in chest cts,'' \emph{arXiv}, 2021.

\bibitem{GLTrans-he2021globallocal}
S.~He, P.~E. Grant, and Y.~Ou, ``Global-local transformer for brain age
  estimation,'' 2021.

\bibitem{PureTrans-BrazilViT-costa2021covid}
G.~S.~S. Costa, A.~C. Paiva, G.~B. Junior, and M.~M. Ferreira, ``Covid-19
  automatic diagnosis with ct images using the novel transformer
  architecture,'' in \emph{Anais do XXI Simp{\'o}sio Brasileiro de
  Computa{\c{c}}{\~a}o Aplicada {\`a} Sa{\'u}de}.\hskip 1em plus 0.5em minus
  0.4em\relax SBC, 2021, pp. 293--301.

\bibitem{CNN+Trans-Multi-View-van2021multi}
G.~van Tulder, Y.~Tong, and E.~Marchiori, ``Multi-view analysis of unregistered
  medical images using cross-view transformers,'' \emph{arXiv preprint
  arXiv:2103.11390}, 2021.

\bibitem{zhang2021pyramid}
Z.~Zhang, B.~Sun, and W.~Zhang, ``Pyramid medical transformer for medical image
  segmentation,'' \emph{arXiv preprint arXiv:2104.14702}, 2021.

\bibitem{xie2021cotr}
Y.~Xie, J.~Zhang, C.~Shen, and Y.~Xia, ``Cotr: Efficiently bridging cnn and
  transformer for 3d medical image segmentation,'' \emph{arXiv preprint
  arXiv:2103.03024}, 2021.

\bibitem{watanabe2021generative}
S.~Watanabe, T.~Ueno, Y.~Kimura, M.~Mishina, and N.~Sugimoto, ``Generative
  image transformer (git): unsupervised continuous image generative and
  transformable model for [123i] fp-cit spect images,'' \emph{Annals of Nuclear
  Medicine}, vol.~35, no.~11, pp. 1203--1213, 2021.

\bibitem{bahdanau_neural_2016}
\BIBentryALTinterwordspacing
D.~Bahdanau, K.~Cho, and Y.~Bengio, ``Neural {Machine} {Translation} by
  {Jointly} {Learning} to {Align} and {Translate},'' \emph{arXiv:1409.0473 [cs,
  stat]}, May 2016, arXiv: 1409.0473. [Online]. Available:
  \url{http://arxiv.org/abs/1409.0473}
\BIBentrySTDinterwordspacing

\bibitem{hu_squeeze-and-excitation_2019}
\BIBentryALTinterwordspacing
J.~Hu, L.~Shen, S.~Albanie, G.~Sun, and E.~Wu, ``Squeeze-and-{Excitation}
  {Networks},'' \emph{arXiv:1709.01507 [cs]}, May 2019, arXiv: 1709.01507.
  [Online]. Available: \url{http://arxiv.org/abs/1709.01507}
\BIBentrySTDinterwordspacing

\bibitem{chen2021vit}
J.~Chen, Y.~He, E.~C. Frey, Y.~Li, and Y.~Du, ``Vit-v-net: Vision transformer
  for unsupervised volumetric medical image registration,'' \emph{arXiv
  preprint arXiv:2104.06468}, 2021.

\bibitem{Data-Efficient-Deit}
H.~Touvron, M.~Cord, M.~Douze, F.~Massa, A.~Sablayrolles, and H.~Jégou,
  ``Training data-efficient image transformers \& distillation through
  attention,'' 2021.

\bibitem{liu2021swin}
Z.~Liu, Y.~Lin, Y.~Cao, H.~Hu, Y.~Wei, Z.~Zhang, S.~Lin, and B.~Guo, ``Swin
  transformer: Hierarchical vision transformer using shifted windows,''
  \emph{arXiv preprint arXiv:2103.14030}, 2021.

\bibitem{Tolstikhin2021mlp-mixer}
I.~Tolstikhin, N.~Houlsby, A.~Kolesnikov, L.~Beyer, X.~Zhai, T.~Unterthiner,
  J.~Yung, D.~Keysers, J.~Uszkoreit, M.~Lucic \emph{et~al.}, ``Mlp-mixer: An
  all-mlp architecture for vision,'' \emph{arXiv preprint arXiv:2105.01601},
  2021.

\bibitem{liu202gmlp}
H.~Liu, Z.~Dai, D.~R. So, and Q.~V. Le, ``Pay attention to mlps,'' 2021.

\bibitem{Touvron2021resmlp}
H.~Touvron, P.~Bojanowski, M.~Caron, M.~Cord, A.~El-Nouby, E.~Grave,
  G.~Izacard, A.~Joulin, G.~Synnaeve, J.~Verbeek \emph{et~al.}, ``Resmlp:
  Feedforward networks for image classification with data-efficient training,''
  \emph{arXiv preprint arXiv:2105.03404}, 2021.

\bibitem{lian2021asmlp}
D.~Lian, Z.~Yu, X.~Sun, and S.~Gao, ``As-mlp: An axial shifted mlp architecture
  for vision,'' \emph{arXiv preprint arXiv:2107.08391}, 2021.

\bibitem{Chen2021cyclemlp}
S.~Chen, E.~Xie, C.~Ge, D.~Liang, and P.~Luo, ``Cyclemlp: A mlp-like
  architecture for dense prediction,'' \emph{arXiv preprint arXiv:2107.10224},
  2021.

\bibitem{CNN+Trans-TransHybrid-liang2021hybrid}
S.~Liang, ``A hybrid deep learning framework for covid-19 detection via 3d
  chest ct images,'' \emph{arXiv preprint arXiv:2107.03904}, 2021.

\bibitem{CNN+Trans-Scopeformer-barhoumi2021scopeformer}
Y.~Barhoumi and R.~Ghulam, ``Scopeformer: n-cnn-vit hybrid model for
  intracranial hemorrhage classification,'' 2021.

\bibitem{li2021medical}
J.~Li, Z.~Yang, and Y.~Yu, ``A medical ai diagnosis platform based on vision
  transformer for coronavirus,'' in \emph{2021 IEEE International Conference on
  Computer Science, Electronic Information Engineering and Intelligent Control
  Technology (CEI)}.\hskip 1em plus 0.5em minus 0.4em\relax IEEE, 2021, pp.
  246--252.

\bibitem{than2021preliminary}
J.~C. Than, P.~L. Thon, O.~M. Rijal, R.~M. Kassim, A.~Yunus, N.~M. Noor, and
  P.~Then, ``Preliminary study on patch sizes in vision transformers (vit) for
  covid-19 and diseased lungs classification,'' in \emph{2021 IEEE National
  Biomedical Engineering Conference (NBEC)}.\hskip 1em plus 0.5em minus
  0.4em\relax IEEE, 2021, pp. 146--150.

\bibitem{rahimzadeh2021fully}
M.~Rahimzadeh, A.~Attar, and S.~M. Sakhaei, ``A fully automated deep
  learning-based network for detecting covid-19 from a new and large lung ct
  scan dataset,'' \emph{Biomedical Signal Processing and Control}, vol.~68, p.
  102588, 2021.

\bibitem{xia2021effective}
Y.~Xia, J.~Yao, L.~Lu, L.~Huang, G.~Xie, J.~Xiao, A.~Yuille, K.~Cao, and
  L.~Zhang, ``Effective pancreatic cancer screening on non-contrast ct scans
  via anatomy-aware transformers,'' in \emph{International Conference on
  Medical Image Computing and Computer-Assisted Intervention}.\hskip 1em plus
  0.5em minus 0.4em\relax Springer, 2021, pp. 259--269.

\bibitem{Pre-training-COVID-19-park2021vision}
S.~Park, G.~Kim, Y.~Oh, J.~B. Seo, S.~M. Lee, J.~H. Kim, S.~Moon, J.-K. Lim,
  and J.~C. Ye, ``Vision transformer for covid-19 cxr diagnosis using chest
  x-ray feature corpus,'' \emph{arXiv preprint arXiv:2103.07055}, 2021.

\bibitem{tanzi-femur-fracture-x-ray}
L.~Tanzi, A.~Audisio, G.~Cirrincione, A.~Aprato, and E.~Vezzetti, ``Vision
  transformer for femur fracture classification,'' 2021.

\bibitem{CNN+Trans-MLL-ATT-verenich2021pulmonary}
E.~Verenich, T.~Martin, A.~Velasquez, N.~Khan, and F.~Hussain, ``Pulmonary
  disease classification using globally correlated maximum likelihood: an
  auxiliary attention mechanism for convolutional neural networks,''
  \emph{arXiv preprint arXiv:2109.00573}, 2021.

\bibitem{dataset19-https://doi.org/10.48550/arxiv.2003.11597}
\BIBentryALTinterwordspacing
J.~P. Cohen, P.~Morrison, and L.~Dao, ``Covid-19 image data collection,'' 2020.
  [Online]. Available: \url{https://arxiv.org/abs/2003.11597}
\BIBentrySTDinterwordspacing

\bibitem{covid19-database}
M.~E.~H. Chowdhury, T.~Rahman, A.~Khandakar, R.~Mazhar, M.~A. Kadir, Z.~B.
  Mahbub, K.~R. Islam, M.~S. Khan, A.~Iqbal, N.~A. Emadi, M.~B.~I. Reaz, and
  M.~T. Islam, ``Can ai help in screening viral and covid-19 pneumonia?''
  \emph{IEEE Access}, vol.~8, pp. 132\,665--132\,676, 2020.

\bibitem{PureTrans-VOLO-liu2021automatic}
C.~Liu and Q.~Yin, ``Automatic diagnosis of covid-19 using a tailored
  transformer-like network,'' in \emph{Journal of Physics: Conference Series},
  vol. 2010, no.~1.\hskip 1em plus 0.5em minus 0.4em\relax IOP Publishing,
  2021, p. 012175.

\bibitem{cohen2020covid19}
J.~P. Cohen, P.~Morrison, L.~Dao, K.~Roth, T.~Q. Duong, and M.~Ghassemi,
  ``Covid-19 image data collection: Prospective predictions are the future,''
  2020.

\bibitem{shome2021covid}
D.~Shome, T.~Kar, S.~N. Mohanty, P.~Tiwari, K.~Muhammad, A.~AlTameem, Y.~Zhang,
  and A.~K.~J. Saudagar, ``Covid-transformer: Interpretable covid-19 detection
  using vision transformer for healthcare,'' \emph{International Journal of
  Environmental Research and Public Health}, vol.~18, no.~21, p. 11086, 2021.

\bibitem{krishnan2021vision}
K.~S. Krishnan and K.~S. Krishnan, ``Vision transformer based covid-19
  detection using chest x-rays,'' in \emph{2021 6th International Conference on
  Signal Processing, Computing and Control (ISPCC)}.\hskip 1em plus 0.5em minus
  0.4em\relax IEEE, 2021, pp. 644--648.

\bibitem{GraphLearning-DynamicGraph-kim2021learning}
B.-H. Kim, J.~C. Ye, and J.-J. Kim, ``Learning dynamic graph representation of
  brain connectome with spatio-temporal attention,'' 2021.

\bibitem{CNN+Trans-mfTrans-zhao2021mftrans}
J.~Zhao, X.~Xiao, D.~Li, J.~Chong, Z.~Kassam, B.~Chen, and S.~Li,
  ``mftrans-net: Quantitative measurement of hepatocellular carcinoma via
  multi-function transformer regression network,'' in \emph{International
  Conference on Medical Image Computing and Computer-Assisted
  Intervention}.\hskip 1em plus 0.5em minus 0.4em\relax Springer, 2021, pp.
  75--84.

\bibitem{wang20213dmet}
S.~Wang, Z.~Zhuang, K.~Xuan, D.~Qian, Z.~Xue, J.~Xu, Y.~Liu, Y.~Chai, L.~Zhang,
  Q.~Wang \emph{et~al.}, ``3dmet: 3d medical image transformer for knee
  cartilage defect assessment,'' in \emph{International Workshop on Machine
  Learning in Medical Imaging}.\hskip 1em plus 0.5em minus 0.4em\relax
  Springer, 2021, pp. 347--355.

\bibitem{CNN+Trans-i-ViT-gao2021instance}
Z.~Gao, B.~Hong, X.~Zhang, Y.~Li, C.~Jia, J.~Wu, C.~Wang, D.~Meng, and C.~Li,
  ``Instance-based vision transformer for subtyping of papillary renal cell
  carcinoma in histopathological image,'' in \emph{International Conference on
  Medical Image Computing and Computer-Assisted Intervention}.\hskip 1em plus
  0.5em minus 0.4em\relax Springer, 2021, pp. 299--308.

\bibitem{CNN+Trans-GasHis-Transformer-chen2021gashis}
H.~Chen, C.~Li, X.~Li, G.~Wang, W.~Hu, Y.~Li, W.~Liu, C.~Sun, Y.~Yao, Y.~Teng
  \emph{et~al.}, ``Gashis-transformer: A multi-scale visual transformer
  approach for gastric histopathology image classification,'' \emph{arXiv
  preprint arXiv:2104.14528}, 2021.

\bibitem{zeid2021multiclass}
M.~A.-E. Zeid, K.~El-Bahnasy, and S.~Abo-Youssef, ``Multiclass colorectal
  cancer histology images classification using vision transformers,'' in
  \emph{2021 Tenth International Conference on Intelligent Computing and
  Information Systems (ICICIS)}.\hskip 1em plus 0.5em minus 0.4em\relax IEEE,
  2021, pp. 224--230.

\bibitem{janowczyk2016deep}
A.~Janowczyk and A.~Madabhushi, ``Deep learning for digital pathology image
  analysis: A comprehensive tutorial with selected use cases,'' \emph{Journal
  of pathology informatics}, vol.~7, 2016.

\bibitem{ikromjanov2022whole}
K.~Ikromjanov, S.~Bhattacharjee, Y.-B. Hwang, R.~I. Sumon, H.-C. Kim, and H.-K.
  Choi, ``Whole slide image analysis and detection of prostate cancer using
  vision transformers,'' in \emph{2022 International Conference on Artificial
  Intelligence in Information and Communication (ICAIIC)}.\hskip 1em plus 0.5em
  minus 0.4em\relax IEEE, 2022, pp. 399--402.

\bibitem{zhao2022improving}
C.~Zhao, R.~Shuai, L.~Ma, W.~Liu, and M.~Wu, ``Improving cervical cancer
  classification with imbalanced datasets combining taming transformers with
  t2t-vit,'' \emph{Multimedia tools and applications}, pp. 1--36, 2022.

\bibitem{LightTrans-USNet-perera2021pocformer}
S.~Perera, S.~Adhikari, and A.~Yilmaz, ``Pocformer: A lightweight transformer
  architecture for detection of covid-19 using point of care ultrasound,''
  \emph{arXiv preprint arXiv:2105.09913}, 2021.

\bibitem{gheflati2021vision}
B.~Gheflati and H.~Rivaz, ``Vision transformer for classification of breast
  ultrasound images,'' \emph{arXiv preprint arXiv:2110.14731}, 2021.

\bibitem{BUSI}
\BIBentryALTinterwordspacing
W.~Al-Dhabyani, M.~Gomaa, H.~Khaled, and A.~Fahmy, ``Dataset of breast
  ultrasound images,'' \emph{Data in Brief}, vol.~28, p. 104863, 2020.
  [Online]. Available:
  \url{https://www.sciencedirect.com/science/article/pii/S2352340919312181}
\BIBentrySTDinterwordspacing

\bibitem{Dataset-B}
M.~H. Yap, G.~Pons, J.~Martí, S.~Ganau, M.~Sentís, R.~Zwiggelaar, A.~K.
  Davison, and R.~Martí, ``Automated breast ultrasound lesions detection using
  convolutional neural networks,'' \emph{IEEE Journal of Biomedical and Health
  Informatics}, vol.~22, no.~4, pp. 1218--1226, 2018.

\bibitem{ViT+CNN+Ensemble-ViT-CNN-jiang2021method}
Z.~Jiang, Z.~Dong, L.~Wang, and W.~Jiang, ``Method for diagnosis of acute
  lymphoblastic leukemia based on vit-cnn ensemble model,'' \emph{Computational
  Intelligence and Neuroscience}, vol. 2021, 2021.

\bibitem{melane-skin-Xie}
J.~Xie, Z.~Wu, R.~Zhu, and H.~Zhu, ``Melanoma detection based on swin
  transformer and simam,'' in \emph{2021 IEEE 5th Information
  Technology,Networking,Electronic and Automation Control Conference (ITNEC)},
  vol.~5, 2021, pp. 1517--1521.

\bibitem{li2021out}
X.~Li, C.~Desrosiers, and X.~Liu, ``Out-of-distribution detection using vision
  transformers,'' 2021.

\bibitem{yu2021end}
Z.~Yu, V.~Mar, A.~Eriksson, S.~Chandra, P.~Bonnington, L.~Zhang, and Z.~Ge,
  ``End-to-end ugly duckling sign detection for melanoma identification with
  transformers,'' in \emph{International Conference on Medical Image Computing
  and Computer-Assisted Intervention}.\hskip 1em plus 0.5em minus 0.4em\relax
  Springer, 2021, pp. 176--184.

\bibitem{wu2021scale}
W.~Wu, S.~Mehta, S.~Nofallah, S.~Knezevich, C.~J. May, O.~H. Chang, J.~G.
  Elmore, and L.~G. Shapiro, ``Scale-aware transformers for diagnosing
  melanocytic lesions,'' \emph{IEEE Access}, vol.~9, pp. 163\,526--163\,541,
  2021.

\bibitem{yang2021fundus}
H.~Yang, J.~Chen, and M.~Xu, ``Fundus disease image classification based on
  improved transformer,'' in \emph{2021 International Conference on
  Neuromorphic Computing (ICNC)}.\hskip 1em plus 0.5em minus 0.4em\relax IEEE,
  2021, pp. 207--214.

\bibitem{Pure-Trans-Eye-song2021deep}
D.~Song, B.~Fu, F.~Li, J.~Xiong, J.~He, X.~Zhang, and Y.~Qiao, ``Deep relation
  transformer for diagnosing glaucoma with optical coherence tomography and
  visual field function,'' \emph{IEEE Transactions on Medical Imaging}, 2021.

\bibitem{yuan2021volo}
L.~Yuan, Q.~Hou, Z.~Jiang, J.~Feng, and S.~Yan, ``Volo: Vision outlooker for
  visual recognition,'' \emph{arXiv preprint arXiv:2106.13112}, 2021.

\bibitem{selvaraju2017grad}
R.~R. Selvaraju, M.~Cogswell, A.~Das, R.~Vedantam, D.~Parikh, and D.~Batra,
  ``Grad-cam: Visual explanations from deep networks via gradient-based
  localization,'' in \emph{Proceedings of the IEEE international conference on
  computer vision}, 2017, pp. 618--626.

\bibitem{aldahoul2021encoding}
N.~AlDahoul, H.~A. Karim, M.~J.~T. Tan, M.~A. Momo, and J.~L. Fermin,
  ``Encoding retina image to words using ensemble of vision transformers for
  diabetic retinopathy grading,'' \emph{F1000Research}, vol.~10, no. 948, p.
  948, 2021.

\bibitem{Graph-Representation-Learning-Survey}
\BIBentryALTinterwordspacing
F.~Chen, Y.-C. Wang, B.~Wang, and C.-C.~J. Kuo, ``Graph representation
  learning: a survey,'' \emph{APSIPA Transactions on Signal and Information
  Processing}, vol.~9, 2020. [Online]. Available:
  \url{http://dx.doi.org/10.1017/ATSIP.2020.13}
\BIBentrySTDinterwordspacing

\bibitem{Brain-Network-Analysis-liu2017complex}
J.~Liu, M.~Li, Y.~Pan, W.~Lan, R.~Zheng, F.-X. Wu, and J.~Wang, ``Complex brain
  network analysis and its applications to brain disorders: a survey,''
  \emph{Complexity}, vol. 2017, 2017.

\bibitem{Bessadok2021}
A.~Bessadok, M.~A. Mahjoub, and I.~Rekik, ``Graph neural networks in network
  neuroscience,'' \emph{arXiv preprint arXiv:2106.03535}, 2021.

\bibitem{jha2020kvasir}
D.~Jha, P.~H. Smedsrud, M.~A. Riegler, P.~Halvorsen, T.~de~Lange, D.~Johansen,
  and H.~D. Johansen, ``Kvasir-seg: A segmented polyp dataset,'' in
  \emph{International Conference on Multimedia Modeling}.\hskip 1em plus 0.5em
  minus 0.4em\relax Springer, 2020, pp. 451--462.

\bibitem{bernal2015wm}
J.~Bernal, F.~J. S{\'a}nchez, G.~Fern{\'a}ndez-Esparrach, D.~Gil,
  C.~Rodr{\'\i}guez, and F.~Vilari{\~n}o, ``Wm-dova maps for accurate polyp
  highlighting in colonoscopy: Validation vs. saliency maps from physicians,''
  \emph{Computerized Medical Imaging and Graphics}, vol.~43, pp. 99--111, 2015.

\bibitem{tajbakhsh2015automated}
N.~Tajbakhsh, S.~R. Gurudu, and J.~Liang, ``Automated polyp detection in
  colonoscopy videos using shape and context information,'' \emph{IEEE
  transactions on medical imaging}, vol.~35, no.~2, pp. 630--644, 2015.

\bibitem{vazquez2017benchmark}
D.~V{\'a}zquez, J.~Bernal, F.~J. S{\'a}nchez, G.~Fern{\'a}ndez-Esparrach, A.~M.
  L{\'o}pez, A.~Romero, M.~Drozdzal, and A.~Courville, ``A benchmark for
  endoluminal scene segmentation of colonoscopy images,'' \emph{Journal of
  healthcare engineering}, vol. 2017, 2017.

\bibitem{silva2014toward}
J.~Silva, A.~Histace, O.~Romain, X.~Dray, and B.~Granado, ``Toward embedded
  detection of polyps in wce images for early diagnosis of colorectal cancer,''
  \emph{International journal of computer assisted radiology and surgery},
  vol.~9, no.~2, pp. 283--293, 2014.

\bibitem{chen2021transunet}
J.~Chen, Y.~Lu, Q.~Yu, X.~Luo, E.~Adeli, Y.~Wang, L.~Lu, A.~L. Yuille, and
  Y.~Zhou, ``Transunet: Transformers make strong encoders for medical image
  segmentation,'' \emph{arXiv preprint arXiv:2102.04306}, 2021.

\bibitem{chang2021transclaw}
Y.~Chang, H.~Menghan, Z.~Guangtao, and Z.~Xiao-Ping, ``Transclaw u-net: Claw
  u-net with transformers for medical image segmentation,'' \emph{arXiv
  preprint arXiv:2107.05188}, 2021.

\bibitem{xu2021levit}
G.~Xu, X.~Wu, X.~Zhang, and X.~He, ``Levit-unet: Make faster encoders with
  transformer for medical image segmentation,'' \emph{arXiv preprint
  arXiv:2107.08623}, 2021.

\bibitem{sha2021Transformer}
Y.~Sha, Y.~Zhang, X.~Ji, and L.~Hu, ``Transformer-unet: Raw image processing
  with unet,'' \emph{arXiv preprint arXiv:2109.08417}, 2021.

\bibitem{li2021more}
Y.~Li, W.~Cai, Y.~Gao, and X.~Hu, ``More than encoder: Introducing transformer
  decoder to upsample,'' \emph{arXiv preprint arXiv:2106.10637}, 2021.

\bibitem{simpson2019large}
A.~L. Simpson, M.~Antonelli, S.~Bakas, M.~Bilello, K.~Farahani,
  B.~Van~Ginneken, A.~Kopp-Schneider, B.~A. Landman, G.~Litjens, B.~Menze
  \emph{et~al.}, ``A large annotated medical image dataset for the development
  and evaluation of segmentation algorithms,'' \emph{arXiv preprint
  arXiv:1902.09063}, 2019.

\bibitem{gao2021utnet}
Y.~Gao, M.~Zhou, and D.~N. Metaxas, ``Utnet: a hybrid transformer architecture
  for medical image segmentation,'' in \emph{International Conference on
  Medical Image Computing and Computer-Assisted Intervention}.\hskip 1em plus
  0.5em minus 0.4em\relax Springer, 2021, pp. 61--71.

\bibitem{campello2021multi}
V.~M. Campello, P.~Gkontra, C.~Izquierdo, C.~Mart{\'\i}n-Isla, A.~Sojoudi,
  P.~M. Full, K.~Maier-Hein, Y.~Zhang, Z.~He, J.~Ma \emph{et~al.},
  ``Multi-centre, multi-vendor and multi-disease cardiac segmentation: the
  m\&ms challenge,'' \emph{IEEE Transactions on Medical Imaging}, vol.~40,
  no.~12, pp. 3543--3554, 2021.

\bibitem{fu2022tf}
Z.~Fu, J.~Zhang, R.~Luo, Y.~Sun, D.~Deng, and L.~Xia, ``Tf-unet: An automatic
  cardiac mri image segmentation method,'' \emph{Mathematical Biosciences and
  Engineering}, vol.~19, no.~5, pp. 5207--5222, 2022.

\bibitem{gao2022multi}
Y.~Gao, M.~Zhou, D.~Liu, and D.~Metaxas, ``A multi-scale transformer for
  medical image segmentation: Architectures, model efficiency, and
  benchmarks,'' \emph{arXiv preprint arXiv:2203.00131}, 2022.

\bibitem{sun2021hybridctrm}
Q.~Sun, N.~Fang, Z.~Liu, L.~Zhao, Y.~Wen, and H.~Lin, ``Hybridctrm: Bridging
  cnn and transformer for multimodal brain image segmentation,'' \emph{Journal
  of Healthcare Engineering}, vol. 2021, 2021.

\bibitem{mendrik2015mrbrains}
A.~M. Mendrik, K.~L. Vincken, H.~J. Kuijf, M.~Breeuwer, W.~H. Bouvy,
  J.~De~Bresser, A.~Alansary, M.~De~Bruijne, A.~Carass, A.~El-Baz
  \emph{et~al.}, ``Mrbrains challenge: online evaluation framework for brain
  image segmentation in 3t mri scans,'' \emph{Computational intelligence and
  neuroscience}, vol. 2015, 2015.

\bibitem{wang2019benchmark}
L.~Wang, D.~Nie, G.~Li, {\'E}.~Puybareau, J.~Dolz, Q.~Zhang, F.~Wang, J.~Xia,
  Z.~Wu, J.-W. Chen \emph{et~al.}, ``Benchmark on automatic six-month-old
  infant brain segmentation algorithms: the iseg-2017 challenge,'' \emph{IEEE
  transactions on medical imaging}, vol.~38, no.~9, pp. 2219--2230, 2019.

\bibitem{zhang2021transfuse}
Y.~Zhang, H.~Liu, and Q.~Hu, ``Transfuse: Fusing transformers and cnns for
  medical image segmentation,'' \emph{arXiv preprint arXiv:2102.08005}, 2021.

\bibitem{codella2018skin}
N.~C. Codella, D.~Gutman, M.~E. Celebi, B.~Helba, M.~A. Marchetti, S.~W. Dusza,
  A.~Kalloo, K.~Liopyris, N.~Mishra, H.~Kittler \emph{et~al.}, ``Skin lesion
  analysis toward melanoma detection: A challenge at the 2017 international
  symposium on biomedical imaging (isbi), hosted by the international skin
  imaging collaboration (isic),'' in \emph{2018 IEEE 15th international
  symposium on biomedical imaging (ISBI 2018)}.\hskip 1em plus 0.5em minus
  0.4em\relax IEEE, 2018, pp. 168--172.

\bibitem{you2022class}
C.~You, R.~Zhao, F.~Liu, S.~Chinchali, U.~Topcu, L.~Staib, and J.~S. Duncan,
  ``Class-aware generative adversarial transformers for medical image
  segmentation,'' \emph{arXiv preprint arXiv:2201.10737}, 2022.

\bibitem{xu2021ect}
S.~Xu and H.~Quan, ``Ect-nas: Searching efficient cnn-transformers architecture
  for medical image segmentation,'' in \emph{2021 IEEE International Conference
  on Bioinformatics and Biomedicine (BIBM)}.\hskip 1em plus 0.5em minus
  0.4em\relax IEEE, 2021, pp. 1601--1604.

\bibitem{luo2021hybrid}
H.~Luo, Y.~Changdong, and R.~Selvan, ``Hybrid ladder transformers with
  efficient parallel-cross attention for medical image segmentation,'' 2021.

\bibitem{liu2022phtrans}
W.~Liu, T.~Tian, W.~Xu, H.~Yang, and X.~Pan, ``Phtrans: Parallelly aggregating
  global and local representations for medical image segmentation,''
  \emph{arXiv preprint arXiv:2203.04568}, 2022.

\bibitem{zhou2021nnformer}
H.-Y. Zhou, J.~Guo, Y.~Zhang, L.~Yu, L.~Wang, and Y.~Yu, ``nnformer:
  Interleaved transformer for volumetric segmentation,'' \emph{arXiv preprint
  arXiv:2109.03201}, 2021.

\bibitem{sirinukunwattana2017gland}
K.~Sirinukunwattana, J.~P. Pluim, H.~Chen, X.~Qi, P.-A. Heng, Y.~B. Guo, L.~Y.
  Wang, B.~J. Matuszewski, E.~Bruni, U.~Sanchez \emph{et~al.}, ``Gland
  segmentation in colon histology images: The glas challenge contest,''
  \emph{Medical image analysis}, vol.~35, pp. 489--502, 2017.

\bibitem{kumar2017dataset}
N.~Kumar, R.~Verma, S.~Sharma, S.~Bhargava, A.~Vahadane, and A.~Sethi, ``A
  dataset and a technique for generalized nuclear segmentation for
  computational pathology,'' \emph{IEEE transactions on medical imaging},
  vol.~36, no.~7, pp. 1550--1560, 2017.

\bibitem{valanarasu2021medical}
J.~M.~J. Valanarasu, P.~Oza, I.~Hacihaliloglu, and V.~M. Patel, ``Medical
  transformer: Gated axial-attention for medical image segmentation,''
  \emph{arXiv preprint arXiv:2102.10662}, 2021.

\bibitem{valanarasu2020learning}
J.~M.~J. Valanarasu, R.~Yasarla, P.~Wang, I.~Hacihaliloglu, and V.~M. Patel,
  ``Learning to segment brain anatomy from 2d ultrasound with less data,''
  \emph{IEEE Journal of Selected Topics in Signal Processing}, vol.~14, no.~6,
  pp. 1221--1234, 2020.

\bibitem{landman20152015}
B.~Landman, Z.~Xu, J.~Igelsias, M.~Styner, T.~Langerak, and A.~Klein, ``2015
  miccai multi-atlas labeling beyond the cranial vault workshop and
  challenge,'' \emph{web}, 2015.

\bibitem{ji2021multi}
Y.~Ji, R.~Zhang, H.~Wang, Z.~Li, L.~Wu, S.~Zhang, and P.~Luo, ``Multi-compound
  transformer for accurate biomedical image segmentation,'' in
  \emph{International Conference on Medical Image Computing and
  Computer-Assisted Intervention}.\hskip 1em plus 0.5em minus 0.4em\relax
  Springer, 2021, pp. 326--336.

\bibitem{gamper2019pannuke}
J.~Gamper, N.~A. Koohbanani, K.~Benet, A.~Khuram, and N.~Rajpoot, ``Pannuke: an
  open pan-cancer histology dataset for nuclei instance segmentation and
  classification,'' in \emph{European Congress on Digital Pathology}.\hskip 1em
  plus 0.5em minus 0.4em\relax Springer, 2019, pp. 11--19.

\bibitem{codella2019skin}
N.~Codella, V.~Rotemberg, P.~Tschandl, M.~E. Celebi, S.~Dusza, D.~Gutman,
  B.~Helba, A.~Kalloo, K.~Liopyris, M.~Marchetti \emph{et~al.}, ``Skin lesion
  analysis toward melanoma detection 2018: A challenge hosted by the
  international skin imaging collaboration (isic),'' \emph{arXiv preprint
  arXiv:1902.03368}, 2019.

\bibitem{wu2022d}
Y.~Wu, K.~Liao, J.~Chen, D.~Z. Chen, J.~Wang, H.~Gao, and J.~Wu, ``D-former: A
  u-shaped dilated transformer for 3d medical image segmentation,'' \emph{arXiv
  preprint arXiv:2201.00462}, 2022.

\bibitem{hatamizadeh2021unetr}
A.~Hatamizadeh, D.~Yang, H.~Roth, and D.~Xu, ``Unetr: Transformers for 3d
  medical image segmentation,'' \emph{arXiv preprint arXiv:2103.10504}, 2021.

\bibitem{hatamizadeh2022swin}
A.~Hatamizadeh, V.~Nath, Y.~Tang, D.~Yang, H.~Roth, and D.~Xu, ``Swin unetr:
  Swin transformers for semantic segmentation of brain tumors in mri images,''
  \emph{arXiv preprint arXiv:2201.01266}, 2022.

\bibitem{chen2021transattunet}
B.~Chen, Y.~Liu, Z.~Zhang, G.~Lu, and D.~Zhang, ``Transattunet: Multi-level
  attention-guided u-net with transformer for medical image segmentation,''
  \emph{arXiv preprint arXiv:2107.05274}, 2021.

\bibitem{tang2019xlsor}
Y.-B. Tang, Y.-X. Tang, J.~Xiao, and R.~M. Summers, ``Xlsor: A robust and
  accurate lung segmentor on chest x-rays using criss-cross attention and
  customized radiorealistic abnormalities generation,'' in \emph{International
  Conference on Medical Imaging with Deep Learning}.\hskip 1em plus 0.5em minus
  0.4em\relax PMLR, 2019, pp. 457--467.

\bibitem{he2020benchmarking}
X.~He, S.~Wang, S.~Shi, X.~Chu, J.~Tang, X.~Liu, C.~Yan, J.~Zhang, and G.~Ding,
  ``Benchmarking deep learning models and automated model design for covid-19
  detection with chest ct scans,'' \emph{medRxiv}, 2020.

\bibitem{caicedo2019nucleus}
J.~C. Caicedo, A.~Goodman, K.~W. Karhohs, B.~A. Cimini, J.~Ackerman,
  M.~Haghighi, C.~Heng, T.~Becker, M.~Doan, C.~McQuin \emph{et~al.}, ``Nucleus
  segmentation across imaging experiments: the 2018 data science bowl,''
  \emph{Nature methods}, vol.~16, no.~12, pp. 1247--1253, 2019.

\bibitem{wang2021mixed}
H.~Wang, S.~Xie, L.~Lin, Y.~Iwamoto, X.-H. Han, Y.-W. Chen, and R.~Tong,
  ``Mixed transformer u-net for medical image segmentation,'' \emph{arXiv
  preprint arXiv:2111.04734}, 2021.

\bibitem{yan2021after}
X.~Yan, H.~Tang, S.~Sun, H.~Ma, D.~Kong, and X.~Xie, ``After-unet: Axial fusion
  transformer unet for medical image segmentation,'' \emph{arXiv preprint
  arXiv:2110.10403}, 2021.

\bibitem{chen2021deep}
X.~Chen, S.~Sun, N.~Bai, K.~Han, Q.~Liu, S.~Yao, H.~Tang, C.~Zhang, Z.~Lu,
  Q.~Huang \emph{et~al.}, ``A deep learning-based auto-segmentation system for
  organs-at-risk on whole-body computed tomography images for radiation
  therapy,'' \emph{Radiotherapy and Oncology}, vol. 160, pp. 175--184, 2021.

\bibitem{lambert2020segthor}
Z.~Lambert, C.~Petitjean, B.~Dubray, and S.~Kuan, ``Segthor: Segmentation of
  thoracic organs at risk in ct images,'' in \emph{2020 Tenth International
  Conference on Image Processing Theory, Tools and Applications (IPTA)}.\hskip
  1em plus 0.5em minus 0.4em\relax IEEE, 2020, pp. 1--6.

\bibitem{zhang3992963s2wintounet}
J.~Zhang, Y.~Liu, Q.~Wu, Y.~Liu, Y.~Wang, X.~Xu, and B.~Song, ``S2wintounet:
  Star-shaped window transformer onion u-net for medical image segmentation,''
  \emph{Available at SSRN 3992963}.

\bibitem{karimi2021convolution}
D.~Karimi, S.~Vasylechko, and A.~Gholipour, ``Convolution-free medical image
  segmentation using transformers,'' \emph{arXiv preprint arXiv:2102.13645},
  2021.

\bibitem{cao2021swin}
H.~Cao, Y.~Wang, J.~Chen, D.~Jiang, X.~Zhang, Q.~Tian, and M.~Wang,
  ``Swin-unet: Unet-like pure transformer for medical image segmentation,''
  \emph{arXiv preprint arXiv:2105.05537}, 2021.

\bibitem{lin2021ds}
A.~Lin, B.~Chen, J.~Xu, Z.~Zhang, and G.~Lu, ``Ds-transunet: Dual swin
  transformer u-net for medical image segmentation,'' \emph{arXiv preprint
  arXiv:2106.06716}, 2021.

\bibitem{huang2021missformer}
X.~Huang, Z.~Deng, D.~Li, and X.~Yuan, ``Missformer: An effective medical image
  segmentation transformer,'' \emph{arXiv preprint arXiv:2109.07162}, 2021.

\bibitem{ning2021cac}
Y.~Ning, S.~Zhang, X.~Xi, J.~Guo, P.~Liu, and C.~Zhang, ``Cac-emvt: Efficient
  coronary artery calcium segmentation with multi-scale vision transformers,''
  in \emph{2021 IEEE International Conference on Bioinformatics and Biomedicine
  (BIBM)}.\hskip 1em plus 0.5em minus 0.4em\relax IEEE, 2021, pp. 1462--1467.

\bibitem{wang2021transbts}
W.~Wang, C.~Chen, M.~Ding, J.~Li, H.~Yu, and S.~Zha, ``Transbts: Multimodal
  brain tumor segmentation using transformer,'' \emph{arXiv preprint
  arXiv:2103.04430}, 2021.

\bibitem{Self-Pre-training-BrainEncoder-jun2021medical}
E.~Jun, S.~Jeong, D.-W. Heo, and H.-I. Suk, ``Medical transformer: Universal
  brain encoder for 3d mri analysis,'' \emph{arXiv preprint arXiv:2104.13633},
  2021.

\bibitem{ranem2022continual}
A.~Ranem, C.~Gonz{\'a}lez, and A.~Mukhopadhyay, ``Continual hippocampus
  segmentation with transformers,'' \emph{arXiv preprint arXiv:2204.08043},
  2022.

\bibitem{laiton2022deep}
C.~Laiton-Bonadiez, G.~Sanchez-Torres, and J.~Branch-Bedoya, ``Deep 3d neural
  network for brain structures segmentation using self-attention modules in mri
  images,'' \emph{Sensors}, vol.~22, no.~7, p. 2559, 2022.

\bibitem{rao2022improving}
V.~M. Rao, Z.~Wan, D.~J. Ma, P.-Y. Lee, Y.~Tian, A.~F. Laine, and J.~Guo,
  ``Improving across-dataset brain tissue segmentation using transformer,''
  \emph{arXiv preprint arXiv:2201.08741}, 2022.

\bibitem{liang2022transconver}
J.~Liang, C.~Yang, M.~Zeng, and X.~Wang, ``Transconver: transformer and
  convolution parallel network for developing automatic brain tumor
  segmentation in mri images,'' \emph{Quantitative Imaging in Medicine and
  Surgery}, vol.~12, no.~4, p. 2397, 2022.

\bibitem{hatamizadeh2022unetformer}
A.~Hatamizadeh, Z.~Xu, D.~Yang, W.~Li, H.~Roth, and D.~Xu, ``Unetformer: A
  unified vision transformer model and pre-training framework for 3d medical
  image segmentation,'' \emph{arXiv preprint arXiv:2204.00631}, 2022.

\bibitem{wang2022medical}
Q.~Wang, L.~Li, B.~Ni, Y.~Li, D.~Kong, C.~Wang, and Z.~Li, ``Medical image
  segmentation using transformer,'' in \emph{Artificial Intelligence in
  China}.\hskip 1em plus 0.5em minus 0.4em\relax Springer, 2022, pp. 92--99.

\bibitem{huang2021transde}
Z.~Huang, J.~Liao, J.~Wei, G.~Cai, and G.~Zhang, ``Transde: A transformer and
  double encoder network for medical image segmentation,'' in \emph{2021 11th
  International Conference on Information Technology in Medicine and Education
  (ITME)}.\hskip 1em plus 0.5em minus 0.4em\relax IEEE, 2021, pp. 374--378.

\bibitem{hille2022joint}
G.~Hille, S.~Agrawal, C.~Wybranski, M.~Pech, A.~Surov, and S.~Saalfeld, ``Joint
  liver and hepatic lesion segmentation using a hybrid cnn with transformer
  layers,'' \emph{arXiv preprint arXiv:2201.10981}, 2022.

\bibitem{wang2021multi}
L.~Wang, X.~Wang, B.~Zhang, X.~Huang, C.~Bai, M.~Xia, and P.~Sun, ``Multi-scale
  hierarchical transformer structure for 3d medical image segmentation,'' in
  \emph{2021 IEEE International Conference on Bioinformatics and Biomedicine
  (BIBM)}.\hskip 1em plus 0.5em minus 0.4em\relax IEEE, 2021, pp. 1542--1545.

\bibitem{li2022rdctrans}
L.~Li and H.~Ma, ``Rdctrans u-net: A hybrid variable architecture for liver ct
  image segmentation,'' \emph{Sensors}, vol.~22, no.~7, p. 2452, 2022.

\bibitem{wang2022multiscale}
B.~Wang, P.~Dong \emph{et~al.}, ``Multiscale transunet++: dense hybrid u-net
  with transformer for medical image segmentation,'' \emph{Signal, Image and
  Video Processing}, pp. 1--8, 2022.

\bibitem{wang2021boundary}
J.~Wang, L.~Wei, L.~Wang, Q.~Zhou, L.~Zhu, and J.~Qin, ``Boundary-aware
  transformers for skin lesion segmentation,'' in \emph{International
  Conference on Medical Image Computing and Computer-Assisted
  Intervention}.\hskip 1em plus 0.5em minus 0.4em\relax Springer, 2021, pp.
  206--216.

\bibitem{zhang2021multi}
Y.~Zhang, R.~Higashita, H.~Fu, Y.~Xu, Y.~Zhang, H.~Liu, J.~Zhang, and J.~Liu,
  ``A multi-branch hybrid transformer networkfor corneal endothelial cell
  segmentation,'' \emph{arXiv preprint arXiv:2106.07557}, 2021.

\bibitem{prangemeier2020attention}
T.~Prangemeier, C.~Reich, and H.~Koeppl, ``Attention-based transformers for
  instance segmentation of cells in microstructures,'' in \emph{2020 IEEE
  International Conference on Bioinformatics and Biomedicine (BIBM)}.\hskip 1em
  plus 0.5em minus 0.4em\relax IEEE, 2020, pp. 700--707.

\bibitem{chen2022pcat}
D.~Chen, W.~Yang, L.~Wang, S.~Tan, J.~Lin, and W.~Bu, ``Pcat-unet: Unet-like
  network fused convolution and transformer for retinal vessel segmentation,''
  \emph{PloS one}, vol.~17, no.~1, p. e0262689, 2022.

\bibitem{yun2021spectr}
B.~Yun, Y.~Wang, J.~Chen, H.~Wang, W.~Shen, and Q.~Li, ``Spectr: Spectral
  transformer for hyperspectral pathology image segmentation,'' \emph{arXiv
  preprint arXiv:2103.03604}, 2021.

\bibitem{marek2011parkinson}
K.~Marek, D.~Jennings, S.~Lasch, A.~Siderowf, C.~Tanner, T.~Simuni, C.~Coffey,
  K.~Kieburtz, E.~Flagg, S.~Chowdhury \emph{et~al.}, ``The parkinson
  progression marker initiative (ppmi),'' \emph{Progress in neurobiology},
  vol.~95, no.~4, pp. 629--635, 2011.

\bibitem{kamran2021vtgan}
S.~A. Kamran, K.~F. Hossain, A.~Tavakkoli, S.~L. Zuckerbrod, K.~M. Sanders, and
  S.~A. Baker, ``Vtgan: Semi-supervised retinal image synthesis and disease
  prediction using vision transformers,'' \emph{arXiv preprint
  arXiv:2104.06757}, 2021.

\bibitem{hajeb2012diabetic}
S.~Hajeb Mohammad~Alipour, H.~Rabbani, and M.~R. Akhlaghi, ``Diabetic
  retinopathy grading by digital curvelet transform,'' \emph{Computational and
  mathematical methods in medicine}, vol. 2012, 2012.

\bibitem{shin2020ganbert}
H.-C. Shin, A.~Ihsani, S.~Mandava, S.~T. Sreenivas, C.~Forster, J.~Cha, and
  A.~D.~N. Initiative, ``Ganbert: Generative adversarial networks with
  bidirectional encoder representations from transformers for mri to pet
  synthesis,'' \emph{arXiv preprint arXiv:2008.04393}, 2020.

\bibitem{hu2021data}
Z.~Hu, H.~Liu, Z.~Li, and Z.~Yu, ``Data-enabled intelligence in complex
  industrial systems cross-model transformer method for medical image
  synthesis,'' \emph{Complexity}, vol. 2021, 2021.

\bibitem{korkmaz2021unsupervised}
Y.~Korkmaz, S.~U. Dar, M.~Yurt, M.~{\"O}zbey, and T.~{\c{C}}ukur,
  ``Unsupervised mri reconstruction via zero-shot learned adversarial
  transformers,'' \emph{arXiv preprint arXiv:2105.08059}, 2021.

\bibitem{knoll2020fastmri}
F.~Knoll, J.~Zbontar, A.~Sriram, M.~J. Muckley, M.~Bruno, A.~Defazio,
  M.~Parente, K.~J. Geras, J.~Katsnelson, H.~Chandarana \emph{et~al.},
  ``fastmri: A publicly available raw k-space and dicom dataset of knee images
  for accelerated mr image reconstruction using machine learning,''
  \emph{Radiology: Artificial Intelligence}, vol.~2, no.~1, p. e190007, 2020.

\bibitem{ristea2021cytran}
N.-C. Ristea, A.-I. Miron, O.~Savencu, M.-I. Georgescu, N.~Verga, F.~S. Khan,
  and R.~T. Ionescu, ``Cytran: Cycle-consistent transformers for non-contrast
  to contrast ct translation,'' \emph{arXiv preprint arXiv:2110.06400}, 2021.

\bibitem{dalmaz2021resvit}
O.~Dalmaz, M.~Yurt, and T.~{\c{C}}ukur, ``Resvit: Residual vision transformers
  for multi-modal medical image synthesis,'' \emph{arXiv preprint
  arXiv:2106.16031}, 2021.

\bibitem{menze2014multimodal}
B.~H. Menze, A.~Jakab, S.~Bauer, J.~Kalpathy-Cramer, K.~Farahani, J.~Kirby,
  Y.~Burren, N.~Porz, J.~Slotboom, R.~Wiest \emph{et~al.}, ``The multimodal
  brain tumor image segmentation benchmark (brats),'' \emph{IEEE transactions
  on medical imaging}, vol.~34, no.~10, pp. 1993--2024, 2014.

\bibitem{bakas2017advancing}
S.~Bakas, H.~Akbari, A.~Sotiras, M.~Bilello, M.~Rozycki, J.~S. Kirby, J.~B.
  Freymann, K.~Farahani, and C.~Davatzikos, ``Advancing the cancer genome atlas
  glioma mri collections with expert segmentation labels and radiomic
  features,'' \emph{Scientific data}, vol.~4, no.~1, pp. 1--13, 2017.

\bibitem{bakas2018identifying}
S.~Bakas, M.~Reyes, A.~Jakab, S.~Bauer, M.~Rempfler, A.~Crimi, R.~T. Shinohara,
  C.~Berger, S.~M. Ha, M.~Rozycki \emph{et~al.}, ``Identifying the best machine
  learning algorithms for brain tumor segmentation, progression assessment, and
  overall survival prediction in the brats challenge,'' \emph{arXiv preprint
  arXiv:1811.02629}, 2018.

\bibitem{nyholm2018mr}
T.~Nyholm, S.~Svensson, S.~Andersson, J.~Jonsson, M.~Sohlin, C.~Gustafsson,
  E.~Kjell{\'e}n, K.~S{\"o}derstr{\"o}m, P.~Albertsson, L.~Blomqvist
  \emph{et~al.}, ``Mr and ct data with multiobserver delineations of organs in
  the pelvic area—part of the gold atlas project,'' \emph{Medical physics},
  vol.~45, no.~3, pp. 1295--1300, 2018.

\bibitem{feng2021task}
C.-M. Feng, Y.~Yan, H.~Fu, L.~Chen, and Y.~Xu, ``Task transformer network for
  joint mri reconstruction and super-resolution,'' \emph{arXiv preprint
  arXiv:2106.06742}, 2021.

\bibitem{zhang2021ptnet}
X.~Zhang, X.~He, J.~Guo, N.~Ettehadi, N.~Aw, D.~Semanek, J.~Posner, A.~Laine,
  and Y.~Wang, ``Ptnet: A high-resolution infant mri synthesizer based on
  transformer,'' \emph{arXiv preprint arXiv:2105.13993}, 2021.

\bibitem{makropoulos2018developing}
A.~Makropoulos, E.~C. Robinson, A.~Schuh, R.~Wright, S.~Fitzgibbon, J.~Bozek,
  S.~J. Counsell, J.~Steinweg, K.~Vecchiato, J.~Passerat-Palmbach
  \emph{et~al.}, ``The developing human connectome project: A minimal
  processing pipeline for neonatal cortical surface reconstruction,''
  \emph{Neuroimage}, vol. 173, pp. 88--112, 2018.

\bibitem{wang2021ted}
D.~Wang, Z.~Wu, and H.~Yu, ``Ted-net: Convolution-free t2t vision
  transformer-based encoder-decoder dilation network for low-dose ct
  denoising,'' \emph{arXiv preprint arXiv:2106.04650}, 2021.

\bibitem{mccollough2016tu}
C.~McCollough, ``Tu-fg-207a-04: Overview of the low dose ct grand challenge,''
  \emph{Medical physics}, vol.~43, no. 6Part35, pp. 3759--3760, 2016.

\bibitem{luthra2021eformer}
A.~Luthra, H.~Sulakhe, T.~Mittal, A.~Iyer, and S.~Yadav, ``Eformer: Edge
  enhancement based transformer for medical image denoising,'' \emph{arXiv
  preprint arXiv:2109.08044}, 2021.

\bibitem{jiang2021RDFNet}
H.~Jiang, P.~Zhang, C.~Che, and B.~Jin, ``Rdfnet: A fast caries detection
  method incorporating transformer mechanism,'' \emph{Computational and
  Mathematical Methods in Medicine}, vol. 2021, 2021.

\bibitem{shen2021COTR}
Z.~Shen, C.~Lin, and S.~Zheng, ``Cotr: Convolution in transformer network for
  end to end polyp detection,'' \emph{arXiv preprint arXiv:2105.10925}, 2021.

\bibitem{ma2021Transformer}
X.~Ma, G.~Luo, W.~Wang, and K.~Wang, ``Transformer network for significant
  stenosis detection in ccta of coronary arteries,'' in \emph{International
  Conference on Medical Image Computing and Computer-Assisted
  Intervention}.\hskip 1em plus 0.5em minus 0.4em\relax Springer, 2021, pp.
  516--525.

\bibitem{kong2021ctcad}
Q.~Kong, Y.~Wu, C.~Yuan, and Y.~Wang, ``Ct-cad: Context-aware transformers for
  end-to-end chest abnormality detection on x-rays,'' in \emph{2021 IEEE
  International Conference on Bioinformatics and Biomedicine (BIBM)}.\hskip 1em
  plus 0.5em minus 0.4em\relax IEEE, 2021, pp. 1385--1388.

\bibitem{tao2021spine}
R.~Tao and G.~Zheng, ``Spine-transformers: Vertebra detection and localization
  in arbitrary field-of-view spine ct with transformers,'' in
  \emph{International Conference on Medical Image Computing and
  Computer-Assisted Intervention}.\hskip 1em plus 0.5em minus 0.4em\relax
  Springer, 2021, pp. 93--103.

\bibitem{li2021x}
Y.~Li, Z.~Wang, L.~Yin, Z.~Zhu, G.~Qi, and Y.~Liu, ``X-net: a dual
  encoding--decoding method in medical image segmentation,'' \emph{The Visual
  Computer}, pp. 1--11, 2021.

\bibitem{liang2021swinir}
J.~Liang, J.~Cao, G.~Sun, K.~Zhang, L.~Van~Gool, and R.~Timofte, ``Swinir:
  Image restoration using swin transformer,'' in \emph{Proceedings of the
  IEEE/CVF International Conference on Computer Vision}, 2021, pp. 1833--1844.

\bibitem{yang2020learning}
F.~Yang, H.~Yang, J.~Fu, H.~Lu, and B.~Guo, ``Learning texture transformer
  network for image super-resolution,'' in \emph{Proceedings of the IEEE/CVF
  Conference on Computer Vision and Pattern Recognition}, 2020, pp. 5791--5800.

\bibitem{luo20213d}
Y.~Luo, Y.~Wang, C.~Zu, B.~Zhan, X.~Wu, J.~Zhou, D.~Shen, and L.~Zhou, ``3d
  transformer-gan for high-quality pet reconstruction,'' in \emph{International
  Conference on Medical Image Computing and Computer-Assisted
  Intervention}.\hskip 1em plus 0.5em minus 0.4em\relax Springer, 2021, pp.
  276--285.

\bibitem{balakrishnan2019voxelmorph}
G.~Balakrishnan, A.~Zhao, M.~R. Sabuncu, J.~Guttag, and A.~V. Dalca,
  ``Voxelmorph: a learning framework for deformable medical image
  registration,'' \emph{IEEE transactions on medical imaging}, vol.~38, no.~8,
  pp. 1788--1800, 2019.

\bibitem{chen2021transmorph}
J.~Chen, Y.~Du, Y.~He, W.~P. Segars, Y.~Li, and E.~C. Frey, ``Transmorph:
  Transformer for unsupervised medical image registration,'' \emph{arXiv
  preprint arXiv:2111.10480}, 2021.

\bibitem{ji2021progressively}
G.-P. Ji, Y.-C. Chou, D.-P. Fan, G.~Chen, H.~Fu, D.~Jha, and L.~Shao,
  ``Progressively normalized self-attention network for video polyp
  segmentation,'' \emph{arXiv preprint arXiv:2105.08468}, 2021.

\bibitem{kondo2021lapformer}
S.~Kondo, ``Lapformer: surgical tool detection in laparoscopic surgical video
  using transformer architecture,'' \emph{Computer Methods in Biomechanics and
  Biomedical Engineering: Imaging \& Visualization}, vol.~9, no.~3, pp.
  302--307, 2021.

\bibitem{czempiel2021opera}
T.~Czempiel, M.~Paschali, D.~Ostler, S.~T. Kim, B.~Busam, and N.~Navab,
  ``Opera: Attention-regularized transformers for surgical phase recognition,''
  \emph{arXiv preprint arXiv:2103.03873}, 2021.

\bibitem{reynaud2021ultrasound}
H.~Reynaud, A.~Vlontzos, B.~Hou, A.~Beqiri, P.~Leeson, and B.~Kainz,
  ``Ultrasound video transformers for cardiac ejection fraction estimation,''
  in \emph{International Conference on Medical Image Computing and
  Computer-Assisted Intervention}.\hskip 1em plus 0.5em minus 0.4em\relax
  Springer, 2021, pp. 495--505.

\bibitem{long2021dssr}
Y.~Long, Z.~Li, C.~H. Yee, C.~F. Ng, R.~H. Taylor, M.~Unberath, and Q.~Dou,
  ``E-dssr: efficient dynamic surgical scene reconstruction with
  transformer-based stereoscopic depth perception,'' in \emph{International
  Conference on Medical Image Computing and Computer-Assisted
  Intervention}.\hskip 1em plus 0.5em minus 0.4em\relax Springer, 2021, pp.
  415--425.

\bibitem{Prognosis-CLIMAT-nguyen2021climat}
H.~H. Nguyen, S.~Saarakkala, M.~B. Blaschko, and A.~Tiulpin, ``Climat:
  Clinically-inspired multi-agent transformers for disease trajectory
  forecasting from multi-modal data,'' \emph{arXiv preprint arXiv:2104.03642},
  2021.

\bibitem{GraphLearning-GrapBrain-zheng2021multimodal}
S.~Zheng, Z.~Zhu, Z.~Liu, Z.~Guo, Y.~Liu, and Y.~Zhao, ``Multi-modal graph
  learning for disease prediction,'' 2021.

\bibitem{GraphLearning-GraphAD-9433842}
Y.~Qiu, S.~Yu, Y.~Zhou, D.~Liu, X.~Song, T.~Wang, and B.~Lei, ``Multi-channel
  sparse graph transformer network for early alzheimer’s disease
  identification,'' in \emph{2021 IEEE 18th International Symposium on
  Biomedical Imaging (ISBI)}, 2021, pp. 1794--1797.

\bibitem{monajatipoor2021berthop}
M.~Monajatipoor, M.~Rouhsedaghat, L.~H. Li, A.~Chien, C.-C.~J. Kuo, F.~Scalzo,
  and K.-W. Chang, ``Berthop: An effective vision-and-language model for chest
  x-ray disease diagnosis,'' \emph{arXiv preprint arXiv:2108.04938}, 2021.

\bibitem{CNN+Trans-TransMed-dai2021transmed}
Y.~Dai, Y.~Gao, and F.~Liu, ``Transmed: Transformers advance multi-modal
  medical image classification,'' \emph{Diagnostics}, vol.~11, no.~8, p. 1384,
  2021.

\bibitem{prior-cxr}
G.~Jacenków, A.~Q. O’Neil, and S.~A. Tsaftaris, ``Indication as prior
  knowledge for multimodal disease classification in chest radiographs with
  transformers,'' in \emph{2022 IEEE 19th International Symposium on Biomedical
  Imaging (ISBI)}, 2022, pp. 1--5.

\bibitem{azzuni2022color}
H.~Azzuni, M.~Ridzuan, M.~Xu, and M.~Yaqub, ``Color space-based hover-net for
  nuclei instance segmentation and classification,'' \emph{arXiv preprint
  arXiv:2203.01940}, 2022.

\bibitem{MT-TransUNet}
\BIBentryALTinterwordspacing
J.~Chen, J.~Chen, Z.~Zhou, B.~Li, A.~Yuille, and Y.~Lu, ``Mt-transunet:
  Mediating multi-task tokens in transformers for skin lesion segmentation and
  classification,'' 2021. [Online]. Available:
  \url{https://arxiv.org/abs/2112.01767}
\BIBentrySTDinterwordspacing

\bibitem{sui2021cst}
D.~Sui, K.~Zhang, W.~Liu, J.~Chen, X.~Ma, and Z.~Tian, ``Cst: A multitask
  learning framework for colorectal cancer region mining based on
  transformer,'' \emph{BioMed Research International}, vol. 2021, 2021.

\bibitem{han2020survey}
K.~Han, Y.~Wang, H.~Chen, X.~Chen, J.~Guo, Z.~Liu, Y.~Tang, A.~Xiao, C.~Xu,
  Y.~Xu \emph{et~al.}, ``A survey on visual transformer,'' \emph{arXiv preprint
  arXiv:2012.12556}, 2020.

\bibitem{radford2021learning}
A.~Radford, J.~W. Kim, C.~Hallacy, A.~Ramesh, G.~Goh, S.~Agarwal, G.~Sastry,
  A.~Askell, P.~Mishkin, J.~Clark, G.~Krueger, and I.~Sutskever, ``Learning
  transferable visual models from natural language supervision,'' 2021.

\bibitem{MIL-Self-Attention-MILTrans-li2020deep}
Z.~Li, L.~Yuan, H.~Xu, R.~Cheng, and X.~Wen, ``Deep multi-instance learning
  with induced self-attention for medical image classification,'' in \emph{2020
  IEEE International Conference on Bioinformatics and Biomedicine
  (BIBM)}.\hskip 1em plus 0.5em minus 0.4em\relax IEEE, 2020, pp. 446--450.

\bibitem{MIL-SA-AbMILP-rymarczyk2021kernel}
D.~Rymarczyk, A.~Borowa, J.~Tabor, and B.~Zielinski, ``Kernel self-attention
  for weakly-supervised image classification using deep multiple instance
  learning,'' in \emph{Proceedings of the IEEE/CVF Winter Conference on
  Applications of Computer Vision}, 2021, pp. 1721--1730.

\bibitem{MIL-NoduleSATs-yang2020relational}
J.~Yang, H.~Deng, X.~Huang, B.~Ni, and Y.~Xu, ``Relational learning between
  multiple pulmonary nodules via deep set attention transformers,'' in
  \emph{2020 IEEE 17th International Symposium on Biomedical Imaging
  (ISBI)}.\hskip 1em plus 0.5em minus 0.4em\relax IEEE, 2020, pp. 1875--1878.

\bibitem{MIL-MIL-VT-yu2021mil}
S.~Yu, K.~Ma, Q.~Bi, C.~Bian, M.~Ning, N.~He, Y.~Li, H.~Liu, and Y.~Zheng,
  ``Mil-vt: Multiple instance learning enhanced vision transformer for fundus
  image classification,'' in \emph{International Conference on Medical Image
  Computing and Computer-Assisted Intervention}.\hskip 1em plus 0.5em minus
  0.4em\relax Springer, 2021, pp. 45--54.

\bibitem{MIL-TransMIL-shao2021transmil}
Z.~Shao, H.~Bian, Y.~Chen, Y.~Wang, J.~Zhang, X.~Ji, and Y.~Zhang, ``Transmil:
  Transformer based correlated multiple instance learning for whole slide image
  classication,'' \emph{arXiv preprint arXiv:2106.00908}, 2021.

\bibitem{Transfer-Transpath-wang2021transpath}
X.~Wang, S.~Yang, J.~Zhang, M.~Wang, J.~Zhang, J.~Huang, W.~Yang, and X.~Han,
  ``Transpath: Transformer-based self-supervised learning for histopathological
  image classification,'' in \emph{International Conference on Medical Image
  Computing and Computer-Assisted Intervention}.\hskip 1em plus 0.5em minus
  0.4em\relax Springer, 2021, pp. 186--195.

\bibitem{CNN-Transfer-truong2021transferable}
T.~Truong, S.~Mohammadi, and M.~Lenga, ``How transferable are self-supervised
  features in medical image classification tasks?'' \emph{arXiv preprint
  arXiv:2108.10048}, 2021.

\bibitem{Transfer-Learning-COVIDPrognosis-sriram2021covid}
A.~Sriram, M.~Muckley, K.~Sinha, F.~Shamout, J.~Pineau, K.~J. Geras, L.~Azour,
  Y.~Aphinyanaphongs, N.~Yakubova, and W.~Moore, ``Covid-19 prognosis via
  self-supervised representation learning and multi-image prediction,''
  \emph{arXiv preprint arXiv:2101.04909}, 2021.

\bibitem{MIL-Survey2018}
\BIBentryALTinterwordspacing
M.-A. Carbonneau, V.~Cheplygina, E.~Granger, and G.~Gagnon, ``Multiple instance
  learning: A survey of problem characteristics and applications,''
  \emph{Pattern Recognition}, vol.~77, p. 329–353, May 2018. [Online].
  Available: \url{http://dx.doi.org/10.1016/j.patcog.2017.10.009}
\BIBentrySTDinterwordspacing

\bibitem{luo2021semi}
X.~Luo, M.~Hu, T.~Song, G.~Wang, and S.~Zhang, ``Semi-supervised medical image
  segmentation via cross teaching between cnn and transformer,'' \emph{arXiv
  preprint arXiv:2112.04894}, 2021.

\bibitem{zhaocontext}
C.~Zhao, S.~Xiang, Z.~Cai, J.~Shen, S.~Li, S.~Zhou, D.~Zhao, W.~Su, S.~Guo
  \emph{et~al.}, ``Context-aware network for semi-supervised segmentation of 3d
  left atrium,'' \emph{Context-Aware Network for Semi-Supervised Segmentation
  of 3d Left Atrium}.

\bibitem{xiao4081789efficient}
Z.~Xiao, Y.~Su, Z.~Deng, and W.~Zhang, ``Efficient combination of cnn and
  transformer for dual-teacher uncertainty-aware guided semi-supervised medical
  image segmentation,'' \emph{Available at SSRN 4081789}.

\bibitem{Transfer-Survey-zhuang2020comprehensive}
F.~Zhuang, Z.~Qi, K.~Duan, D.~Xi, Y.~Zhu, H.~Zhu, H.~Xiong, and Q.~He, ``A
  comprehensive survey on transfer learning,'' \emph{Proceedings of the IEEE},
  vol. 109, no.~1, pp. 43--76, 2020.

\bibitem{CNN-VIT-Pretrain-matsoukas2021time}
C.~Matsoukas, J.~F. Haslum, M.~S{\"o}derberg, and K.~Smith, ``Is it time to
  replace cnns with transformers for medical images?'' \emph{arXiv preprint
  arXiv:2108.09038}, 2021.

\bibitem{luo2022self}
Y.~Luo, Z.~Chen, and X.~Gao, ``Self-distillation augmented masked autoencoders
  for histopathological image classification,'' \emph{arXiv preprint
  arXiv:2203.16983}, 2022.

\bibitem{malkiel2021pre}
I.~Malkiel, G.~Rosenman, L.~Wolf, and T.~Hendler, ``Pre-training and
  fine-tuning transformers for fmri prediction tasks,'' \emph{arXiv preprint
  arXiv:2112.05761}, 2021.

\bibitem{xie2021unified}
Y.~Xie, J.~Zhang, Y.~Xia, and Q.~Wu, ``Unified 2d and 3d pre-training for
  medical image classification and segmentation,'' \emph{arXiv preprint
  arXiv:2112.09356}, 2021.

\bibitem{chen2022self}
R.~J. Chen and R.~G. Krishnan, ``Self-supervised vision transformers learn
  visual concepts in histopathology,'' \emph{arXiv preprint arXiv:2203.00585},
  2022.

\bibitem{li2022transbtsv2}
J.~Li, W.~Wang, C.~Chen, T.~Zhang, S.~Zha, H.~Yu, and J.~Wang, ``Transbtsv2:
  Wider instead of deeper transformer for medical image segmentation,''
  \emph{arXiv preprint arXiv:2201.12785}, 2022.

\bibitem{CNN-ViT-Transfer-krishnamurthy2021}
S.~Krishnamurthy, K.~Srinivasan, S.~M. Qaisar, P.~Vincent, and C.-Y. Chang,
  ``Evaluating deep neural network architectures with transfer learning for
  pneumonitis diagnosis,'' \emph{Computational and Mathematical Methods in
  Medicine}, vol. 2021, 2021.

\bibitem{CNN-ViT-CellRatio-liu2021aspect}
W.~Liu, C.~Li, M.~M. Rahamana, H.~Sun, W.~Hu, H.~Chen, C.~Sun, Y.~Yao, and
  M.~Grzegorzek, ``Is aspect ratio of cells important in deep learning? a
  robust comparison of deep learning methods for multi-scale cytopathology cell
  image classification: from convolutional neural networks to visual
  transformers,'' 2021.

\bibitem{Compare-Implant-zhou2021shoulder}
M.~Zhou and S.~Mo, ``Shoulder implant x-ray manufacturer classification:
  Exploring with vision transformer,'' 2021.

\bibitem{TransAD-altay2020preclinical}
F.~Altay, G.~R. Sanchez, Y.~James, S.~V. Faraone, S.~Velipasalar, and
  A.~Salekin, ``Preclinical stage alzheimer's disease detection using magnetic
  resonance image scans,'' 2020.

\bibitem{cnn-vit-low-resolution}
I.~Adjei-Mensah, X.~Zhang, A.~A. Baffour, I.~O. Agyemang, S.~B. Yussif,
  B.~L.~Y. Agbley, and C.~Sey, ``Investigating vision transformer models for
  low-resolution medical image recognition,'' in \emph{2021 18th International
  Computer Conference on Wavelet Active Media Technology and Information
  Processing (ICCWAMTIP)}, 2021, pp. 179--183.

\bibitem{cassidy2021diabetic}
B.~Cassidy, C.~Kendrick, N.~D. Reeves, J.~M. Pappachan, C.~O’Shea, D.~G.
  Armstrong, and M.~H. Yap, ``Diabetic foot ulcer grand challenge 2021:
  evaluation and summary,'' in \emph{Diabetic Foot Ulcers Grand
  Challenge}.\hskip 1em plus 0.5em minus 0.4em\relax Springer, 2021, pp.
  90--105.

\end{thebibliography}

%








\end{document}